\documentclass[10pt,journal,compsoc]{IEEEtran}
\ifCLASSOPTIONcompsoc
  \usepackage[nocompress]{cite}
\else
  \usepackage{cite}
\fi
\usepackage{algorithm}
\usepackage{algorithmic}
\usepackage{graphicx}
\usepackage{epstopdf}
\usepackage{subfigure}
\usepackage{soul}
\usepackage{multicol}
\usepackage{multirow}
\usepackage{float}
\usepackage{enumitem}
\usepackage{booktabs}
\usepackage{amsmath}
\usepackage{amsfonts}
\usepackage{bm}
\usepackage{url}
\usepackage[normalem]{ulem}
\useunder{\uline}{\ul}{}

\ifCLASSINFOpdf
\else
\fi
\usepackage{amsmath}

\makeatletter
\def\endthebibliography{%
  \def\@noitemerr{\@latex@warning{Empty `thebibliography' environment}}%
  \endlist
}
\makeatother
\begin{document}
%
% paper title
% Titles are generally capitalized except for words such as a, an, and, as,
% at, but, by, for, in, nor, of, on, or, the, to and up, which are usually
% not capitalized unless they are the first or last word of the title.
% Linebreaks \\ can be used within to get better formatting as desired.
% Do not put math or special symbols in the title.
\title{Graph Adversarial Immunization for Certifiable Robustness}
%
%
% author names and IEEE memberships
% note positions of commas and nonbreaking spaces ( ~ ) LaTeX will not break
% a structure at a ~ so this keeps an author's name from being broken across
% two lines.
% use \thanks{} to gain access to the first footnote area
% a separate \thanks must be used for each paragraph as LaTeX2e's \thanks
% was not built to handle multiple paragraphs
%
%
%\IEEEcompsocitemizethanks is a special \thanks that produces the bulleted
% lists the Computer Society journals use for "first footnote" author
% affiliations. Use \IEEEcompsocthanksitem which works much like \item
% for each affiliation group. When not in compsoc mode,
% \IEEEcompsocitemizethanks becomes like \thanks and
% \IEEEcompsocthanksitem becomes a line break with idention. This
% facilitates dual compilation, although admittedly the differences in the
% desired content of \author between the different types of papers makes a
% one-size-fits-all approach a daunting prospect. For instance, compsoc 
% journal papers have the author affiliations above the "Manuscript
% received ..."  text while in non-compsoc journals this is reversed. Sigh.

\author{Shuchang~Tao, Qi~Cao\IEEEauthorrefmark{1}, Huawei~Shen\IEEEauthorrefmark{1}\thanks{\IEEEauthorrefmark{1}Corresponding authors},~\IEEEmembership{Member,~IEEE}, Yunfan~Wu, Liang~Hou,\\ and Xueqi~Cheng,~\IEEEmembership{Senior Member,~IEEE}%
\IEEEcompsocitemizethanks{
  \IEEEcompsocthanksitem
            Shuchang Tao, Qi Cao, Huawei Shen, Yunfan Wu and Liang Hou are with the Data Intelligence System Research Center, Institute of Computing Technology, Chinese Academy of Sciences (CAS) and University of Chinese Academy of Sciences (UCAS), Beijing, 100190, China.  
            \protect\\ E-mail:
            \protect \{taoshuchang18z,caoqi,shenhuawei,wuyunfan19b,\\houliang17z\}@ict.ac.cn
%  \IEEEcompsocthanksitem
%            Qi Cao, Huawei Shen, Yunfan Wu and Liang Hou are with the Data Intelligence System Research Center, Institute of Computing Technology (ICT), Chinese Academy of Sciences (CAS), Beijing, 100190, China.
%            \protect\\ E-mail:
%            \protect \{caoqi,shenhuawei,wuyunfan19b,houliang\}@ict.ac.cn
  \IEEEcompsocthanksitem
           	Xueqi Cheng is with the CAS Key Laboratory of Network Data Science and Technology, Institute of Computing Technology (ICT), Chinese Academy of Sciences (CAS), Beijing, 100190, China.
            \protect\\ E-mail:
            \protect cxq@ict.ac.cn
}
}
\markboth{Journal of Transactions on Knowledge and Data Engineering,~Vol.~14, No.~8, August~2023}%
{Shell \MakeLowercase{\textit{et al.}}: Bare Demo of IEEEtran.cls for Computer Society Journals}
% The only time the second header will appear is for the odd numbered pages
% after the title page when using the twoside option.
% 
% *** Note that you probably will NOT want to include the author's ***
% *** name in the headers of peer review papers.                   ***
% You can use \ifCLASSOPTIONpeerreview for conditional compilation here if
% you desire.

% The publisher's ID mark at the bottom of the page is less important with
% Computer Society journal papers as those publications place the marks
% outside of the main text columns and, therefore, unlike regular IEEE
% journals, the available text space is not reduced by their presence.
% If you want to put a publisher's ID mark on the page you can do it like
% this:
%\IEEEpubid{0000--0000/00\$00.00~\copyright~2015 IEEE}
% or like this to get the Computer Society new two part style.
%\IEEEpubid{\makebox[\columnwidth]{\hfill 0000--0000/00/\$00.00~\copyright~2015 IEEE}%
%\hspace{\columnsep}\makebox[\columnwidth]{Published by the IEEE Computer Society\hfill}}
% Remember, if you use this you must call \IEEEpubidadjcol in the second
% column for its text to clear the IEEEpubid mark (Computer Society jorunal
% papers don't need this extra clearance.)

% use for special paper notices
%\IEEEspecialpapernotice{(Invited Paper)}

% for Computer Society papers, we must declare the abstract and index terms
% PRIOR to the title within the \IEEEtitleabstractindextext IEEEtran
% command as these need to go into the title area created by \maketitle.
% As a general rule, do not put math, special symbols or citations
% in the abstract or keywords.
\IEEEtitleabstractindextext{%
\begin{abstract}
%Despite achieving great success in semi-supervised node classification task, graph neural networks (GNNs) are vulnerable to adversarial attacks. Existing researches focus on developing either adversarial training or robust GNN models against adversarial attacks on graphs. However, little research attention is paid to the potential and practice of immunization to adversarial attacks on graphs. In this paper, we propose and formulate \emph{graph adversarial immunization}, i.e., vaccinating an affordable fraction of node pairs or nodes to improve the certifiable robustness of graph against any admissible adversarial attack, noted as edge-level and node-level immunization. We further propose two effective algorithms, called \emph{AdvImmune-Edge} and \emph{AdvImmune-Node}, which circumvent the computationally expensive combinatorial optimization when solving the adversarial immunization problem. Experiments conducted on three benchmark datasets demonstrate that the proposed AdvImmune-Edge and AdvImmune-Node significantly outperform baselines. In particular, AdvImmune-Node remarkably improves the ratio of robust nodes by 79$\%$, 294$\%$, 100$\%$, with an affordable immune budget of only 5$\%$ nodes. Furthermore, AdvImmune methods show excellent defensive performance against various attacks, outperforming state-of-the-art defense methods. To the best of our knowledge, this is the first work fto improve robustness from a graph data perspective without losing performance on clean graphs, providing a brand new insight   into graph adversarial learning.
Despite achieving great success, graph neural networks (GNNs) are vulnerable to adversarial attacks. Existing defenses focus on developing adversarial training or model modification. 
In this paper, we propose and formulate \emph{graph adversarial immunization}, i.e., vaccinating part of graph structure to improve certifiable robustness of graph against any admissible adversarial attack. We first propose edge-level immunization to vaccinate node pairs. Unfortunately, such edge-level immunization cannot defend against emerging node injection attacks, since it only immunizes existing node pairs. To this end, we further propose node-level immunization. To avoid computationally intensive combinatorial optimization associated with adversarial immunization, we develop \emph{AdvImmune-Edge} and \emph{AdvImmune-Node} algorithms to effectively obtain the immune node pairs or nodes. Extensive experiments demonstrate the superiority of AdvImmune methods. In particular, AdvImmune-Node remarkably improves the ratio of robust nodes by 79$\%$, 294$\%$, and 100$\%$, after immunizing only 5$\%$ of nodes. Furthermore, AdvImmune methods show excellent defensive performance against various attacks, outperforming state-of-the-art defenses. To the best of our knowledge, this is the first attempt to improve certifiable robustness from graph data perspective without losing performance on clean graphs, providing new insights into graph adversarial learning.  
\end{abstract}

% Note that keywords are not normally used for peerreview papers.
}

% make the title area
\maketitle

% To allow for easy dual compilation without having to reenter the
% abstract/keywords data, the \IEEEtitleabstractindextext text will
% not be used in maketitle, but will appear (i.e., to be "transported")
% here as \IEEEdisplaynontitleabstractindextext when the compsoc 
% or transmag modes are not selected <OR> if conference mode is selected 
% - because all conference papers position the abstract like regular
% papers do.
\IEEEdisplaynontitleabstractindextext
% \IEEEdisplaynontitleabstractindextext has no effect when using
% compsoc or transmag under a non-conference mode.

% For peer review papers, you can put extra information on the cover
% page as needed:
% \ifCLASSOPTIONpeerreview
% \begin{center} \bfseries EDICS Category: 3-BBND \end{center}
% \fi
%
% For peerreview papers, this IEEEtran command inserts a page break and
% creates the second title. It will be ignored for other modes.
\IEEEpeerreviewmaketitle

\IEEEraisesectionheading{\section{Introduction}\label{sec:introduction}}
% Computer Society journal (but not conference!) papers do something unusual
% with the very first section heading (almost always called "Introduction").
% They place it ABOVE the main text! IEEEtran.cls does not automatically do
% this for you, but you can achieve this effect with the provided
% \IEEEraisesectionheading{} command. Note the need to keep any \label that
% is to refer to the section immediately after \section in the above as
% \IEEEraisesectionheading puts \section within a raised box.

% The very first letter is a 2 line initial drop letter followed
% by the rest of the first word in caps (small caps for compsoc).
% 
% form to use if the first word consists of a single letter:
% \IEEEPARstart{A}{demo} file is ....
% 
% form to use if you need the single drop letter followed by
% normal text (unknown if ever used by the IEEE):
% \IEEEPARstart{A}{}demo file is ....
% 
% Some journals put the first two words in caps:
% \IEEEPARstart{T}{his demo} file is ....
% 
% Here we have the typical use of a "T" for an initial drop letter
% and "HIS" in caps to complete the first word.

\IEEEPARstart{G}{raph} Neural Networks (GNNs) have achieved significant success in many tasks, such as node classification~\cite{kipf2017semi,Klicpera2018PredictTP,xu2018gwnn}, cascade prediction~\cite{Cao2020PopularityPO}, recommender systems~\cite{fan2019graph}, and fraud detection~\cite{Ma2021Survey,Cheng2022Fraud}. However, GNNs have been shown to be vulnerable to adversarial attacks~\cite{Dai2018AdversarialAO,zugner2018adversarial,Bojchevski2018AdversarialAO,TaoGNIA,ZouTDGIA}, i.e., imperceptible perturbations on graph data can easily fool GNNs. Such vulnerability of GNNs may cause tremendous security risks, especially in the safety-critical area. Take fraud detection as an example, fraudsters fool GNN by perturbing graphs to misclassify fraudsters, so as to defraud loans and cause serious economic losses.

Numerous works focus on developing defense strategies against adversarial attacks on GNN models, including adversarial training~\cite{feng2019graph,Dai2019AdversarialTM,kong2020flag}, model modification~\cite{ZhangGNNGuard2020,Zhu2019RobustGC}, and so on~\cite{zhang2019comparing, Entezari2020AllYN,Jin2020GraphSL}. 
These methods have emerged swiftly and have demonstrated their effectiveness in enhancing the performance of GNNs~\cite{Wu2019AdversarialEF,Xu2019TopologyAA}. Nonetheless, these approaches are typically heuristic and their efficacy is limited to specific attacks, rather than being universally applicable to all attacks~\cite{Bojchevski2019CertifiableRT}.
As a result, an endless cat-and-mouse game ensues between adversarial attacks and defense strategies~\cite{Bojchevski2019CertifiableRT}. To address this the attack-defense dilemma, recent studies have turned to robustness certification and robust training on graphs~\cite{zugner2020CertiRob,Liu2020CertifiableRT,Bojchevski2019CertifiableRT} against any admissible adversarial attack~\cite{zugner_adversarial_2019,bojchevski_sparsesmoothing_2020, WangJCG21,SchuchardtBKG21}. However, this robust training may inadvertently impair the performance of GNN models on the clean graph, an outcome that is not desirable in the absence of adversarial attacks.

\begin{figure}
\centering
\subfigure[Edge-level adversarial immunization]{
\includegraphics[width=8.2cm]{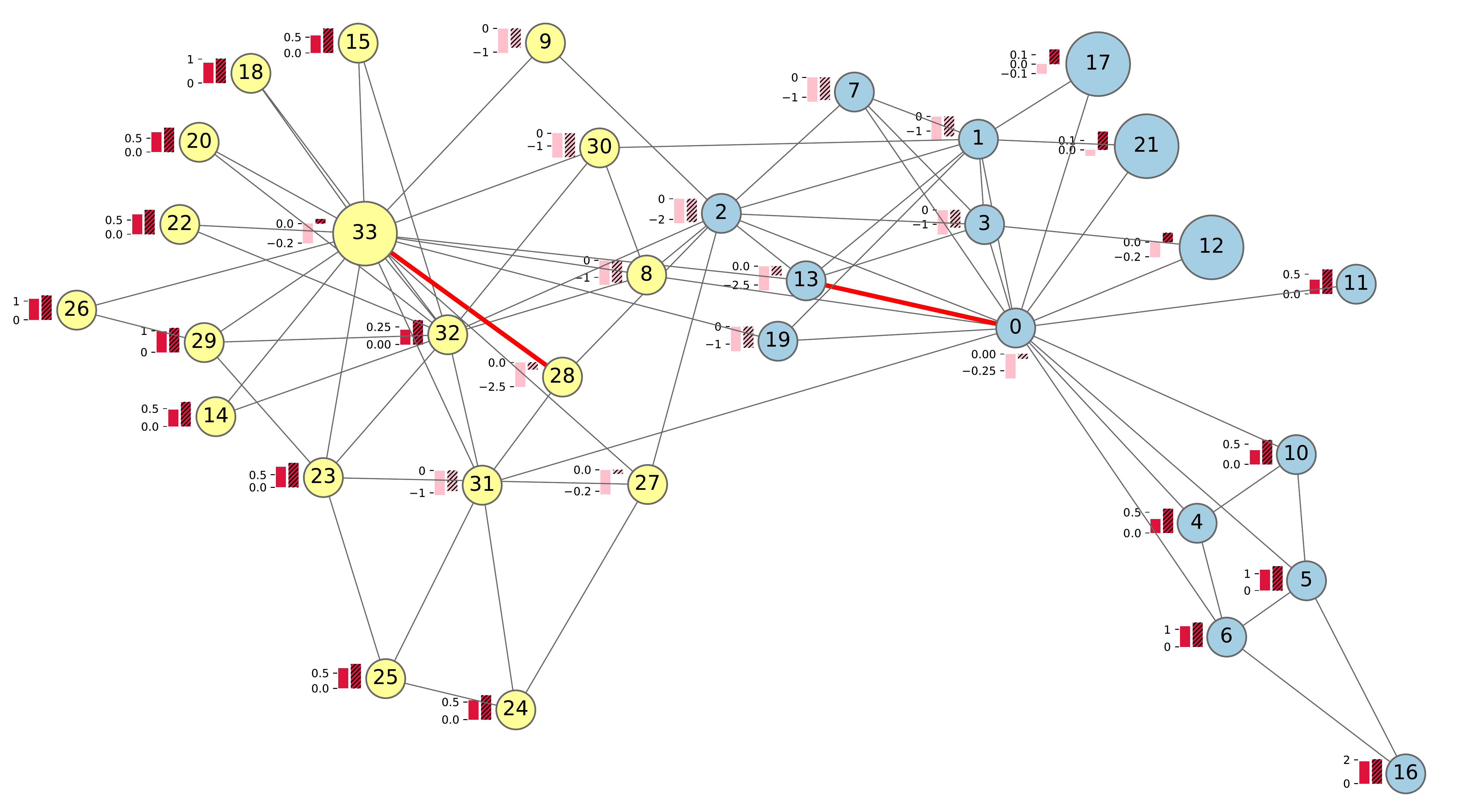}
\label{subfig:karate_edge}
}
\subfigure[Node-level adversarial immunization]{
\includegraphics[width=8.2cm]{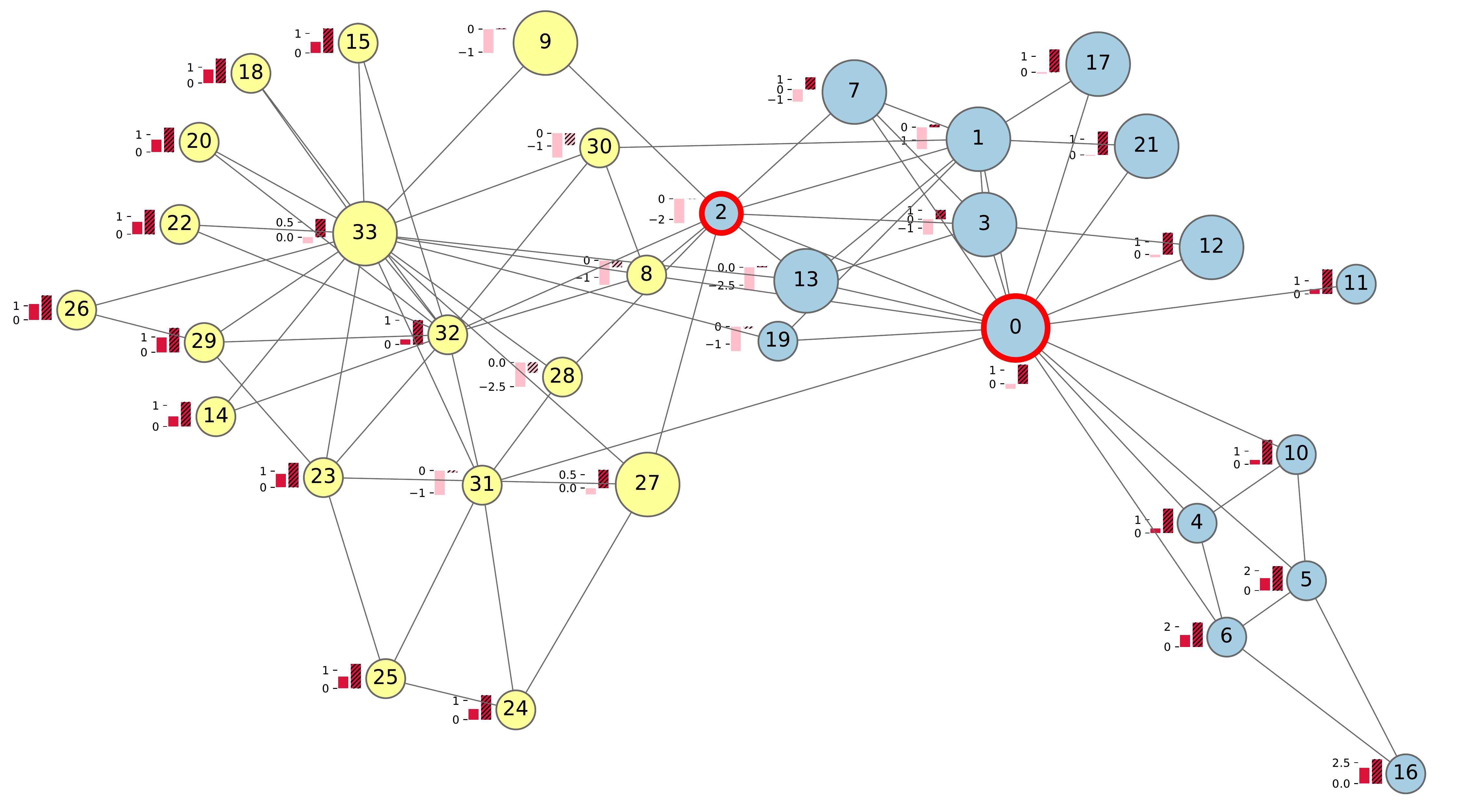}
\label{subfig:karate_node}
}
\subfigure{
\includegraphics[width=6.7cm]{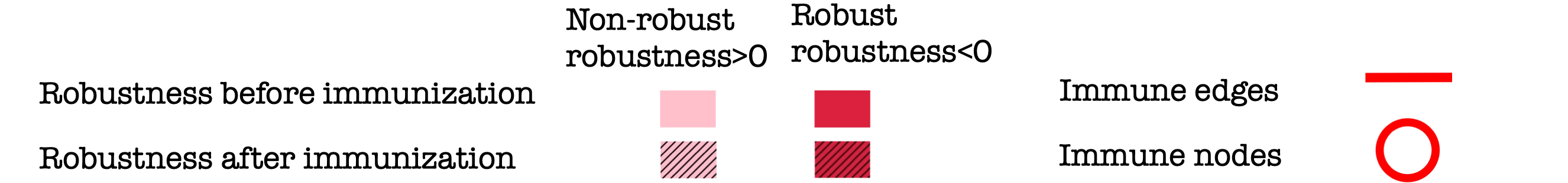}
}
\caption{Effect of graph adversarial immunization on Karate club network. Colors differentiate nodes into two classes. The red edges/circles indicate the immune edges/nodes. Two bars represent the node's robustness before immunization (non-striped background) and after immunization (striped background). The color of bar indicates whether the node is certified as robust: a red bar refers to a robust node (node's robustness $> 0$, such as Node 20), a pink bar refers to a non-robust node (node's robustness $< 0$, such as Node 8). A larger node size denotes a node that becomes robust through the immunization process.
%The node is certified as robust, when its robustness $>$ 0  (red), otherwise as non-robust (pink).
}
\label{fig:karate}
\end{figure}
%% \vspace{-5pt}

We propose and formulate \emph{graph adversarial immunization}, a pioneering guideline for enhancing certifiable robustness against any permissible adversarial attack from graph data perspective. The concept of \emph{adversarial immunization} involves vaccinating a part of node pairs or individual nodes in advance to protect them from attack modifications, thereby enhancing the graph's overall robustness. We propose both node-level and edge-level immunization for different scenarios. Edge-level immunization targets edges or node pairs, suitable for situations requiring fine immunization. In contrast, node-level immunization protects nodes from having their connections or disconnections to other nodes modified, making it more practical and better suited for defending against emerging node injection attacks~\cite{TaoGNIA, ZouTDGIA}. Importantly, adversarial immunization is a versatile strategy that not only enhances certifiable robustness against any permissible attack but also avoids performance degradation on the unattacked clean graph.

We propose two algorithms called AdvImmune-Edge and AdvImmune-Node for edge-level and node-level immunization, respectively. To circumvent the computational cost of combinatorial optimization when solving adversarial immunization, AdvImmune-Edge obtains the immune edges or node pairs with meta-gradient in a discrete way, and AdvImmune-Node solves node-level immunization by designing and computing robustness gain. 
These two methods are collectively referred to as AdvImmune methods.
%% with meta-gradient in a discrete way, 
% circumventing the computationally expensive combinatorial optimization when solving adversarial immunization. 

To provide a clear understanding of adversarial immunization, we demonstrate its effect on the Karate club network. For edge-level immunization, Figure~\ref{subfig:karate_edge} shows that immunizing just 2 edges results in an increase of 4 nodes (larger nodes) that are certified as robust against any permissible attack. 
Intuitively, we have observed that the immune edges (Edge 28-33 and Edge 13-0) connect large-degree nodes with nodes of the same class. Specifically, Node 33 and Node 0 possess the highest degree, respectively. Moreover, Node 28 is connected to Node 2, belonging to a different class. If an attacker removes Edge 28-33, the model may misclassify Node 28 as  Node 2’s class, because the model may overemphasize the information from the neighbor Node 2. Similarly, Node 13 is also connected to Node 33, which belongs to another class.
%Intuitively, this effectiveness stems primarily from the immunization of crucial edges, which mitigates the impact of attacks. In other words, perturbing some crucial edges could lead to a high misclassification rate, however, by immunizing these vital edges, we force the attack to target less influential edges, resulting in a degradation of attack performance.
As for node-level immunization, Figure~\ref{subfig:karate_node} illustrates that 11 more nodes become robust against any admissible attack, only immunizing 2 nodes in advance.
%In Figure~\ref{subfig:karate_node}, node No.2 is immunized, but it does not become robust, because its robustness was too low at the beginning. After immunization, the robustness of this node has been greatly improved, from -2.35 to -0.02. 
Indeed, AdvImmune methods exhibit a significant improvement in certifiable robustness when applied to real large-scale networks. 
Results on Reddit reveal that with a moderate immune budget of merely 5\% nodes, the ratio of robust nodes improves by 294\%. 
Such findings demonstrate the effectiveness of AdvImmune methods, i.e., immunizing a reasonable fraction of critical nodes or node pairs in advance can notably improve the certifiable robustness of graph.

%\begin{figure}
%\centering
%\subfigure[Node-level immunization]{
%\includegraphics[width=8cm]{figures/karate_node.pdf}
%\label{subfig:karate_node}
%}
%\includegraphics[width=7cm]{figures/karate_node.pdf}
%\caption{Effect of node-level adversarial immunization on Karate club network. 
%%Colors differentiate nodes into two classes.  
%%Two bars represent the node's robustness before and after immunization.
%%We use two bars to represent the node's robustness before and after immunization. 
%%The node is certified as robust (red), when its robustness $>$ 0, otherwise as non-robust (pink).
%%The larger node represents the node that becomes robust through immunization.
%Red circles indicate the immune nodes.
%}
%\label{fig:karate_node}
%\end{figure}

Adversarial immunization has much potential in real-world applications. For example, in credit scoring systems~\cite{Jin2020AdversarialAA}, immunization maintains certain critical users or the relations between user pairs to prevent fraudsters from pretending to be normal customers, avoiding serious financial losses. 

The initial publication on edge-level immunization was presented at the 14th ACM International Conference on Web Search and Data Mining (WSDM '21)~\cite{Tao2021AdvImmune}. The conference version highlights the following contributions:
%We summarize the main contributions as follows:
\begin{enumerate}[topsep = 0.5 em]
\item This pioneering work introduces and formalizes \emph{adversarial immunization}, improving certifiable robustness against any admissible attack and contributing fresh perspectives to graph adversarial learning.
\item We propose AdvImmune-Edge, an innovative approach that utilizes meta-gradient for adversarial immunization, thereby circumventing the need for expensive combinatorial optimization.
\item Comprehensive experiments validate the effectiveness of AdvImmune-Edge on three benchmark datasets, substantially enhancing the ratio of robust nodes.
\end{enumerate}

The conference version~\cite{Tao2021AdvImmune} focuses on edge-level immunization.
Recently, the emerging node injection attacks have raised great attention~\cite{Sun2020AdversarialAO, Wang2020ScalableAO, TaoGNIA, ZouTDGIA}, where attacker injects malicious nodes to degrade GNN performance. Edge-level immunization cannot deal with these attacks since it can only immunize the existing edges or node pairs.
Besides, when facing some tough attacks, AdvImmune-Edge may require quite a lot of immune budgets, and has to calculate and immunize them one by one, resulting in high computational cost. 
To deal with these situations, we further propose a new \emph{node-level immunization, which vaccinates a fraction of nodes in advance, to protect the connections/disconnections of them from being modified by attacks}.

However, the transition from edge-level to node-level immunization is not a simple extension, but rather presents unique challenges. 
The worst-case perturbed graphs differ before and after immunization, leading to inaccuracies in estimating the impact of node pairs or nodes. This issue is further magnified in node-level immunization, potentially resulting in suboptimal immune nodes. 
To address these challenges, we propose \emph{AdvImmune-Node}. Specifically, we design robustness gain to accurately estimate impact of node. 
To mitigate time-consuming computations, we propose additional strategies including candidate selection and lazy update.  
AdvImmune-Node defends against node injection attacks by protecting certain nodes from being connected to any malicious node.
It is more flexible to tackle various attack scenarios with an affordable immune budget. 
We summarize the main additional contributions:
\begin{enumerate}[topsep = 0.5 em]
\item We propose node-level  imunization and unify both node-level and edge-level immunization into a general formalization.
\item We propose AdvImmune-Node algorithm to solve node-level immunization effectively, by designing and computing robustness gain.  
\item Experiments demonstrate AdvImmune-Node remarkably improves ratio of robust nodes by 79$\%$, 294$\%$, 100$\%$, with an affordable immune budget of 5$\%$ nodes. 
\item AdvImmune methods exhibit excellent performance against both node injection and graph modification attacks, outperforming state-of-the-art defense methods.~\footnote{The code and data used in the paper is released at  \url{https://github.com/TaoShuchang/AdvImmune}/}

\end{enumerate}

\section{Preliminaries}
Adversarial learning on graphs mainly targets semi-supervised node classification. This section initially introduces this task, along with a widely used GNN to tackle it. Besides, we discuss the robustness certification against any admissible attack. 

\textbf{Semi-supervised node classification.}
Given an attributed graph $G=(\mathbf{A}, \mathbf{X})$, $ \mathbf{A} \in\{0,1\}^{N \times N} $ is the adjacency matrix and $\mathbf{X} \in \mathbb{R}^{N \times d}$ is the attribute matrix consisting of node attributes, $N$ is the number of nodes and $d$ is the dimension of node attributes. 
In semi-supervised node classification, a subset of nodes  are labelled from class sets $\mathcal{K}$.
The goal is to assign a label for each unlabelled node by the learned classifier $\mathbf{Y}=f(\mathbf{A}, \mathbf{X}) \in \mathbb{R}^{N \times K}$, where $K=|\mathcal{K}|$ is the number of classes.

\textbf{Graph neural networks.}
GNNs have shown significant success in semi-supervised node classification task~\cite{kipf2017semi, velickovic2018graph, Klicpera2018PredictTP}. 
Among existing GNNs, $\pi$-PPNP~\cite{Klicpera2018PredictTP,Bojchevski2019CertifiableRT} exhibits exceptional performance.
It links graph convolutional networks (GCN) to PageRank to effectively capture the impact of infinitely neighborhood aggregation, and separates feature transformation from propagation to simplify model structure. In this paper, we follow Bojchevski \textit{et al.}~\cite{Bojchevski2019CertifiableRT} using $\pi$-PPNP to address node classification. The formulation is:
\begin{equation}
	\begin{aligned}
    \mathbf{Y} =\operatorname{softmax}\left(\mathbf{\Pi}\mathbf{H}\right), \quad \mathbf{H} =f_{\theta}\left(\mathbf{X}\right),
    \label{eq:PPNP}
\end{aligned}
\end{equation}
where $\mathbf{H} \in \mathbb{R}^{N \times K}$ is the transformed features computed by a neural network $f_{\theta}$, and $\theta$ is the parameters of the neural network, $\mathbf{H}^{\text {diff }}:=\mathbf{\Pi} \mathbf{H}$ is defined as the \textit{diffused logits}, which are the unnormalized outputs with real numbers ranging from $(-\infty,+\infty)$, often referred to as raw predictions. $\mathbf{\Pi} =(1-\alpha)\left(\mathbf{I}_{N}-\alpha \mathbf{D}^{-1} \mathbf{A} \right)^{-1}$ is personalized PageRank~\cite{Page1999ThePC} that measures distance between nodes, and $\mathbf{D}$ is a diagonal degree matrix with $\mathbf{D}_{i i}=\sum_{j} \mathbf{A}_{i j}$. 
The personalized PageRank  with source node $t$ on graph $G$ is written as: 
$
	\mathbf{\pi}_G\left(\mathbf{e}_{t}\right)= (1-\alpha) \mathbf{e}_{t}\left(\mathbf{I}_{N}-\alpha \mathbf{D}^{-1} \mathbf{A}\right)^{-1}, 
$
where $\mathbf{e}_t$ is the $t$-th canonical basis vector (row vector).  
$\mathbf{\pi}_G\left(\mathbf{e}_{t}\right)=\mathbf{\Pi}_{t,:}$ is the $t$-th row of personalized PageRank matrix $\mathbf{\Pi}$.

\textbf{Robustness certification.} 
\label{sec:graph_cert}
Since GNN models are vulnerable to adversarial attacks, 
Bojchevski \textit{et al.} ~\cite{Bojchevski2019CertifiableRT} certify the robustness of each node against any admissible attack on graph by \textit{worst-case margin}.
Specifically, the difference between the raw predictions of node $t$ on label class $y_t$ and other class $k$ defines the \textit{margin} on the perturbed graph $\tilde{G}$. Taking $\pi$-PPNP as a typical GNN model, the formula of the defined margin is as follows:
 \begin{equation}
 	 m_{y_t, k}(t,\tilde{G})
 	 =\mathbf{H}_{t, y_t}^{\text {diff}}-\mathbf{H}_{t, k}^{\mathrm{diff}}
 	 =\mathbf{\pi}_{\tilde{G}}\left(\mathbf{e}_{t}\right)\left(\mathbf{H}_{:, y_t}-\mathbf{H}_{:, k}\right),
 	  	\label{eq:w-margin cert}
 \end{equation}
where $y_t$ is the label class of node $t$.

For target node $t$, the \textit{worst-case margin} between class $y_t$ and class $k$ under any admissible perturbation $\tilde{G} \in \mathcal{Q}$ is:
\begin{equation}
\begin{aligned}
m_{y_{t}, k}(t,\tilde{G}^{*})&=\min _{\tilde{G} \in \mathcal{Q}} m_{y_{t}, k}(t,\tilde{G})\\
&=\min _{\tilde{G} \in \mathcal{Q}} \mathbf{\pi}_{\tilde{G}}\left(\mathbf{e}_{t}\right)\left(\mathbf{H}_{:, y_{t}}-\mathbf{H}_{:, k}\right),	
\end{aligned}
\label{eq:PPNP_cert}
\end{equation}
where $\tilde{G}$ is the perturbed graph, and $\mathcal{Q}$ is the set of admissible perturbed graphs.
Note that the goal of threat model/adversary is to minimize worst-case margin (Eq.~\ref{eq:PPNP_cert}), to misclassify the GNN. 
The perturbations satisfy both global budget and local budget, where global budget $B$ means there are at most $B$ perturbing edges, and local budget $ b_{t}$ limits node $t$ to have no more than $b_t$ perturbing edges. 
Node $t$ is certifiably robust when: 
\begin{equation}
	m_{y_{t}, k_t}(t,\tilde{G}^{*})=\min _{k \neq y_{t}} m_{y_{t}, k}(t,\tilde{G}^{*})>0,
	\label{eq:min w-margin}
\end{equation}
where $k_t$ is the most likely class among other classes.
%where the logits $\mathbf{H}$ and the set $\mathcal{Q}$. 
In other words, whether a node is robust is determined by the worst-case margin against any admissible perturbed graph. 
This certification method directly measures the robustness of node/graph under GNNs. 
Bojchevski \textit{et al.}~\cite{Bojchevski2019CertifiableRT} use policy iteration with reward $\mathbf{r}=-\left(\mathbf{H}_{:, y_t}\right.\left.-\mathbf{H}_{:, k}\right)$ to find the worst-case perturbed graph. 
%Under certain set of admissible perturbed graph $\mathcal{Q}_\mathcal{F}$, running policy iteration $K \times (K-1)$ times can obtain the certificates for all $N$ nodes.
%For each pair of classes $k_1$ and $k_2$, the reward vector $\mathbf{r}=-\left(\mathbf{H}_{:, k_{1}}-\mathbf{H}_{:, k_{2}}\right)$ are different. 
%The exact worst-case margins $m_{y_{t}, k}^{*}(\cdot)$ for all $N$ nodes of graph can be recovered by computing $\mathbf{\Pi}$ on the resulting $K \times (K-1)$ perturbed graphs $\tilde{G}$. 

% needed in second column of first page if using \IEEEpubid
%\IEEEpubidadjcol

%\section{Adversarial Immunization}
%In this section, we first formalize the problem of graph adversarial immunization and elaborate it with a widely used GNN model. 
%Then we propose an effective algorithm AdvImmune-Edge, using meta-gradient for selecting and immunizing appropriate node pairs in advance, to improve the certifiable robustness of graph.

\section{Adversarial Immunization Formulation}
In this section, we formalize the problem of graph adversarial immunization.
Adversarial immunization aims to improve the certifiable robustness of nodes against any admissible attack, i.e., the minimal \textit{worst-case} margin of nodes under node classification task. Specifically, by vaccinating appropriate node pairs or nodes in advance, GNN model can correctly classify nodes even under the worst case. The general goal of adversarial immunization is formalized as:
%\begin{equation}
%\max _{\mathcal{E}_{\mathbf{c}}\in \mathcal{S}_{\mathbf{c}}}  \min _{k \neq y_{t}}  m_{y_{t}, k}(t, \hat{G}),
%%=\max _{\mathcal{E}_{\mathbf{c}} \in \mathcal{S}_{\mathbf{c}}} \min _{k \neq y_{t}} \mathbf{\pi}_{\hat{G}}\left(\mathbf{e}_{t}\right)\left(\mathbf{H}_{:, y_{t}}-\mathbf{H}_{:, k}\right),
%\label{eq:node_goal}
%\end{equation}
\begin{equation}
\begin{aligned}
&\max _{\mathbf{M}\in \mathcal{M}}  \min _{k \neq y_{t}} \min _{ \tilde{G}\in \mathcal{Q}_\mathbf{M}} m_{y_{t}, k}(t, \hat{G}) \\
&\mathbf{A}_{\hat{G}} = \mathbf{A} + \mathbf{A}^{\prime}_{\tilde{G}} \circ \mathbf{M} ,
\label{eq:node_goal}
\end{aligned}
\end{equation}
where $\mathbf{M}\in \{0,1\}^{N \times N}$ is the immune graph,  $\boldsymbol{A}_{\hat{G}}$ is the adjacency matrix of modified graph $\hat{G}$ with the contribution of both perturbing graph $\boldsymbol{A}^{\prime}_{\tilde{G}}$ and immune graph $\mathbf{M}$,
and $\circ$ is the Hadamard product indicating element-wise multiplication.
$\mathcal{Q}_\mathbf{M}$ is the admissible perturbed graph set corresponding to the current immune graph $\mathbf{M}$, i.e., the perturbations involving the immune nodes or node pairs are removed from the original $\mathcal{Q}$.
In immune graph $\mathbf{M}$, an element=$0$ indicates that the corresponding node pair is immunized, which filters the influence of perturbations, while an element=$1$ implies that the corresponding node pair is not immunized and may be attacked.  
In other words,  $\mathbf{M}$ can be regarded as a mask, protecting these immune node pairs from being attacked. 
Note that the assumption of adversarial immunization is that it can protect a portion of data from attacks, a reasonable and acceptable assumption given the defense role of immunization. This defense mechanism can be incorporated into the system by creating a filter that blocks connections between specific nodes, thereby safeguarding immune edges and nodes.

Notably, we use $\mathbf{M}_E$ to denote the immune graph of edge-level immunization, and $\mathbf{M}_V$ for that of node-level immunization.
In particular, $\mathbf{M}_V=\mathbf{m}\otimes \mathbf{m}$, where $\otimes$ means outer product, $ \mathbf{m} \in \{0,1\}^{N\times 1}$ represents the immune nodes, and each element of $\mathbf{m}$ indicates whether the node is immune node or not.
Due to the limited immune budget in reality, we cannot immunize all node pairs or all nodes. 
For edge-level adversarial immunization, we provide immune edge budget $D_E$ to constrain the choice of immune node pairs.
The number of immune node pairs should be no more than  $D_E$, i.e., $\#_{0}(\mathbf{M}_E) \leq D_E$, where $\#_{0}(\mathbf{M}_E)$ denotes the number of zero elements in $\mathbf{M}_E$.
The set of admissible $\mathbf{M}_E$ is defined as:
\begin{equation}
\mathcal{M}_{E} = \left\{\mathbf{M}_E | \mathbf{M}_E \in \{0,1\}^{N \times N} , \#_{0}(\mathbf{M}_E) \leq D_E\right\}.
\end{equation}
For node-level adversarial immunization, we also provide immune node budget $D_V$ to constrain the number of immune nodes.
We define the set of admissible immune nodes $\mathbf{m}$ as:
\begin{equation}
\mathcal{M}_{V} = \left\{  \mathbf{m} |\mathbf{m} \in \{0,1\}^{N} , \#_{0}(\mathbf{m}) \leq D_V\right\}.
\end{equation}
%
%Globally, the number of immune node pairs should be no more than \textit{global budget} $C$, i.e., $\left|\mathcal{E}_{\mathbf{c}}\right| \leq C$.
%For each node $t$, the number of immune node pairs cannot exceed the \textit{local budget} $c_t$, i.e., $\left|\mathcal{E}_{\mathbf{c}}^{t}\right| \leq c_t$, where $\mathcal{E}_{\mathbf{c}}^{t}$ represents the node pairs associated with node $t$. 
%The set of admissible immune node pairs $\mathcal{E}_{\mathbf{c}}$ is defined as:
%\begin{equation}
%\mathcal{S}_{\mathbf{c}} = \left\{ \mathcal{E}_{\mathbf{c}}| \mathcal{E}_{\mathbf{c}} \subseteq (\mathcal{V} \times \mathcal{V}),
%\left|\mathcal{E}_{\mathbf{c}}\right| \leq C, 
%\left|\mathcal{E}_{\mathbf{c}}^{t}\right|\leq c_t,\forall t \in \mathcal{V} \right\}.
%\end{equation}

The above immunization objective function (Ep.~\ref{eq:node_goal}) is for a single node $t$. 
To improve certifiable robustness of the entire graph, we take the sum of the worst-case margins over all nodes as overall objective of adversarial immunization:
\begin{equation}
%	\max _{\mathcal{E}_{\mathbf{c}}\in \mathcal{S}_{\mathbf{c}}} \sum _{t \in \mathcal{V} }  \min _{k \neq y_{t}}  \mathbf{\pi}_{\hat{G}}\left(\mathbf{e}_{t}\right)\left(\mathbf{H}_{:, y_{t}}-\mathbf{H}_{:, k}\right).
\max _{\mathbf{M}\in \mathcal{M}}J(\mathbf{M}) = \max _{\mathbf{M}\in \mathcal{M}}  \sum _{t \in \mathcal{V} }  \min _{k \neq y_{t}} \min _{ \tilde{G}\in \mathcal{Q}_\mathbf{M}} m_{y_{t}, k}(t, \hat{G}),
	\label{eq:goal}
\end{equation}
where $J(\mathbf{M})$ denotes the objective fuction, $\mathcal{V}$ is the node set of graph $G$.

\textbf{Challenges:} 
Obtaining the optimal immune node pairs or nodes presents challenges due to two primary issues. 
Firstly, the computational expense of selecting a set of specific node pairs or nodes from the total pool is high. Given an immune budget of $D$, potential immunization plans are {\small$\left(\begin{array}{c}N^{2} \\ D\end{array}\right)$} for edge-level and {\small$\left(\begin{array}{c}N \\ D\end{array}\right)$} for node-level. 
This results in a prohibitive search cost of $\mathcal{O}\left(N^{2 D}\right)$ or $\mathcal{O}\left(N^{D}\right)$, making the efficient identification of optimal immune node pairs or nodes challenging. 
Secondly, the discrete nature of graph data poses an additional challenge. The resulting non-differentiability obstructs the back-propagation of gradients, thereby hindering the optimization of immune node pairs.

%It's not easy to obtain the optimal immune node pairs or nodes due to two issues.
%First, the computational cost of selecting the set of certain node pairs or nodes from totality is expensive. Given an immune budget $D$, possible immunization plans are $\left(\begin{array}{c}N^{2} \\ D\end{array}\right)$ for edge-level and $\left(\begin{array}{c}N \\ D\end{array}\right)$ for node-level. 
%This leads to an unbearable search cost $\mathcal{O}\left(N^{2 D}\right)$ or $\mathcal{O}\left(N^{D}\right)$, making it difficult to find the optimal immune node pairs or nodes efficiently. 
%The second challenge comes from the discrete nature of graph data. The resulting non-differentiability hinders back-propagating gradients to guide the optimization of immune node pairs. 

\section{Edge-Level Adversarial Immunization}
In this section, we propose an effective algorithm AdvImmune-Edge, which utilizes meta-gradient to select and immunize suitable node pairs in advance, thereby enhancing the certifiable robustness of the graph.

\subsection{AdvImmune-Edge Method}
\label{sec:AdvImmune-Edge}
The objective function of edge-level immunization is:
\begin{equation}
\small
\begin{aligned}
\max _{\mathbf{M}_E\in \mathcal{M}_E} J(\mathbf{M}_E) &= \max _{\mathbf{M}_E\in \mathcal{M}_E} \sum _{t \in \mathcal{V} }  \min _{k \neq y_{t}} \min _{ \tilde{G}\in \mathcal{Q}_{\mathbf{M}_E}} m_{y_{t}, k}(t, \hat{G}) \\
\mathbf{A}_{\hat{G}} &= \mathbf{A}+\mathbf{A}^{\prime}_{\tilde{G}} \circ \mathbf{M}_E.
\label{eq:edge_obj}
\end{aligned}
\end{equation}
We innovatively tackle it by greedy algorithm via meta-gradient to obtain the optimal immune graph matrix.

\textbf{Meta-gradient of immune graph matrix.}
Meta-gradient is the gradient of hyperparameters~\cite{Finn2017ModelAgnosticMF, zugner_adversarial_2019}. Regarding immune graph $\mathbf{M}_E$ as a hyperparameter, we compute the meta-gradient of $\mathbf{M}_E$ using the objective function:
\begin{equation}
\nabla_{\mathbf{M}_E}^{\mathrm{meta}}
=\nabla_{\mathbf{M}_E}
\left[\sum _{t \in \mathcal{V}} \min_{k \neq y_{t}} \min _{ \tilde{G}\in \mathcal{Q}_{\mathbf{M}_E}} m_{y_{t}, k}(t, \hat{G})\right],
\label{eq:meta}
\end{equation}
where $\nabla_{\mathbf{M}_E}^{\mathrm{meta}}$ is the meta-gradient of immune graph matrix $\mathbf{M}_E$.
Each entry in $\nabla_{\mathbf{M}_E}^{\mathrm{meta}}$ signifies the impact of the corresponding node pair on the objective function, i.e., worst-case margin.  
Before computing meta-gradient, we first determine the most likely class $k_t$ which minimizes the \textit{worst-case} margin of each node $t$: 
\begin{equation}
	k_t = \underset{k\neq y_t }{\arg \min } \ m_{y_{t}, k}(t, \hat{G}).
	\label{eq:k}
\end{equation}
We also compute perturbed graph that minimizes the \textit{worst-case} margin with the current ${\mathbf{M}_E}$, according to 
$
\tilde{G}^{*} = \arg \min_{\tilde{G}\in \mathcal{Q}_{\mathbf{M}_E}} \ m_{ y_{t}, k}(t, \tilde{G})
$.
And $\tilde{G}^{*}$ is updated every immunizing a certain number of node pairs.

Performing gradient optimization directly may result in decimals and elements greater than 1 or smaller than 0 in immune graph matrix, making $\mathbf{M}_E$ no longer a matrix indicating the immune choice of node pairs in graph. To address this issue, we optimize the immune graph matrix discretely using greedy algorithm.

\textbf{Meta-gradient greedy selection.}
The objective of immunization is a maximization problem, which we address using discrete gradient ascent. Specifically, we compute the meta-gradient for each entry of the matrix $\mathbf{M}_E$, and select node pairs with the greatest impact greedily. 
To maximize objective $J$, it is necessary to have a positive $\Delta J$, meaning $\Delta J > 0$.
The variation in the objective function is determined by $\Delta J= \Delta \mathbf{M}_E \cdot \nabla_{\mathbf{M}_E}^{\mathrm{meta}}$. 
Considering that $\mathbf{M}_E$ is initialized with all elements equal to 1 and can only change from 1 to 0, it implies that $\Delta \mathbf{M}_E  < 0$. As a result, $\nabla_{\mathbf{M}_E}^{\mathrm{meta}} < 0$, signifying that only the negative gradient contributes to the maximization of the objective function. The element with the highest absolute value in the negative gradient corresponds to the direction where the loss increases most rapidly~\cite{de2011second}.
Therefore, we define the inverse of meta-gradient as the value of corresponding node pair as
$
	V_{(i,j)} = -{\nabla_{\mathbf{M}_E}^{\mathrm{meta}}}_{(i,j)}
$, which signifies the impact of the corresponding node pair on the adversarial immunization objective.

 In order to preserve node pairs with the greatest impact, we greedily select entries with the maximum value in $V$:
 \begin{equation}
 	 (i^*,j^*) =\underset{(i,j)}{\arg \max } \  V_{(i,j)},
 \end{equation}
 and set corresponding entries in $\mathbf{M}_E$ to zero to immunize node pairs, protecting them from being attacked.
%The discrete optimization solution is very suitable for processing discrete graph data.

\begin{algorithm}[t]
\caption{AdvImmune-Edge immunization on graphs}
\label{alg:A}
\begin{algorithmic}[1]
	\STATE \textbf{Input:} Graph $G=(\mathbf{A}, \mathbf{X})$, immune edge budget $D_E$
	\STATE \textbf{Output:} Immune graph matrix $\mathbf{M}_E$
	\STATE $\mathbf{M}_E = ones(N,N)$ initialization
	\STATE \textbf{while} number of immune node pairs in $\mathbf{M}_E$  $< D_E$  \textbf{do}
	\STATE \quad $\tilde{G}^{*} \leftarrow$ update attack every $D_E/c$ edges immunized
	\STATE \quad $k_t \leftarrow $ the class which minimizes the worst-case margin of each node $t$ as Eq.~\ref{eq:k}
	\STATE \quad $\hat{G} \leftarrow$ perturbing with $\tilde{G}$  and immunizing with $\mathbf{M}_E$
%	\STATE \quad $\mathbf{\pi}_{\hat{G}} \leftarrow $ concatenate $\mathbf{\pi}_{\hat{G}}(\mathbf{e}_t)$ of corresponding node $t$.
	\STATE \quad $\nabla_{\mathbf{M}_E}^{\mathrm{meta}}
\leftarrow $ the meta-gradient computing in Eq.~\ref{eq:meta}
	\STATE \quad $ V \leftarrow -\nabla_{\mathbf{M}_E}^{\mathrm{meta}}$ 
	\STATE \quad $V_{(i,j)}=0, \ \forall{(i,j) \in \mathbf{M}_E}$ exclude node pairs that have already been immunized.
	\STATE \quad $(i^*,j^*)\leftarrow $ maximal entry in $V$
	\STATE \quad \small $\mathbf{M}_E[i^*,j^*] = 0 \leftarrow $ immunize one more node pair $(i^*,j^*)$
	\STATE \textbf{end while}
	\STATE \textbf{Return: } $\mathbf{M}_E$
\end{algorithmic}
\end{algorithm}

\subsection{Algorithm}
This section elucidates AdvImmune-Edge algorithm, which obtains immune node pairs to improve the certifiable robustness of graph (Algorithm \ref{alg:A}). 
Initially, $\mathbf{M}_E$ is set as a matrix with all elements as 1, indicating no immunized node pair.
The algorithm then iteratively selects node pairs with the highest impact. 
In each iteration, we choose the most likely class $k_t$ minimizing the worst-case margin. 
The core step is to calculate the meta-gradient of objective function for  $\mathbf{M}_E$ and the corresponding value $V_{(i,j)}$.
Node pairs that have already been immunized (in $M_E$) are excluded to avoid repetition.
The maximum entry in $V_{(i,j)}$ is selected and the corresponding entry in $\mathbf{M}_E$ is set to zero.  
The process repeats until sufficient immune node pairs are acquired.

\begin{figure}[t]
	\centering
	\includegraphics[width=6.9cm]{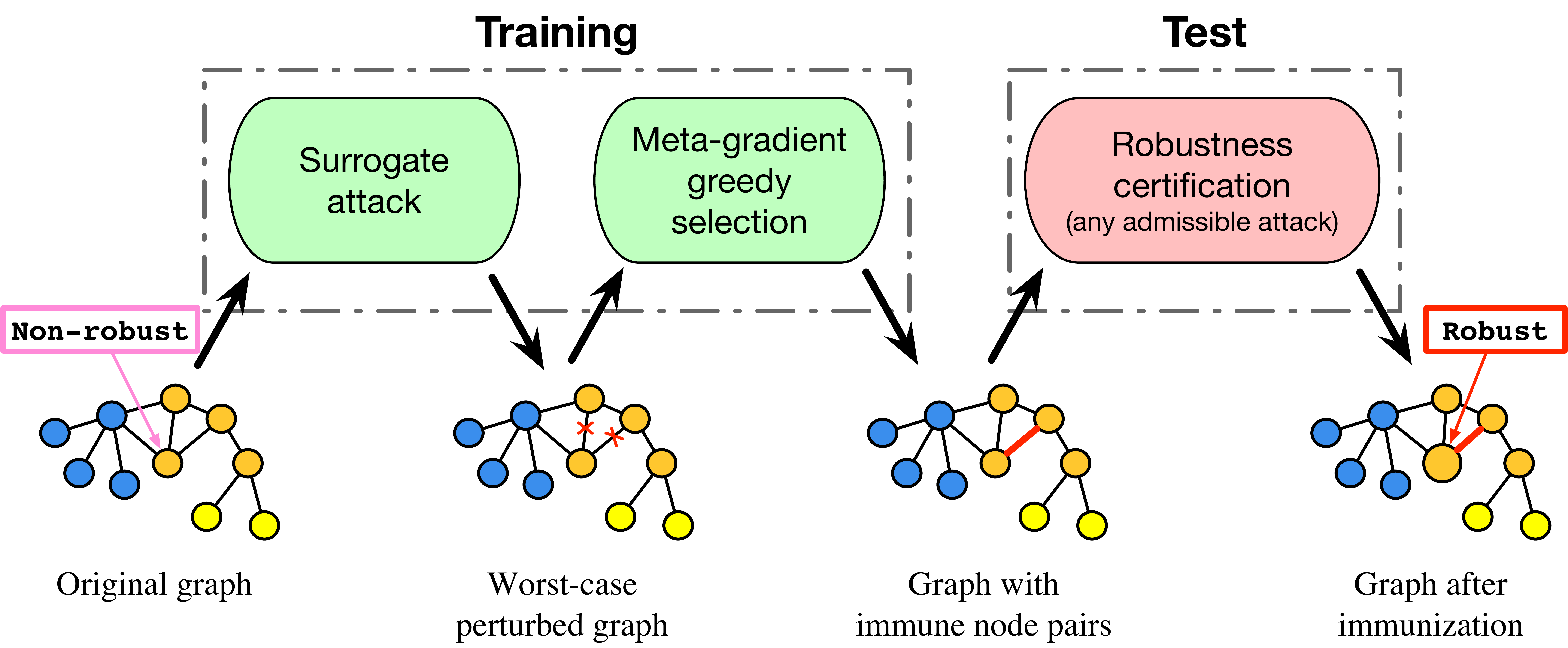}
	\centering
	\caption{The training and test process of the AdvImmune-Edge algorithm. }
	\label{fig:process}
\end{figure}

Figure~\ref{fig:process} illustrates the training and test process of our AdvImmune-Edge. For the training phase, we adopt surrogate attack to generate the worst-case perturbed graph, and then we select and immunize  suitable edges or node pairs using the meta-gradient (Algorithm~\ref{alg:A}). For test, the immune graph serves as a mask, protecting certain node pairs from being modified, thereby improving the certifiable robustness of the graph against any admissible adversarial attack.

%We illustrate training and test process of AdvImmune-Edge in Figure~\ref{fig:process}.  
%During training, we first use surrogate model to obtain the worst-case perturbed graph. 
%Then, we select and immunize appropriate node pairs through Algorithm~\ref{alg:A} via meta-gradient.
%As for the test, we use the immune graph as a mask to protect certain node pairs from being modified, and improve the certifiable robustness of nodes against any admissible adversarial attack.
%It is worth noting that adversarial immunization is performed before the attack of robustness certification. 
%With some immune node pairs, nodes become acquired robust against any admissible attack.

\subsection{Complexity Analysis}
The computational complexity of AdvImmune-Edge algorithm depends on GNN model. Assuming the computational complexity of GNN model, i.e., the complexity of the surrogate attack, is $\mathcal{O}(T)$. Specifically, under $\pi$-PPNP, the personalized PageRank matrix $\mathbf{\pi}$ involves the inversion operation, resulting in a computational complexity of  $\mathcal{O}(T) = \mathcal{O}(N^{3})$. 
Calculate the minimal worst-case margins for each pair of classes $(k_1, k_2)$ results in a computational complexity of $\mathcal{O}\left( K^{2}\cdot T \right)$ in each iteration. To improve the estimate of meta-gradients, we also update the surrogate attack constant times, represented as $c$, where $c$ typically hovers around $100$. The optimization process of immunization, which includes operations of both element-wise multiplication and meta-gradient, demands a computational complexity of $\mathcal{O}\left(N^2\right)$. Considering the total of $D_E$ iterations to identify the optimal $D_V$ node pairs for immunization, AdvImmune-Edge method has a computational complexity of $\mathcal{O}\left( c \cdot K^{2}\cdot T + D_E \cdot  N^2 \right)$.

%Since we have to calculate the minimal worst-case margins for each pair of classes $(k_1, k_2)$, the computational complexity in each iteration is $\mathcal{O}\left( K^{2}\cdot T \right) $.
%For a better estimate on meta-gradients, we also update the surrogate attack constant times, denoted as $c$. Usually, $c$ is a constant around $100$ in our experiments.
%As for the optimization process of immunization, operations of both element-wise multiplication and meta-gradient require a computational complexity of $\mathcal{O}\left(N^2\right)$. 
%Considering the total of $D_E$ iterations to find the optimal $D_V$ node pairs to be immunized, AdvImmune-Edge method has a computational complexity of $\mathcal{O}\left( c \cdot K^{2}\cdot T + D_E \cdot  N^2 \right)$.

\section{Node-Level Adversarial Immunization}
In this section, we focus on node-level adversarial immunization.  We propose AdvImmune-Node method to select immune nodes effectively and efficiently, and protect them from attacks to improve the certifiable robustness of graph. We also provide the algorithm and complexity analysis.
%% \vspace{-10pt}

\subsection{AdvImmune-Node Method}
The node-level immunization vaccinates a fraction of nodes, protecting all their relationships from being modified by attacks. That is, the attacker cannot add any edges to these nodes, nor delete the existing edges of them.
We present the node-level immunization problem in matrix form:
\begin{equation}
\begin{aligned}
\max _{ \mathbf{m}\in \mathcal{M}_V} J(\mathbf{m}) &= \max _{ \mathbf{m}\in \mathcal{M}_V} \sum _{t \in \mathcal{V} } \min _{k \neq y_{t}}  \min _{ \tilde{G}\in \mathcal{Q}_\mathbf{m}} m_{y_{t}, k_{t}}(t, \hat{G})\\
%&\mathbf{\pi}_{\hat{G}}\left(\mathbf{e}_{t}\right)=(1-\alpha)\mathbf{e}_{t}\left(\mathbf{I}_{N}-\alpha \mathbf{D}^{-1}_{\hat{G}} \mathbf{A}_{\hat{G}}\right)^{-1} \\
%\mathbf{A}_{\hat{G}} =& \mathbf{A} +  \mathbf{M}_V \cdot \mathbf{A}^{\prime}_{\tilde{G}^{*}} \cdot \mathbf{M}_V,
\mathbf{A}_{\hat{G}} &= \mathbf{A} +  \mathbf{A}^{\prime}_{\tilde{G}} \circ \left(\mathbf{m}\otimes \mathbf{m}\right),
\end{aligned}
\end{equation}
where $ \mathbf{m}\otimes \mathbf{m}$ is the immune graph. In the vector $ \mathbf{m}$, $0$ represents the immune node, whose relationships will be protected, while $1$ implies the corresponding node is not immunized and may be attacked. 
Similar to Section~\ref{sec:AdvImmune-Edge}, we also adopt discrete gradient ascent, in which the element with the greatest absolute value in the negative gradient represents the direction of most rapid growth of loss~\cite{de2011second}.
%We define the inverse of meta-gradient as the value of node:
The node's value is defined as the meta-gradient inverse:
%The value of node is:
\begin{equation}
\begin{aligned}
V &= -\nabla_{ \mathbf{m}}^{\mathrm{meta}} \\
\nabla_{ \mathbf{m}}^{\mathrm{meta}}
& =\nabla_{ \mathbf{m}}
\left[\sum _{t \in \mathcal{V}} \min_{k \neq y_{t}} \min _{ \tilde{G}\in \mathcal{Q}_\mathbf{m}} m_{y_{t}, k}(t, \hat{G})\right].
\end{aligned}
\label{eq:meta-node}
\end{equation}
However, node-level adversarial immunization faces its own unique challenges that meta-gradient-based greedy algorithm cannot solve well.

\textbf{Challenges.}
As shown in Figure~\ref{fig:process}, during training, the worst-case perturbed graph used by immune node pairs is calculated from the original graph before immunization. For the test, the worst-case perturbed graph in robustness certification is calculated according to the graph after immunization. The two perturbations are different. As a result, the immune node pairs or nodes may not be optimal on the new worst-case perturbation.
Fortunately, it is tolerable for edge-level immunization since there is only several immune edges (node pairs) difference between the perturbations before and after immunization.
However, node-level immunization amplifies the inaccuracy. In other words, immunizing one node means protecting $N$ node pairs, so it will be unreliable to immunize nodes calculated only by inaccurate estimation.

To address the problem, we propose a novel method, AdvImmune-Node based on robustness gain to obtain immune nodes.

\textbf{Robustness gain.}
The challenges mainly come from the inaccurate estimation caused by different worst-case perturbed graphs before and after immunization.
It is best to select immune nodes based on the real worst-case perturbed graph after immunization.
A direct way is to calculate the gain on robustness before and after immunizing the node, namely \textit{robustness gain} $\Delta$. Specifically, assuming that one more node $j$ is immunized on the current immune graph $\mathbf{m}\otimes \mathbf{m}$, we calculate new robustness based on the new worst-case perturbed graph after immunizing $j$, and the additional robustness brought by immunizing node $j$ is denoted as $\Delta_{j}$. We formalize the robustness gain $\Delta_{j}$:

\begin{small}
\begin{equation}
\begin{aligned}
   \Delta_{j} = \sum _{t \in \mathcal{V} }  m_{y_{t}, k_{t}}(t, \hat{G}^{j}) - \sum _{t \in \mathcal{V} }  m_{y_{t}, k_{t}}(t, \hat{G}) 
 \end{aligned}
 \label{eq:gain}
\end{equation}  
\begin{equation*}
\begin{aligned}
%\mathbf{A}_{\hat{G}^{j}} & = \mathbf{A} +  \mathbf{M}_V^{j} \cdot \mathbf{A}^{\prime}_{\tilde{G}^{j^*}} \cdot \mathbf{M}_V^{j}  
\mathbf{A}_{\hat{G}^{j}} = \mathbf{A} +  \mathbf{A}^{\prime}_{\tilde{G}^{j^*}} \circ \left(\mathbf{m}^{j}\otimes \mathbf{m}^{j}\right),& \
\tilde{G}^{j^*}  = \arg\min_{\tilde{G}^{j} \in \mathcal{Q}_\mathbf{m}^{j} } m_{y_{t}, k}(t,\tilde{G}^{j})\\
\mathbf{A}_{\hat{G}}  = \mathbf{A} +  \mathbf{A}^{\prime}_{\tilde{G}^{*}} \circ \left( \mathbf{m}\otimes \mathbf{m}\right),& \
\tilde{G}^{*} = \arg\min_{\tilde{G} \in \mathcal{Q}_\mathbf{m}} m_{y_{t}, k}(t,\tilde{G}),
\end{aligned}
\end{equation*}
\end{small}
\hspace{-5pt}where the robustness gain $\Delta_{j}$ of node $j$ means the difference between new robustness on new immune graph $\mathbf{m}^{j}\otimes \mathbf{m}^{j}$ (after immunizing node $j$) and the original robustness on current immune graph $\mathbf{m}\otimes \mathbf{m}$.
$\mathbf{m}$ indicates the current immune nodes, and $\mathbf{m}^{j}$  immunizes one more node $j$ than the $\mathbf{m}$. 
$\mathcal{Q}_\mathbf{m}$ means the current admissible perturbed graph set, and $\mathcal{Q}^{j}_\mathbf{m}$ denotes new perturbed graph set after immunizing $j$, i.e., removing all perturbations involving node $j$ and its corresponding edges and node pairs. $\tilde{G}^{j^*}$ is the new worst-case perturbed graph assuming immunizing one more node $j$.
Robustness gain reflects how much the robustness of graph can be gained if we immunize node $j$.

As for practical immunization, we can vaccinate the node with the greatest robustness gain, and update gain after each immunization.
However, robustness gain requires searching for the new worst-case perturbed graph, which is time-consuming. We consider a more efficient approach.

\begin{algorithm}[tp]
\caption{AdvImmune-Node immunization on graphs}
\label{alg:N}
\begin{algorithmic}[1]
	\STATE \textbf{Input:} Graph $G=(\mathbf{A}, \mathbf{X})$, immune node budget $D_V$
	\STATE \textbf{Output:} Immune nodes $\mathbf{m}$
	\STATE  $\mathbf{m} = ones(N,1)$ initialization
	\STATE $\hat{G} \leftarrow$ graph with current $\mathbf{m}$
	\STATE $V \leftarrow $ values computed in Eq.~\ref{eq:meta-node}
	\STATE $P \leftarrow $ candidate nodes with large $V$
	\STATE $\Delta_{j} \leftarrow$ robustness gain for each node $j$ in $P$ as Eq.~\ref{eq:gain}.
	\STATE Sort $P$ by $\Delta_{j}$ from large to small
	\STATE \textbf{while} number of immune nodes in $\mathbf{m} < D_V$ \textbf{do}
	\STATE \quad \textbf{while} True \textbf{do}
	\STATE \quad \quad Select node $j_1 = \arg\max_{j \in P} (\Delta_j) $  in  $P$
	\STATE \quad \quad \small Update $\Delta_{j_1}$ for node $j_1$ as Ep.~\ref{eq:gain}
	\STATE \quad \quad $j_2 = \arg\max_{j \in P}(\Delta_{j})$  the maximum gain
	\STATE \quad \quad \textbf{if}  $j_1 == j_2 $   \textbf{do}
	\STATE \quad \quad \quad \textbf{break}
	\STATE \quad \quad \textbf{end if}
	\STATE \quad \textbf{end while}
	\STATE \quad $\mathbf{m}[j_1] = 0 \leftarrow $ immunize one more node $j_1$
	\STATE \quad Remove node $j_1$ from candidate set $P$
	\STATE \quad $\hat{G} \leftarrow$ graph with current $\mathbf{m}$
	\STATE \textbf{end while}
	\STATE \textbf{Return: } $\mathbf{M}_V$
\end{algorithmic}
\end{algorithm}

\textbf{Candidate selection.}
To narrow down the range of nodes for which robustness gains need to be calculated, we select important nodes as the candidate node set and only compute $\Delta$ for these candidate nodes. Because most nodes do not contribute much to the graph robustness. 
We select candidate nodes based on their values $V$ via Eq.~\ref{eq:meta-node}.
We take the top-k nodes with the largest $V$ as candidate node set $P$, and calculate $\Delta_{j}$ for each node $j$ in $P$.

\textbf{Lazy update.}
For the first-time immunization, we select the node with the largest robustness gain $\Delta$, and remove the node from the candidate set $P$.
As for updating $\Delta$, it is time-consuming to update for all candidates after each immunization.
For efficiency, we stop the update as long as the updated robustness gain of the current node is the largest in the current candidate set~\cite{Leskovec2007CELF}.
Specifically, we update the robustness gain for the node $j_1 = \arg\max_{j \in P} (\Delta_j)$ with the largest gain, according to  Eq.~\ref{eq:gain}. If the updated gain is still the largest in current $P$, then we select and vaccinate node $j$ in this immunization. Otherwise, we continue to update the robustness gain from large to small until the updated gain is the largest.
In this way, the gains of immune nodes are computed and updated efficiently. Further complexity analysis is discussed in Section~\ref{sec:node-comp}.

To demonstrate the reasonability of AdvImmune-Node, we conduct experiments to collect the ground-truth immune nodes by updating the robustness gains of all nodes.
Here, the immune budget is 0.5$\%$ nodes, i.e., 11 nodes.
We observe that the ground-truth immune nodes are exactly the same as the immune nodes by AdvImmune-Node, which only updates robustness gain for a few nodes.

\subsection{Algorithm}
We describe our AdvImmune-Node algorithm in Algorithm~\ref{alg:N}. We first initialize $\mathbf{m}$  as a vector with all elements of 1, indicating no node is immunized. Then we select the candidate nodes $P$ according to the node value $V$, and compute the robustness gain $\Delta$ for each candidate. After that, we choose node $j_1$ with the highest gain $\Delta$ based on previously computed results (Step 11). Next, we update the gain $\Delta_{j_1}$ for $j_1$ by setting $j$ in Eq.~\ref{eq:gain} to $j_1$ (Step 12). Subsequently, we select node $j_2$ with the maximum gain from the updated $\Delta$. If $j_1$ equals $j_2$, it indicates that the maximum updated gain is larger than the gain of other nodes prior to the update, thus confirming that $j_1$ yields the highest gain.
% we update $\Delta$ from large to small until there is no greater gain than the current updated gain $\Delta_{j_1}$. 
 We immunize the current node $j_1$ and remove it from the candidate set $P$. This process is repeated until we obtain enough immune nodes.

\subsection{Complexity Analysis}
\label{sec:node-comp}
Similar to AdvImmune-Edge, AdvImmune-Node also computes the worst-case margin with a computational complexity of  $\mathcal{O}\left( K^{2}\cdot T \right) $.
As for node-level immunization, we need $D_V$ immune nodes, and robustness gains may be updated multiple times, denoted as $d$. 
Usually, $d$ is a very small constant. In our experiments,  $d$ is usually 1 or 2, rarely 3 or 4. Therefore, robustness gain computation performs $d \cdot D_V$ times.
The robustness gain $\Delta$ has a demand for computing the worst-case margin.
%The computational complexity of selecting and sorting candidates is $\mathcal{O}\left( N \cdot \log N\right)$
In total, AdvImmune-Node has a computational complexity of $\mathcal{O}\left( d \cdot D_V \cdot (K^{2}\cdot T) \right) $.

Note that such a robustness gain-based algorithm is not suitable for edge-level immunization.
%the magnitude of node pairs is much larger than that of nodes, 
%AdvImmune-Node has to update worst-case perturbation more than once before each immunization, which is too expensive.
This is because $D_E$ in edge-level immunization is generally much larger than $D_V$ in node-level immunization, resulting in unaffordable computational cost with $\mathcal{O}(d\cdot D_E \cdot (K^2\cdot T))$.

%\begin{table}[t]
%  \caption{Statistics of the evaluation datasets}
%  \label{tab:dataset}
%  \begin{tabular}{ccccccc}
%    \toprule
%    Dataset & Type & $N_{LCC}$ & $\left|\mathcal{E}_{LCC}\right|$  & $d$ & $K$ \\
%    \midrule
%    Citeseer & Citation network & 2110 & 3668 & 3703 & 6\\
%    Cora-ML & Citation network & 2810 & 7981  & 2879 & 7\\
%    Reddit & Social network & 3069 & 7009 & 602 & 5 \\
%  \bottomrule
%\end{tabular}
%\end{table}

\begin{figure*}[!h]
\centering
\subfigure[Ratio of robust nodes on Citeseer]{
\includegraphics[width=5.8cm]{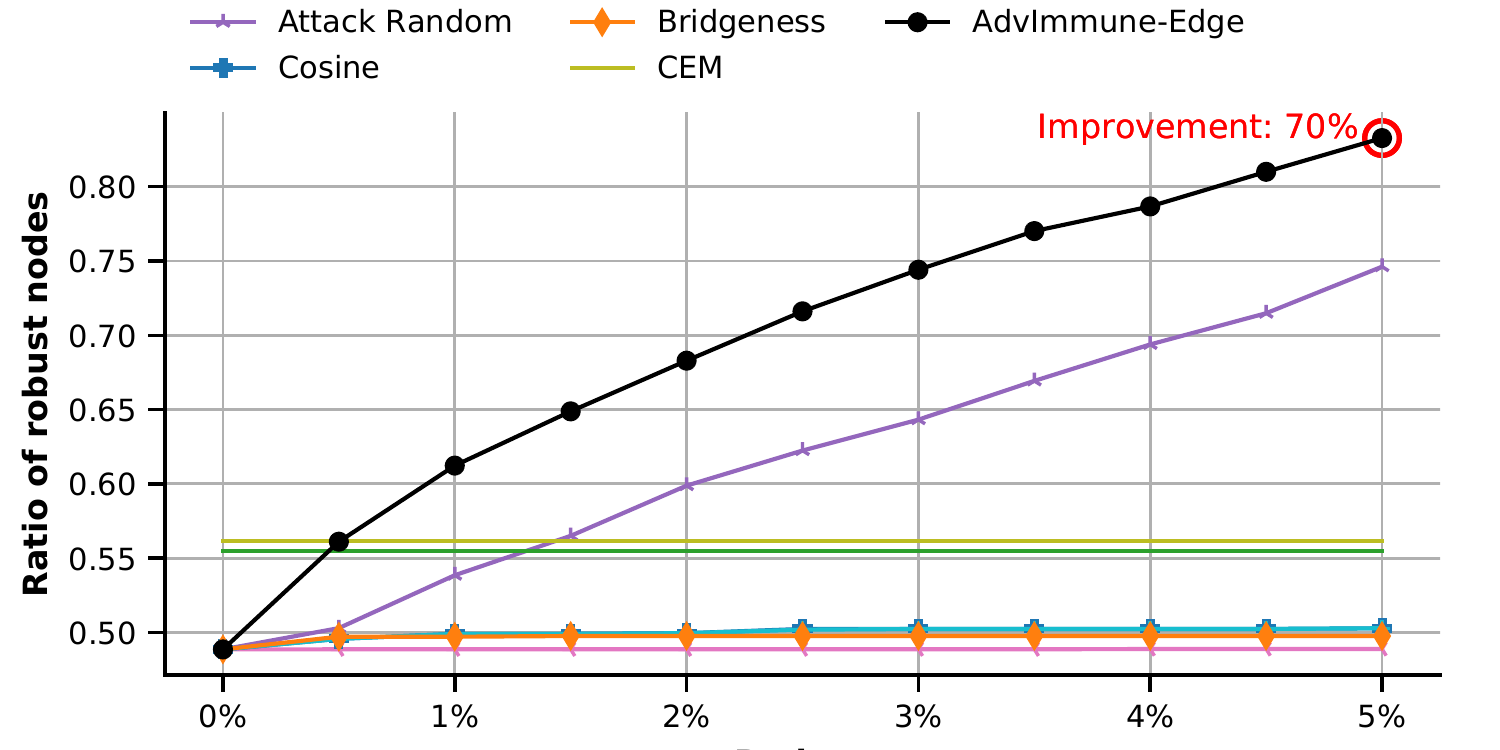}
\label{subfig:citeseer_r}
}
\subfigure[Ratio of robust nodes on Cora-ML]{
\includegraphics[width=5.8cm]{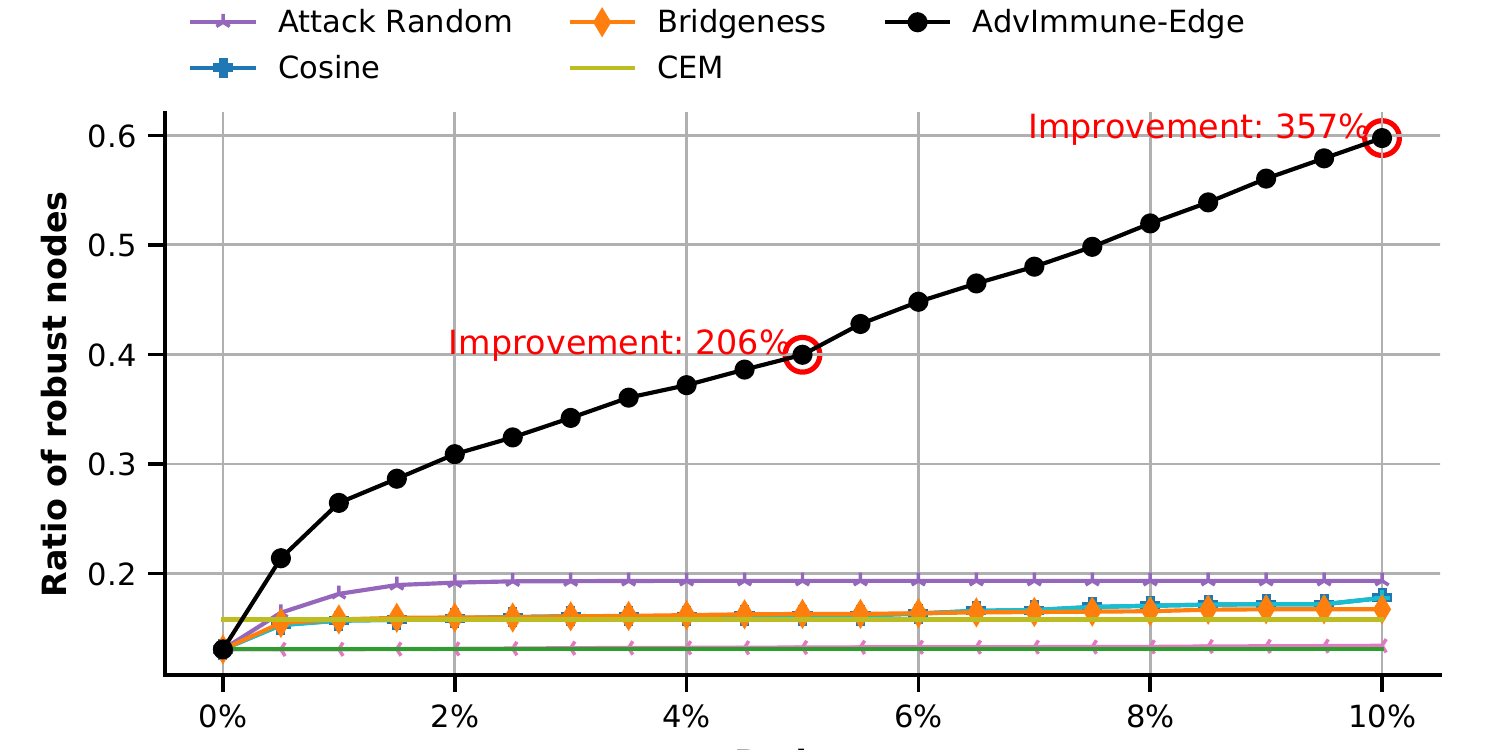}
\label{subfig:cora_r}
}
\subfigure[Ratio of robust nodes on Reddit]{
\includegraphics[width=5.8cm]{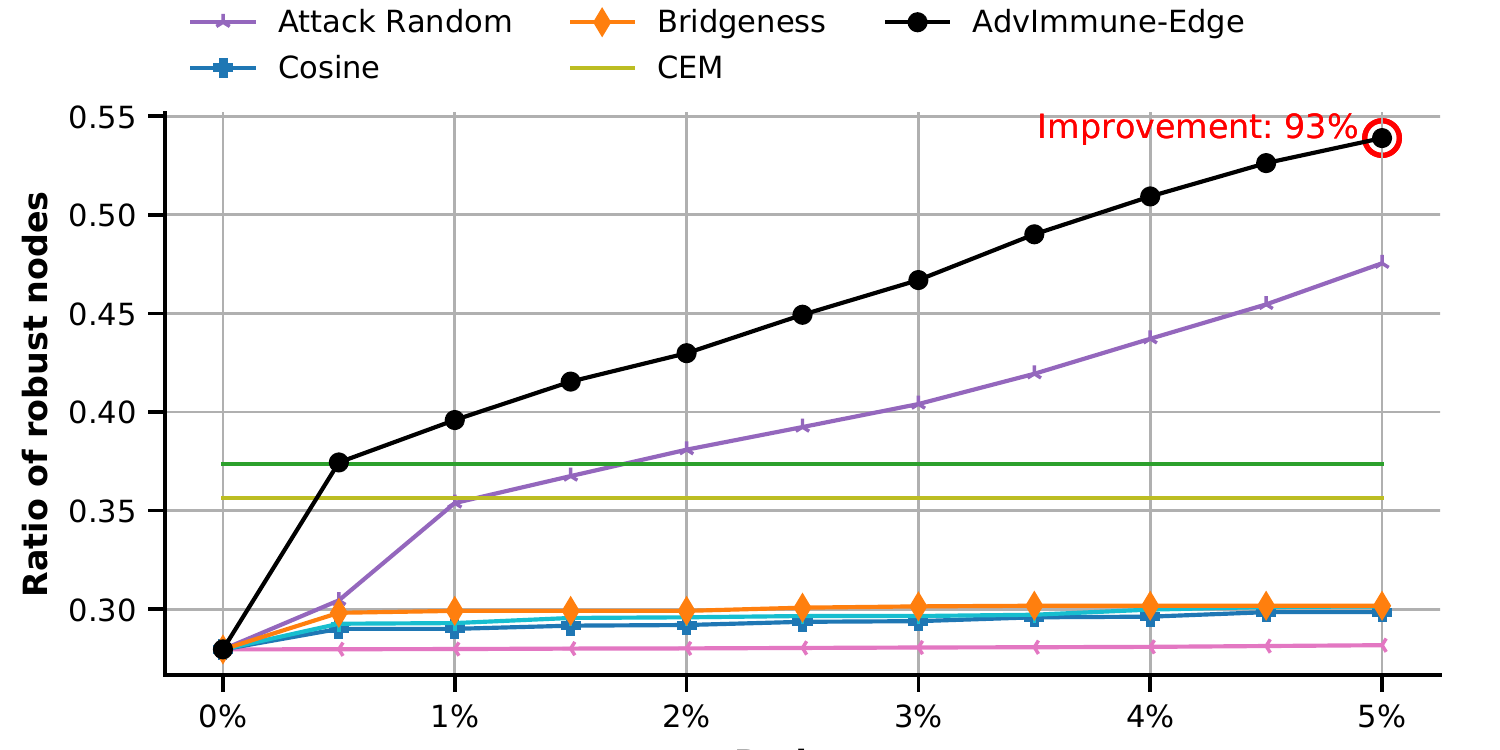}
\label{subfig:reddit_r}
}
\subfigure[Worst-case margin on Citeseer]{
\includegraphics[width=5.8cm]{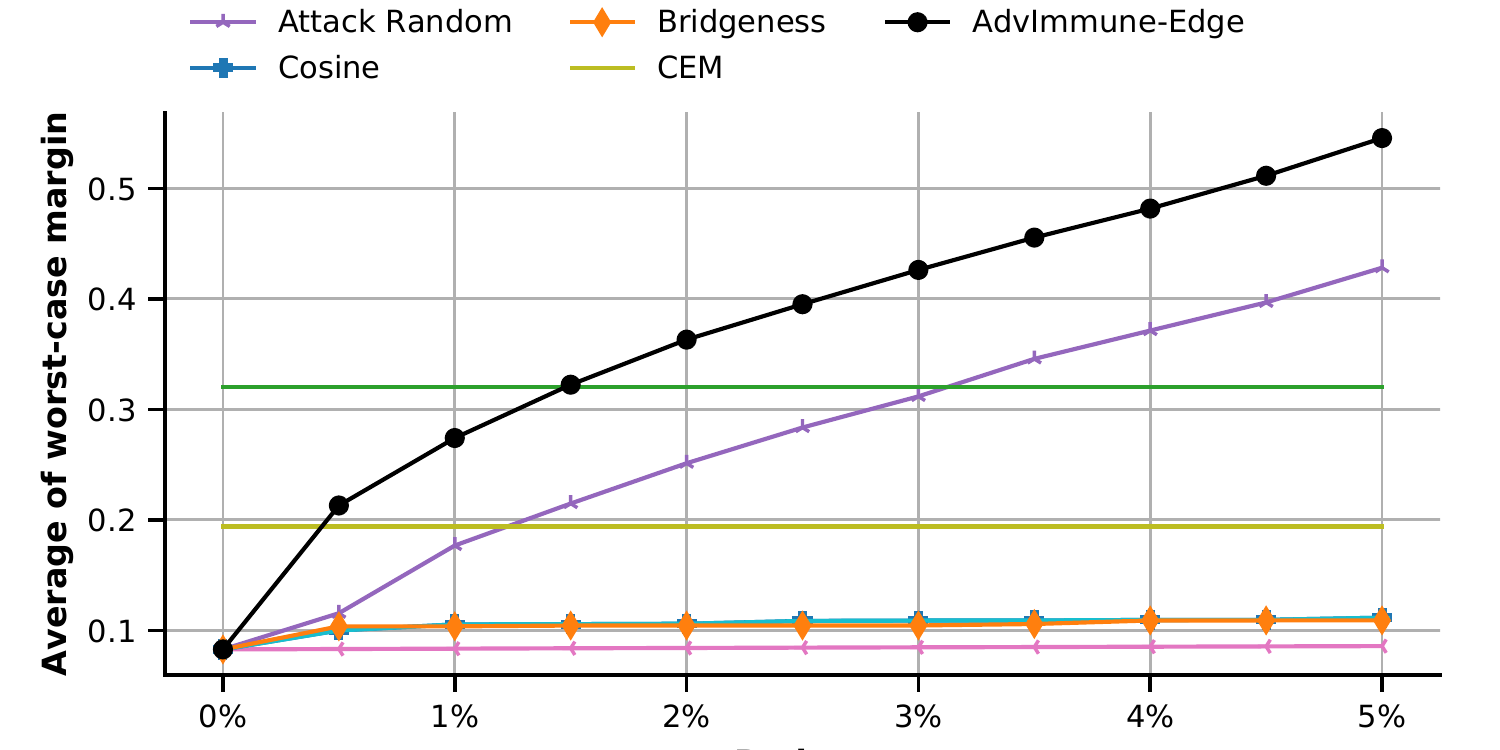}
\label{subfig:citeseer_w}
}
\subfigure[Worst-case margin on Cora-ML]{
\includegraphics[width=5.8cm]{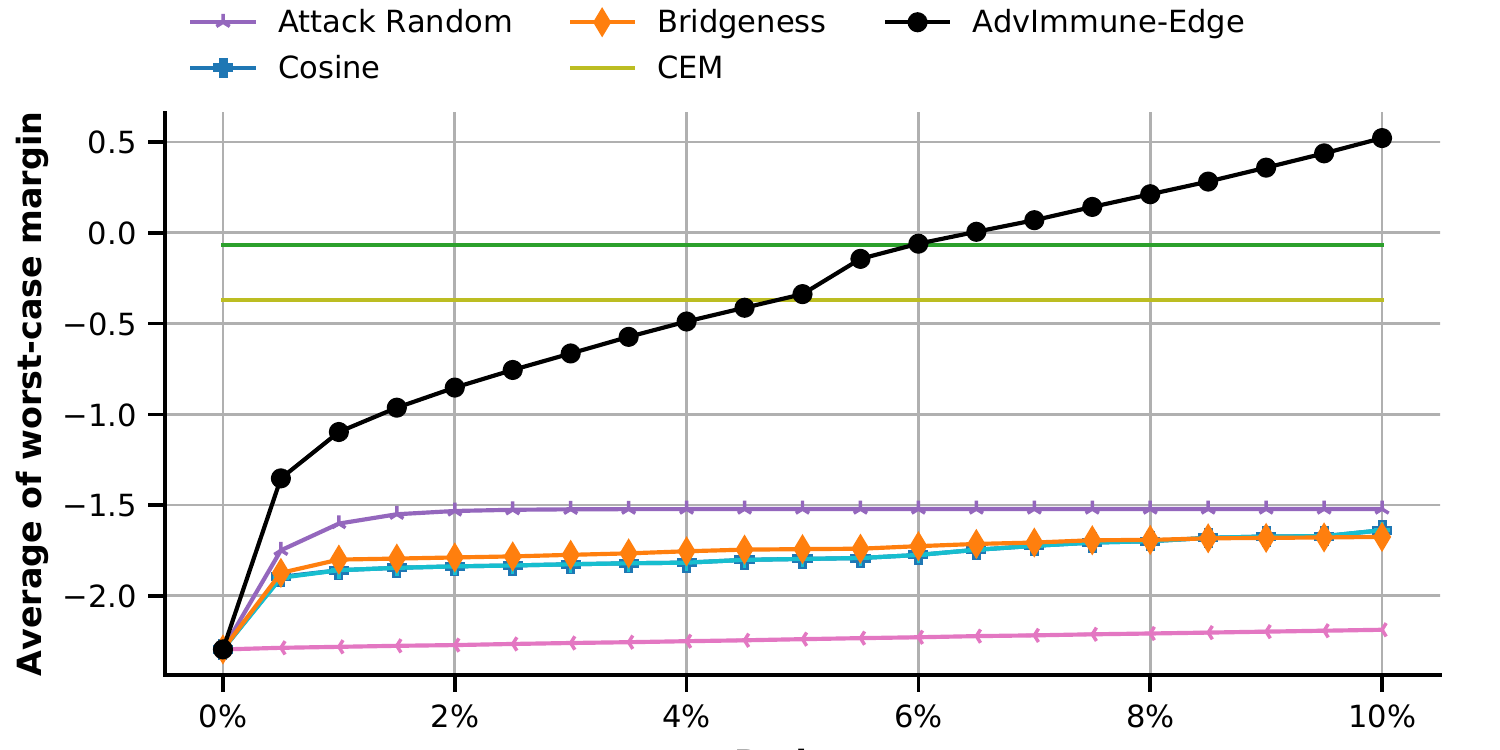}
\label{subfig:cora_w}
}
\subfigure[Worst-case margin on Reddit]{
\includegraphics[width=5.8cm]{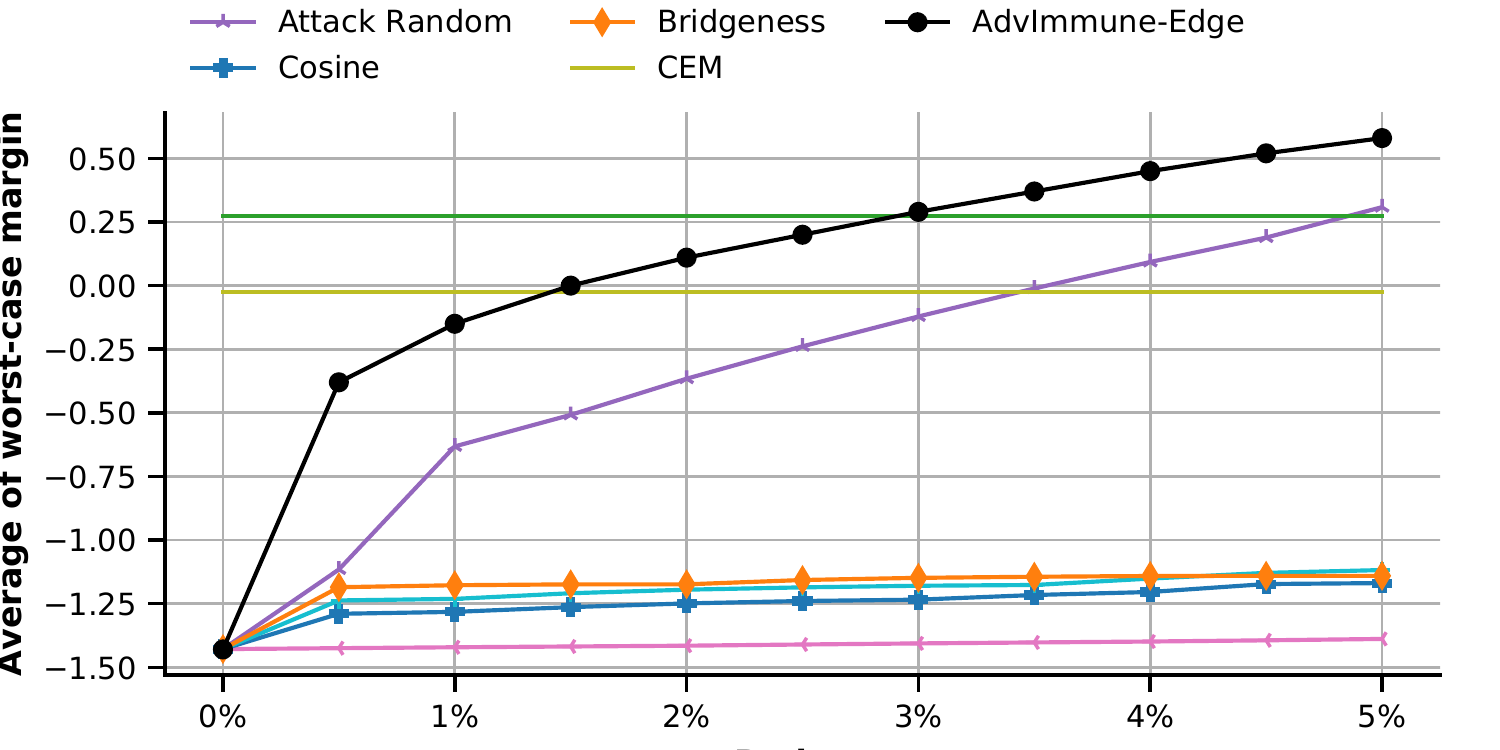}
\label{subfig:reddit_w}
}
\caption{
The effectiveness of edge-level adversarial immunization on three datasets. The upper figures (a-c) illustrate the changes in the ratio of robust nodes, whereas the lower figures (d-f) depict the changes in the average of worst-case margins.
}
\label{fig:edge}
\end{figure*}

\section{Experiments}
In this section, we demonstrate the robustness superiority of AdvImmune methods on three benchmark datasets, and verify the practical defense effectiveness against five attack methods, compared to five defense methods. Further case studies of immune nodes and node pairs are also provided.
\subsection{Experimental Datasets and Settings}
\subsubsection{Datasets}
AdvImmune methods are assessed on two commonly used citation networks: Citeseer (2110 nodes, 3668 edges, 6 classes) and Cora-ML(2810 nodes, 7981 edges, 7 classes)~\cite{Bojchevski2019CertifiableRT}, and a social network Reddit (3069 nodes, 7009 edges, 5 classes)~\cite{Zeng2019GraphSAINTGS}. 
In the citation networks, nodes symbolize papers with keywords as attributes and paper class as the label, while edges symbolize citation relationships. For Reddit, each node symbolizes a post with word vectors as attributes and community as the label, and each edge symbolizes the relationship between posts.

%In citation networks, a node represents a paper with keywords as attributes and paper class as label, and the edge represents the citation relationship.  
%In Reddit, each node represents a post, with word vectors as attributes and community as label, while each edge represents the post-to-post relationship.

%Due to the high complexity of $\pi$-PPNP, it is difficult to apply it to large graphs, leading to the limitation of our method.
%Therefore, we only keep a subgraph of Reddit to conduct experiments. 
%Specifically, we first randomly select 10,000 nodes, and four classes with the most nodes are selected as our target classes.
%All nodes in these target classes are kept.
%Then, to retain the network structure as much as possible, we further include the first-order neighbors of the kept nodes.
%Nodes that don't belong to the target four classes are marked as the fifth class, i.e., other-class.
%Experiments are conducted on the largest connected component (LCC) for three benchmark datasets~\cite{zugner2018adversarial}. 
%The statistics of each dataset are summarized in Table \ref{tab:dataset}. 

\subsubsection{Settings}
\label{subsec:scen}
Our experiments utilize the same GNN settings as in ~\cite{Bojchevski2019CertifiableRT}, where the $\pi$-PPNP model with a transition probability $\alpha=0.85$. 
The threat model~\cite{Trustworthy2022} is evasion white-box attack, following certifiable robustness~\cite{Bojchevski2019CertifiableRT}, since the white-box attack is the strongest attack, better evaluating the ability of defense methods. The attacker knows the input dataset, output prediction, as well as the structure and parameters of the GNN model. 
For the surrogate attack, we adopt the challenging and practical Remove-Add scenario~\cite{Bojchevski2019CertifiableRT}, where the attacker can add or remove the connection of any node pair. Both the surrogate model and robustness certification limit the local budget to $b_t=\max\left(\mathbf{D}_t-6, 0\right)$, where $\mathbf{D}_t$ is the degree of node $t$. We do not impose a global budget limit in both scenarios to better demonstrate our effectiveness. The robustness of nodes is measured using prediction classes, following the settings of robustness certification~\cite{Bojchevski2019CertifiableRT,Zgner2019CertifiableRA}.

For ours, we conduct experiments on node-level and edge-level immunization. 
Our immune budget ranges from $0.5\%$ to $5\%$ of node pairs or nodes in edge-level or node-level immunization. 
We also explore the potential of edge-level immunization on Cora-ML when the immune budget is 10$\%$ node pairs.
%Note that the immune object could be edge or node pair in edge-level immunization, we unify them as immune edges.
%For baselines, random-based methods randomize 10 times in both scenarios. As for attribute-based methods, in \textbf{Remove-only} scenario, they only immunize the connected edges with high attribute similarity between nodes under the same class, while in \textbf{Remove-Add} scenario, they immunize connected edges with a probability of 0.3 and unconnected node pairs with a probability of 0.7. Such settings of probability is based on the fact that there are about 30\% removed edges among the worst-case perturbed edges in all datasets. 

\subsection{Baselines}
\label{sec:baseline}
To our knowledge, we pioneer the proposal of graph adversarial immunization. To validate the efficacy of AdvImmune methods, we design random and heuristic immunization baselines for both edge-level and node-level immunization.

%Besides, we also compare AdvImmunes with state-of-the-art defense methods.
            
\subsubsection{Edge-level immunization baselines}
\quad
$\bullet$ \textbf{Random immunization methods.} We implement two random methods with distinct sets of candidate immune node pairs. \textit{Random} method selects immune node pairs randomly from all node pairs, while \textit{Attack Random} chooses immune node pairs from the worst-case perturbed edges obtained by the surrogate attack. Attack Random, having knowledge of the worst-case perturbed edges from surrogate attacks, is stronger than Random baseline.

$\bullet$ \textbf{Heuristic immunization methods.} 
\emph{(1) Attribute-based Methods.}
Studies indicate that attackers often remove edges between nodes of the same class while adding edges to pairs of nodes from different classes~\cite{zugner_adversarial_2019,Wu2019AdversarialEF}. Thus, we design baselines that preserve the connection of edges with high attribute similarity within the same class and disconnect node pairs with low attribute similarity across different classes. We utilize commonly used measures of similarity, specifically \textit{Jaccard} and \textit{Cosine}.
\emph{(2) Structure-based Methods.} We also devise baselines considering the structural significance of edges. We employ Jaccard similarity between the neighbors of a node pair to reflect the connectivity and importance of edges. We then heuristically select the edges with the highest values and the unconnected node pairs with the lowest values for immunization, namely \textit{Bridgeness}.

% \emph{(1) Attribute-based methods.}
%Researches show that attackers tend to remove edges between nodes from the same class and add edges to pairs of nodes from different classes~\cite{zugner_adversarial_2019,Wu2019AdversarialEF}. Therefore, we design baselines that maintain the connection of edges with the high similarity between the attributes of nodes under the same class and the disconnection of node pairs with low attribute similarity under different classes. We consider the commonly used Jaccard score and cosine score as the measures of similarity, namely \textit{Jaccard} and \textit{Cosine}, respectively. 
%\emph{(2) Structure-based methods.} We also design baselines considering the structural importance of edges. 
%%Specifically, for global structure importance, we use edge-betweenness as a measurement~\cite{Girvan2002CommunitySI}.
%%Locally, 
%We adopt Jaccard similarity between the neighbors of node pair, to reflect the connectivity and importance of edges, and we heuristically choose the edges with the greatest values and the unconnected node pairs with the smallest values to be immunized, namely \textit{Bridgeness}.

\begin{figure*}[ht]
\centering
\subfigure[Ratio of robust nodes on Citeseer]{
\includegraphics[width=5.8cm]{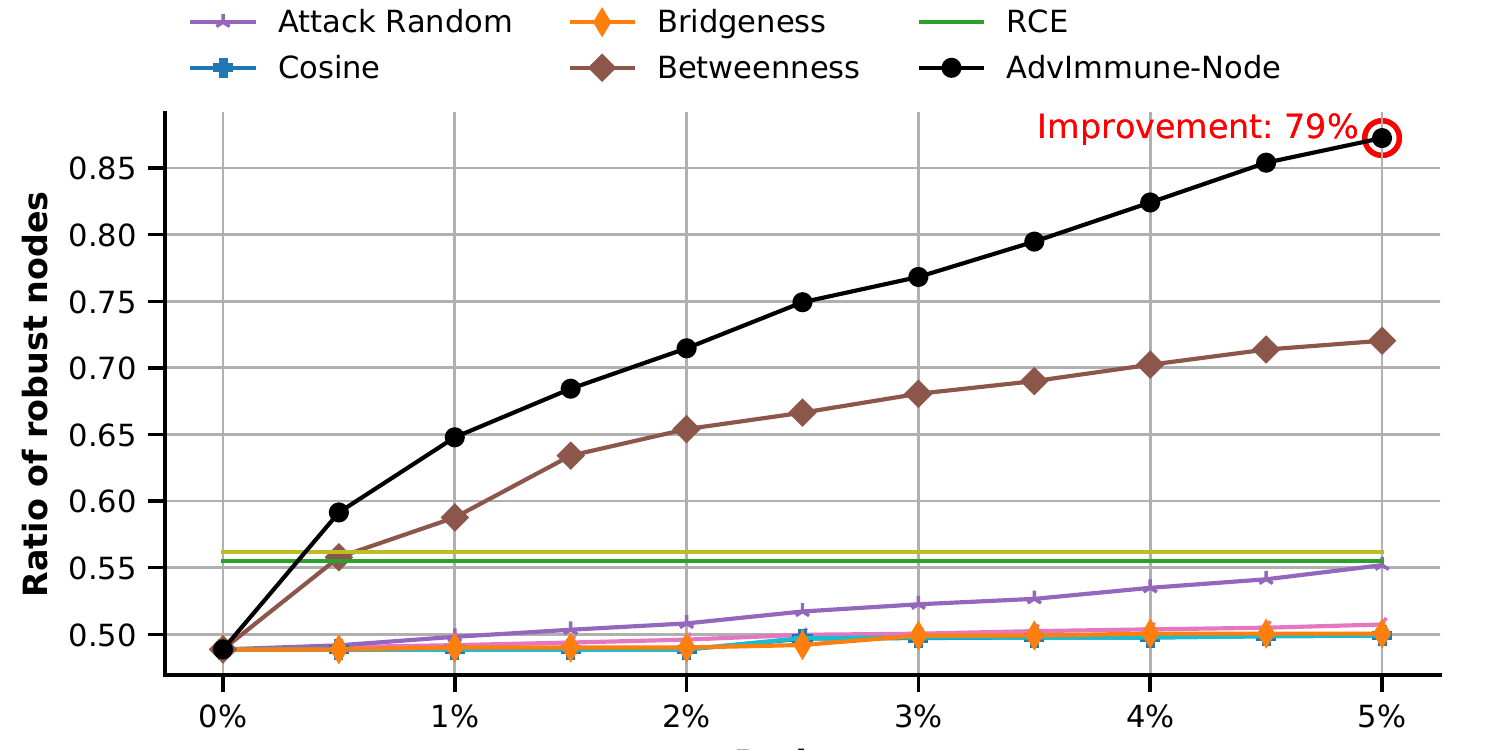}
\label{subfig:citeseer_r_node}
}
\subfigure[Ratio of robust nodes on Cora-ML]{
\includegraphics[width=5.8cm]{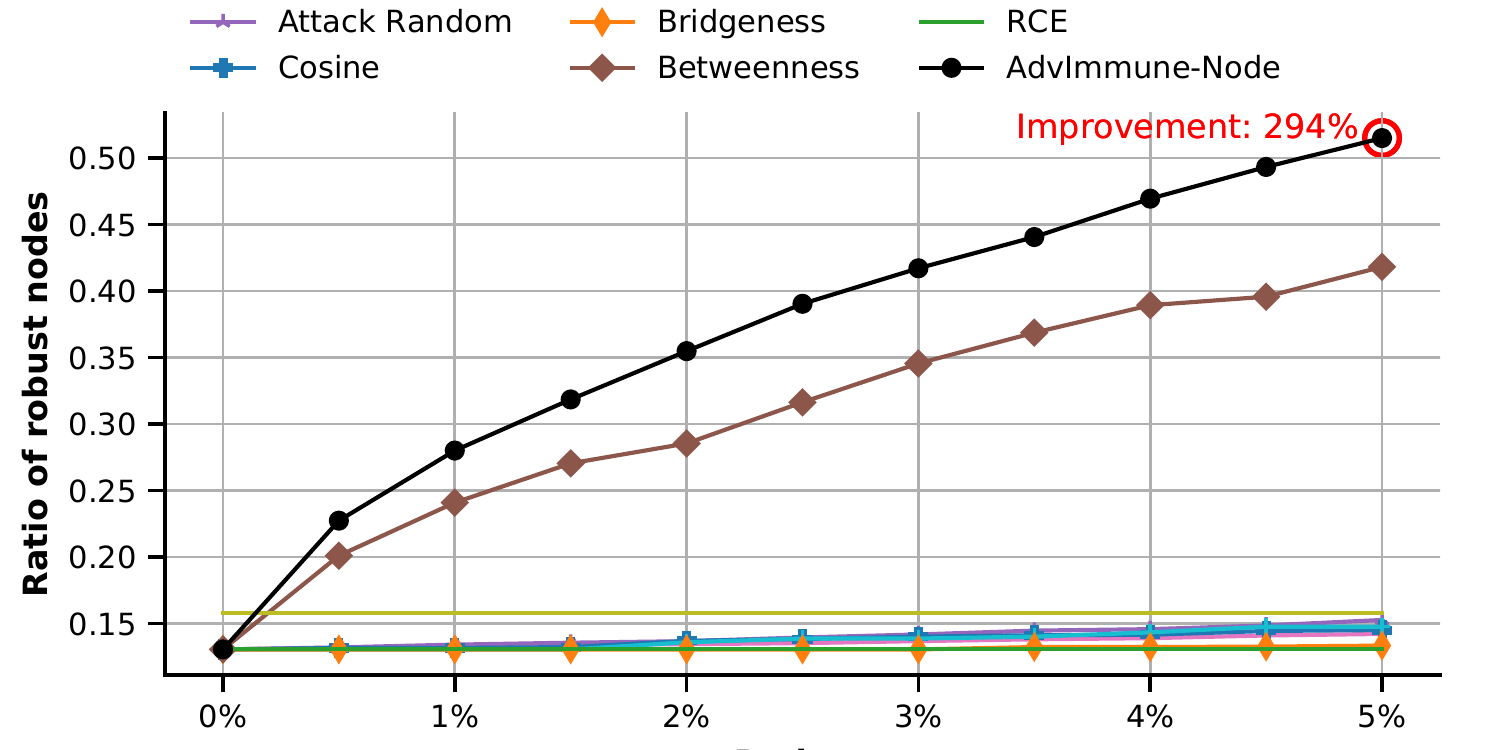}
\label{subfig:cora_r_node}
}
\subfigure[Ratio of robust nodes on Reddit]{
\includegraphics[width=5.8cm]{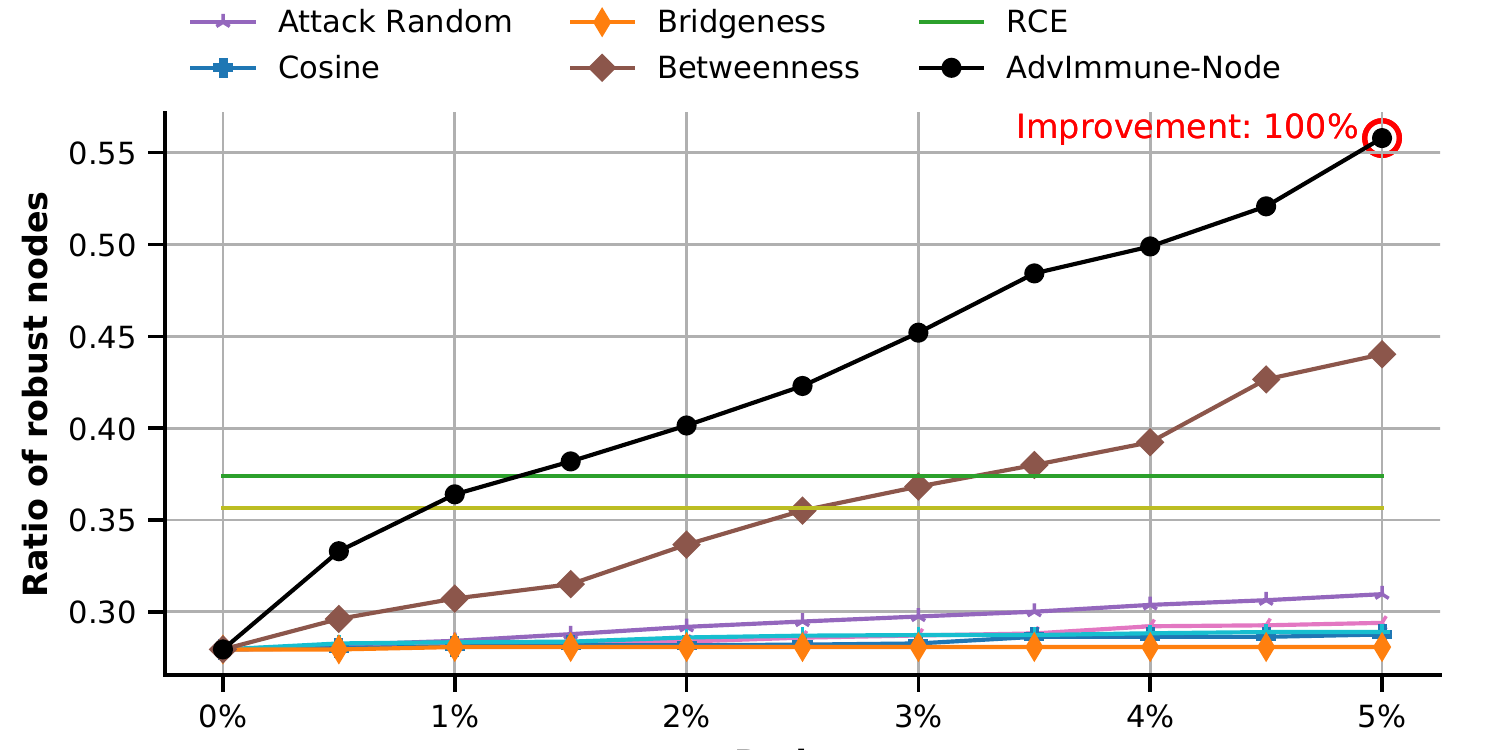}
\label{subfig:reddit_r_node}
}
\subfigure[Worst-case margin on Citeseer]{
\includegraphics[width=5.8cm]{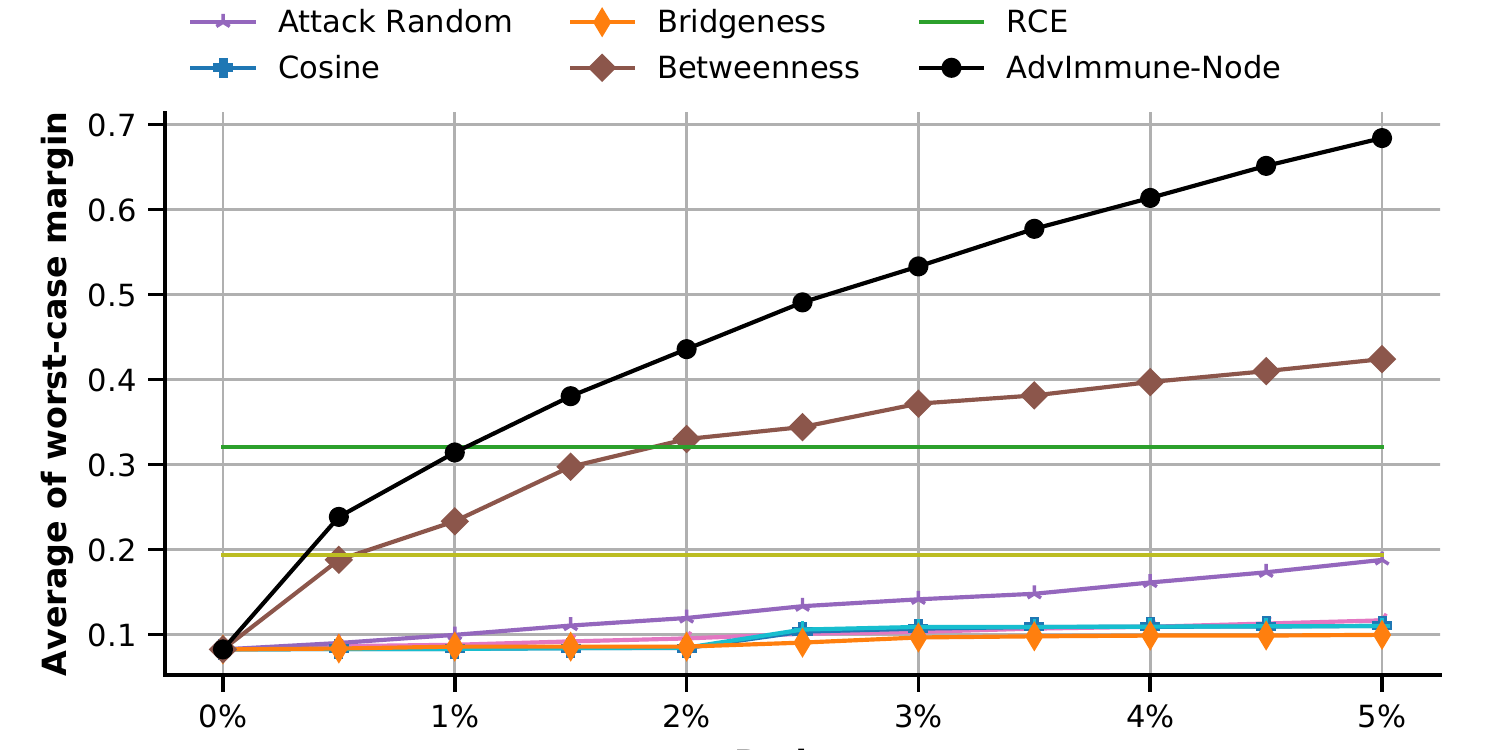}
\label{subfig:citeseer_w_node}
}
\subfigure[Worst-case margin on Cora-ML]{
\includegraphics[width=5.8cm]{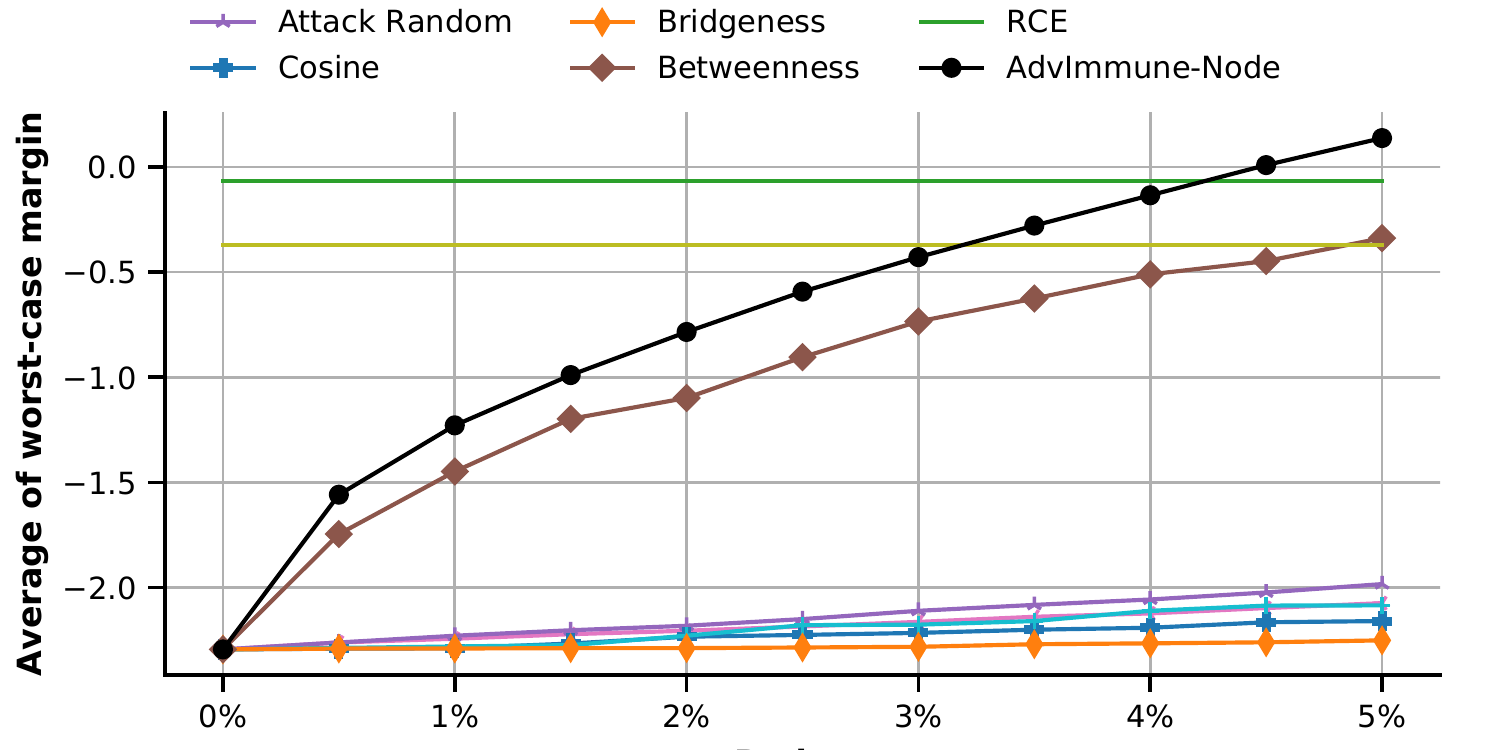}
\label{subfig:cora_w_node}
}
\subfigure[Worst-case margin on Reddit]{
\includegraphics[width=5.8cm]{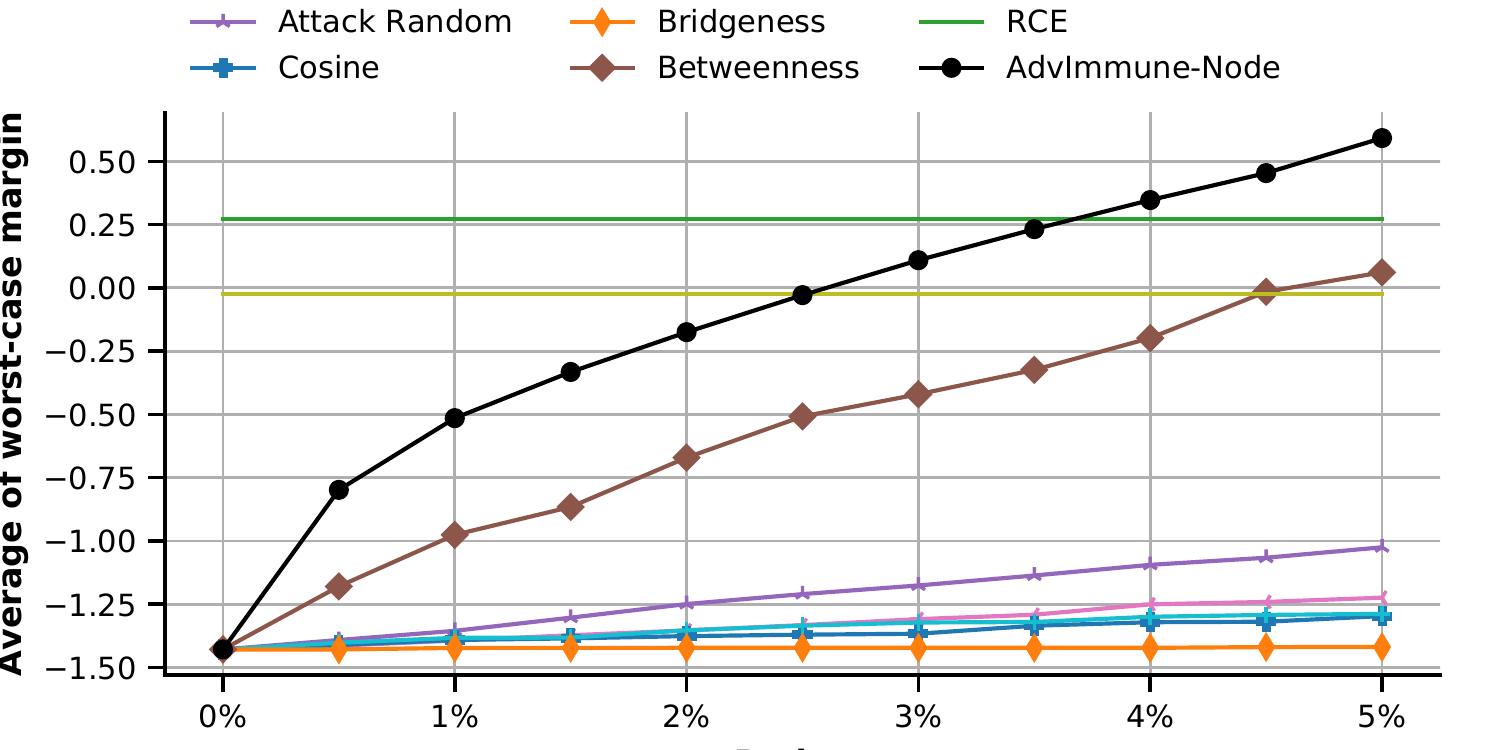}
\label{subfig:reddit_w_node}
}
\caption{
The effectiveness of node-level adversarial immunization. The upper figures (a-c) illustrate the changes in the ratio of robust nodes, whereas the lower figures (d-f) depict the changes in the average of worst-case margins.
}
\label{fig:node}
\end{figure*}

\subsubsection{Node-level immunization baselines}
We also develop node-level immunization baselines.

$\bullet$ \textbf{Random immunization methods.}  \textit{Random} method selects immune nodes randomly from all nodes, while \textit{Attack Random} selects immune nodes from the nodes that connect to the worst-case perturbed edges. 

$\bullet$ \textbf{Heuristic immunization methods.}  \emph{(1) Attribute-based methods.} We design baselines that maintain the node that is similar to its neighbor under the same class and is not similar to the disconnected node under different classes. For similarity measurement, we adopt \textit{Jaccard} and \textit{Cosine}. 
\emph{(2) Structure-based methods.} Considering network structure, we leverage node-betweenness~\cite{Girvan2002CommunitySI} for global structure importance of nodes, namely \textit{Betweenness}. Locally, we protect the node whose neighbors are similar to the neighbors of their connected nodes, and whose neighbors are dissimilar to the neighbors of other disconnected nodes, i.e., \textit{Bridgeness}.

\subsection{Performance of Edge-Level Immunization }
\label{sec:edge result}
In order to validate the efficacy of AdvImmune-Edge, we perform experiments on three standard datasets: Citeseer, Cora-ML, and Reddit. We utilize robust training~\cite{Bojchevski2019CertifiableRT} for improving certifiable robustness, as our strong baseline. The baselines optimize the GNN model by focusing on the robust cross-entropy loss and robust hinge loss, referred to as \textit{RCE} and \textit{CEM}, respectively.

Figure~\ref{fig:edge} illustrates the variation in the ratio of robust nodes with respect to the immune budget (upper figures), as well as the alterations in the average of worst-case margins (lower figures). The figure indicates that an increase in the immune budget improves the performance of all immunization methods, thereby enhancing the certifiable robustness of the graph. It is important to note that the starting point for all immunization methods is the same, which is the robust node ratio of the graph when no edges are immunized.

Regarding random-based immunization methods, Attack Random outperforms Random across all datasets, because Attack Random's knowledge of the worst-case perturbed edges targeted by the surrogate attack. Nevertheless, even with similar access to the worst-case perturbed graph, Attack Random falls significantly short of AdvImmune-Edge, validating the superiority of our approach.
As for heuristic-based baselines, both Jaccard and Cosine methods underperform compared to Attack Random, suggesting that edges with high attribute similarity exert less influence on certifiable robustness. The performance of Bridgeness is comparable to that of Jaccard and Cosine.

For state-of-the-art robust training methods, RCE and CEM exhibit similar performance. However, our AdvImmune-Edge method outperforms RCE and CEM when only immunizing 0.5\% node pairs across all datasets. 
We observe that robust training methods benefit a lot on worst-case margins since their loss is the worst-case margin.
Our AdvImmune-Edge method can increase the worst-case margin by adjusting the budget. Within an affordable budget, AdvImmune-Edge surpasses robust training methods such as RCE and CEM.
Our method can be flexibly adjusted according to different cost budgets.

As for our AdvImmune-Edge method, it significantly surpasses all the baselines on all benchmark datasets.
On Cora-ML (Figure~\ref{subfig:cora_r},~\ref{subfig:cora_w}), AdvImmune-Edge elevates the ratio of robust nodes by 206\%, achieving triple the number of original robust nodes, when immunizing 5\% node pairs. 
On Citeseer and Reddit, AdvImmune-Edge also contributes the improvement of 70\% and 93\%, when immunizing 5\% node pairs. These results demonstrate that our  AdvImmune-Edge can significantly improve the certifiable robustness of graph with an affordable immune budget.

\begin{table*}[ht]
\caption{Graph modification attacks: The accuracy of node classification by GNN under non-targeted attacks. The larger the better, the largest one is bolded, and the second largest is underlined.}
\label{tab:atkmod}
\scalebox{1.02}{
\begin{tabular}{c|ccc|ccc|ccc}
\toprule
Dataset        & \multicolumn{3}{c|}{Citeseer}                     & \multicolumn{3}{c|}{Cora-ML}                      & \multicolumn{3}{c}{Reddit}                       \\
\midrule
Method         & Clean        & Surro. attack & metattack     & Clean        & Surro. attack & metattack     & Clean        & Surro. attack & metattack     \\
\midrule
No   defense   & 0.7460          & 0.6664          & 0.6417          & 0.8480          & 0.7441          & 0.7028          & 0.9006          & 0.6312          & 0.6716          \\
\hline
RCE            & 0.7251          & 0.7052          & 0.5962          & 0.7751          & 0.4623          & 0.6317          & \underline{0.8843}          & 0.6989          & 0.5953          \\
CEM            & 0.7095          & 0.6773          & 0.5863          & 0.8018          & 0.5157          & 0.6644          & 0.8778          & 0.7107          & 0.5934          \\
FLAG           & 0.7047          & 0.5398          & 0.5943          & \textbf{0.8527} & 0.6936          & 0.7167          & 0.8817          & 0.4047          & 0.4128          \\
RobustGNN     & 0.7028          & 0.5863          & 0.5782          & 0.7712          & 0.6021          & 0.6651          & 0.8609          & 0.7680          & 0.7374          \\
GNNGuard          & \underline{0.7308}          & 0.6412          & \underline{0.7190}          & 0.8406          & \underline{0.7765}          & \textbf{0.7665} & 0.8719          & 0.6689          & \underline{0.7801}         \\
\hline
AdvImmune-Edge & \textbf{0.7460} & \textbf{0.7464} & 0.6137          & \underline{0.8480}         & \textbf{0.8246} & 0.7043          & \textbf{0.9006} & \textbf{0.8983} & 0.7084          \\
AdvImmune-Node & \textbf{0.7460} & \underline{0.7123}          & \textbf{0.7251} & \underline{0.8480}          & 0.7626          & \underline{0.7609}         & \textbf{0.9006} & \underline{0.7996}         & \textbf{0.8214}\\
\bottomrule
\end{tabular}
}
\end{table*}

\subsection{Performance of Node-Level Immunization}
We conduct node-level adversarial immunization experiments,  where immunization methods immunize nodes to improve certifiable robustness.
Our AdvImmune-Node is evaluated on three benchmark datasets compared with immunization baselines and robust training methods.
%In particular, we design a strong immunization baseline Betweenness for node-level immunization, which preferentially immunizes nodes with large node-betweenness.

Figure~\ref{fig:node} illustrates the changes in the ratio of robust nodes and worst-case margins, when varying the immune budget from 0.5$\%$ to 5$\%$.
The performances of immunization baselines are similar to edge-level immunization.
%, except for Betweenness which is the unique baseline in node-level immunization.
Betweenness performs best among the immunization baseline, suggesting that robustness benefits more from global structural importance.
For robust training methods, RCE and CEM still cannot bring much improvement to the ratio of robust nodes. However, our AdvImmune-Node outperforms both RCE and CEM, when only immunizing 0.5$\%$ nodes on Citeseer and Cora-ML and immunizing 1.5$\%$ nodes on Reddit.

Our AdvImmune-Node method significantly surpasses all the baselines across all datasets, resulting in an improvement of robust nodes by 79\%, 100\%, and 294\%. Particularly on Cora-ML, our AdvImmune-Node method nearly triples the ratio of robust nodes. On Citeseer and Reddit, our method also notably improves the ratio of robust nodes. These results highlight the robustness superiority of our AdvImmune-Node method.

\begin{table*}[t]
\caption{Node injection attack: The accuracy of node classification by GNN under targeted attacks. Larger is better and the largest accuracy is bolded.}
\label{tab:atkinj}
\scalebox{1.02}{
\begin{tabular}{c|cccc|cccc|cccc}
\toprule
Dataset         & \multicolumn{4}{c|}{Citeseer}                                  & \multicolumn{4}{c|}{Cora-ML}                                   & \multicolumn{4}{c}{Reddit}                                    \\
\midrule
Method          & Clean         & PGD           & TDGIA         & G-NIA         & Clean         & PGD           & TDGIA         & G-NIA         & Clean         & PGD           & TDGIA         & G-NIA         \\
\midrule
No defense      & 1.00          & 0.80          & 0.78          & 0.64          & 1.00          & 0.52          & 0.68          & 0.22          & 1.00          & 0.48          & 0.60          & 0.02          \\
\hline
RCE             & 0.42          & 0.40          & 0.42          & 0.42          & 0.28          & 0.28          & 0.28          & 0.28          & 0.48          & 0.26          & 0.22          & 0.32          \\
RCE*            & 0.44          & 0.42          & 0.40          & 0.44          & 0.20          & 0.14          & 0.16          & 0.34          & 0.36          & 0.26          & 0.22          & 0.34          \\
CEM             & 0.94          & 0.88          & 0.88          & 0.94          & 0.96          & 0.46          & 0.60          & 0.82          & \textbf{1.00} & 0.58          & 0.66          & 0.48          \\
CEM*            & 0.94          & 0.88          & 0.88          & 0.94          & 0.96          & 0.88          & 0.88          & 0.92          & 1.00          & 0.58          & 0.60          & 0.94          \\
FLAG            & \textbf{1.00} & 0.98          & 0.98          & \textbf{1.00} & \textbf{1.00} & 0.32          & 0.42          & 0.56          & \textbf{1.00} & 0.36          & 0.56          & 0.50          \\
RobustGNN      & 0.98          & 0.86          & 0.84          & 0.96          & 0.94          & 0.80          & 0.76          & 0.88          & 0.98          & 0.90          & 0.80          & 0.96          \\
GNNGuard           & \textbf{1.00} & \textbf{1.00} & 0.98          & 0.98          & \textbf{1.00} & 0.74          & 0.72          & \textbf{0.94} & \textbf{1.00} & 0.78          & 0.84          & 0.62          \\
\hline
AdvImmune-Edge  & \textbf{1.00} & 0.80          & 0.78         & 0.64          & \textbf{1.00} & 0.52          & 0.68          & 0.22          & \textbf{1.00} & 0.48          & 0.66          & 0.02          \\
AdvImmune-Node  & \textbf{1.00} & \textbf{1.00} & \textbf{1.00} & 0.98          & \textbf{1.00} & \textbf{1.00} & \textbf{1.00} & 0.58          & \textbf{1.00} & \textbf{1.00} & \textbf{1.00} & 0.70          \\
AdvImmune-Node* & \textbf{1.00} & \textbf{1.00} & \textbf{1.00} & \textbf{1.00} & \textbf{1.00} & \textbf{1.00} & \textbf{1.00} & 0.92          & \textbf{1.00} & \textbf{1.00} & \textbf{1.00} & \textbf{1.00} \\
\bottomrule
\end{tabular}
}
\end{table*}

\subsection{Immunization against Various Attacks}
To validate the practical defense effectiveness of the immune nodes and node pairs obtained by our AdvImmune methods, we evaluate them against various adversarial attacks.
Graph adversarial attacks can be mainly divided into graph modification attacks and node injection attacks~\cite{ZouTDGIA}.
Specifically, graph modification attackers modify network structure or node attributes, while node injection attackers inject malicious nodes, which is more practical.

To demonstrate the effectiveness of AdvImmune methods, we compare them with state-of-the-art defense methods. Existing defense methods can be mainly divided into: adversarial training, model modification, and certified defense~\cite{Sun2018AdversarialAA}. For each category, we adopt representative defense methods. 
For adversarial training method, we utilize an efficient method \textit{FLAG}~\cite{kong2020flag} to defend against attacks. 
For model modification, we adopt two state-of-the-art methods, RobustGCN~\cite{Zhu2019RobustGC} and GNNGuard~\cite{ZhangGNNGuard2020}.  
In particular, we developed two variants using $\pi$-PPNP as the base GNN for a fair comparison, namely, \textit{RobustGNN} and \textit{GNNGuard}. 
For certified defense, robust training methods (RCE and CEM~\cite{Bojchevski2019CertifiableRT}) are applied.

\subsubsection{Graph Modification Attacks}
For graph modification attacks, we adopt state-of-the-art  \textit{metattack}~\cite{zugner_adversarial_2019} and \textit{surrogate attack}~\cite{Bojchevski2019CertifiableRT} obtained by the worst-case perturbed graph, to corroborate the defense performance of our methods.
We first obtain immune nodes by AdvImmune-Node and immune node pairs by AdvImmune-Edge, and keep them unchanged the whole time. 
Then, we evaluate the defensie performance by the accuracy after defense or immunization on three settings, i.e., the clean graph (\textit{Clean}), after surrogate attack (\textit{Surro. attack}), and after metattack, respectively.

For surrogate GNN model of attacks, we adopt $\pi$-PPNP to obtain stronger attacks for better evaluation.
Besides, we employ the Meta-Self variant of metattack known for its high destructiveness~\cite{Jin2020GraphSL}. 
As for the surrogate attack, it is obtained by the perturbed graph in robustness certification.
Regarding budget, metattack can perturb 20\% edges, while immune budget is 5\% nodes or node pairs. 
%The settings can better demonstrate the effectiveness of our  AdvImmune methods.

In Table ~\ref{tab:atkmod}, we observe that most defense methods suffer from performance drops on clean graphs on all three datasets, which is unfavorable before an attack happens.
Except that FLAG brings improvement on Cora-ML, however, FLAG still loses performance on other datasets on clean graphs.
Our AdvImmune-Node and AdvImmune-Edge keep the same accuracy as original GNN model, since they only protect nodes or edges without changing the graph or GNN model.
Under the setting of attacks, not a single defense method achieves consistent superiority in accuracy across both attacks on all datasets.
We observe that defense methods perform differently when faced with different attacks, which may be because different defense methods have certain preferences for attacks.

As for ours, under surrogate attack, AdvImmune methods outperform all baselines on all datasets and bring significant improvement to model performance.
When resisting metattack, our methods achieve the best performance in most cases, except on Cora-ML, where our method achieves comparable performance to GNNGuard.
It is worth noting that the immune nodes and immune node pairs used to defend against metattack and surrogate attack are the same.
These results demonstrate that the immune nodes and immune node pairs are transferable against various attacks, improving the model performance under attacks while maintaining accuracy on clean graphs.

\subsubsection{Node Injection Attacks}
We also validate AdvImmune methods against the emerging and practical node injection attacks.
We employ PGD~\cite{Madry2017TowardsDL,Xu2019TopologyAA} and state-of-the-art attacks TDGIA~\cite{ZouTDGIA} and G-NIA~\cite{TaoGNIA} as the strong node injection attacks. 
We carry out targeted attack for node injection attacks, since G-NIA can only deal with the targeted attack, For target nodes selection, following~\cite{zugner2018adversarial,Wang2020ScalableAO}, we first train a surrogate GNN. Then among all nodes that have been correctly classified, we select 50 nodes as our target nodes to be attacked, including ten nodes with the highest margin of classification, ten nodes with the lowest margin, and thirty nodes selected randomly. 

The defense methods and AdvImmune methods also know the target nodes, thereby defense methods use these targets as training set and AdvImmune methods utilize them as targets in loss function.
The node injection budget is 50 malicious nodes and the degree of each injected node is the average degree.
The immune budgets are 10$\%$ nodes for AdvImmune-Node and 10$\%$ node pairs for AdvImmune-Edge. 
In addition, we consider a new scenario that pre-injects several attributes to get better surrogate attacks, where the attributes are generated by the strongest G-NIA. We design variants for methods that require surrogate attacks to train on this new graph with pre-injected attributes.
For robust training methods and AdvImmune-Node, the better surrogate attack can help them train better GNNs or compute better immune nodes. But for AdvImmune-Edge, the immune node pairs computed according to the surrogate attack contain virtual pre-injection attributes, which cannot correspond to actual malicious nodes injected by different attacks.
 Therefore, the variants RCE*, CEM*, and AdvImmune-Node* are included.

As shown in Table~\ref{tab:atkinj}, we observe that G-NIA is the strongest attack, which catastrophically ruins the GNN performance.
On clean graph, there are still some defense methods that cannot classify all the target nodes correctly, even though all defense baselines are trained with these targets. All of our AdvImmune methods can maintain an entirely correct accuracy.

When resisting node injection attacks, robust training methods (CEM and RCE) show a certain defensive effect, but are still not satisfactory.
The performances of their variants (CEM* and RCE*) are improved under some attacks.
%shows a good defensive performance, while RCE is still not as good as no defense, especially when defending against PGD and TDGIA on all datasets.
The adversarial training performs differently on different datasets.
%, maybe because continuous node attributes on Cora-ML and Reddit make the node injection attacks too powerful to be covered by attacks used in adversarial training.
Model modification methods perform the best among all baselines, indicating that penalizing the model’s weights on malicious edges or nodes is effective~\cite{Jin2020AdversarialAA}.
%do not face this issue, since they modify the model architecture to exclude perturbations, by penalizing the model’s weights on malicious edges or nodes~\cite{Jin2020AdversarialAA}.

As for our methods, AdvImmune-Edge has limitations in handling node injection attacks, since AdvImmune-Edge only maintains existing edges or node pairs that cannot cover the new edges injected by attackers.
AdvImmune-Node shows better defensive performance against these node injection attacks, since immunizing nodes not only protects existing edges or node pairs, but also resists the newly injected edges.
AdvImmune-Node achieves an accuracy of 1.0 in most cases, while defending against G-NIA, AdvImmune-Node still has room for improvement.
AdvImmune-Node* performs the best among all methods on Citeseer and Reddit, and is comparable to GNNGuard on Cora-ML.
In short, node-level adversarial immunization can resist node injection attacks.
% \vspace{-3pt}

\begin{figure}
\centering
\subfigure[Visualization of AdvImmune-Edge]{
\centering
\centering{\includegraphics[width=6cm]{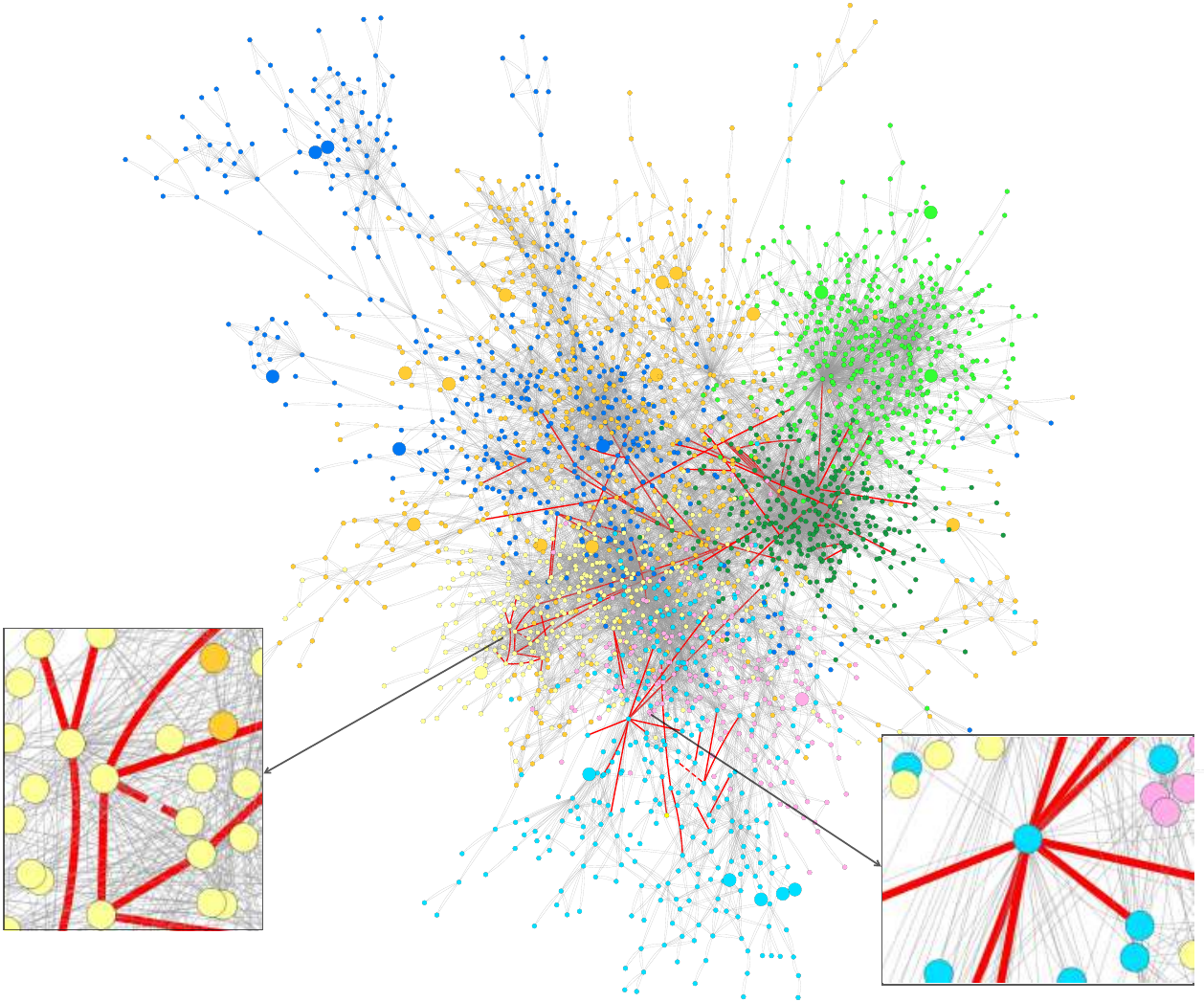}}
\label{subfig:edge_imm}
}
\subfigure[Distribution of immune edges on different indicators]{
  \includegraphics[width=3.2cm]{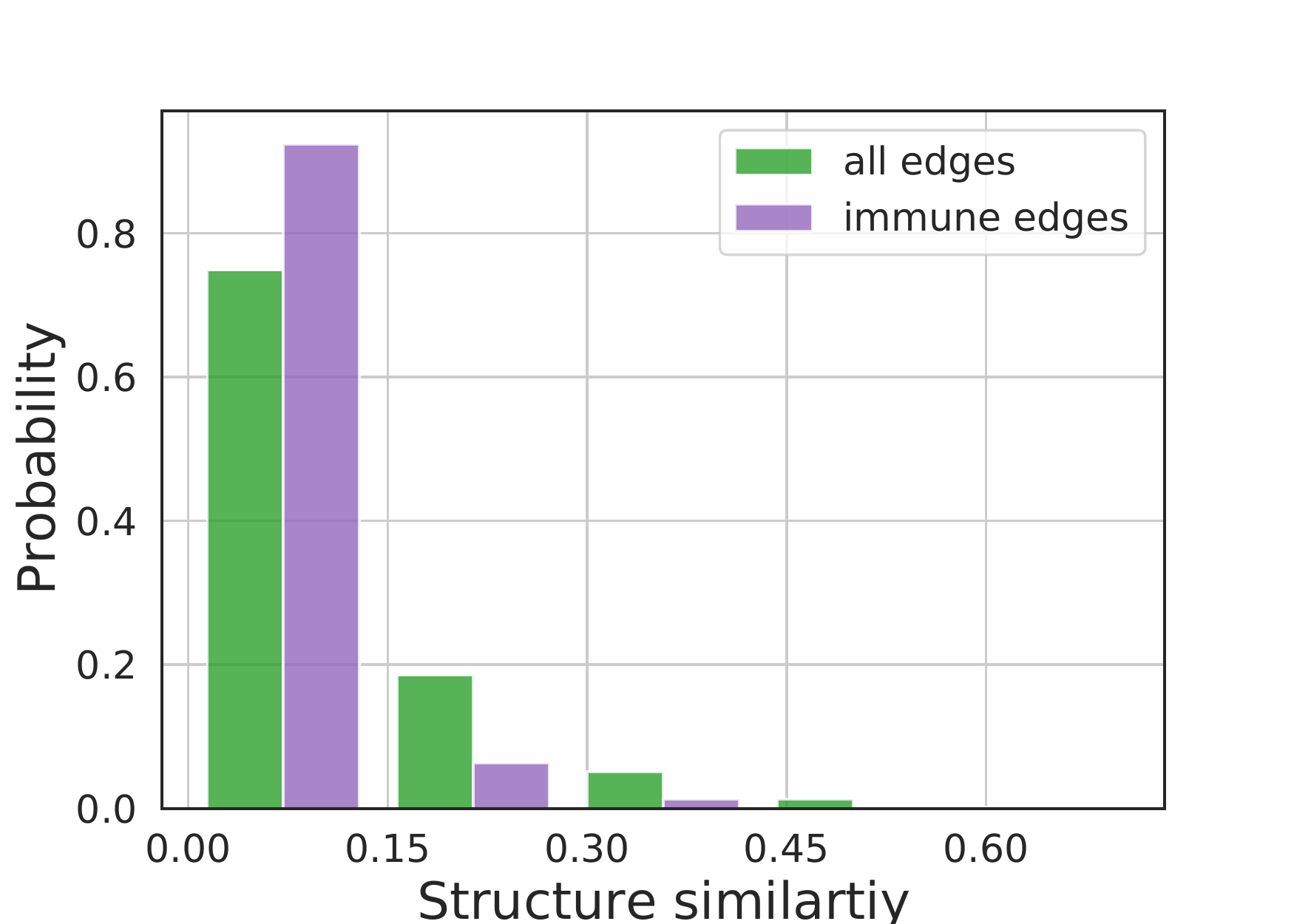}
 \includegraphics[width=2.98cm]{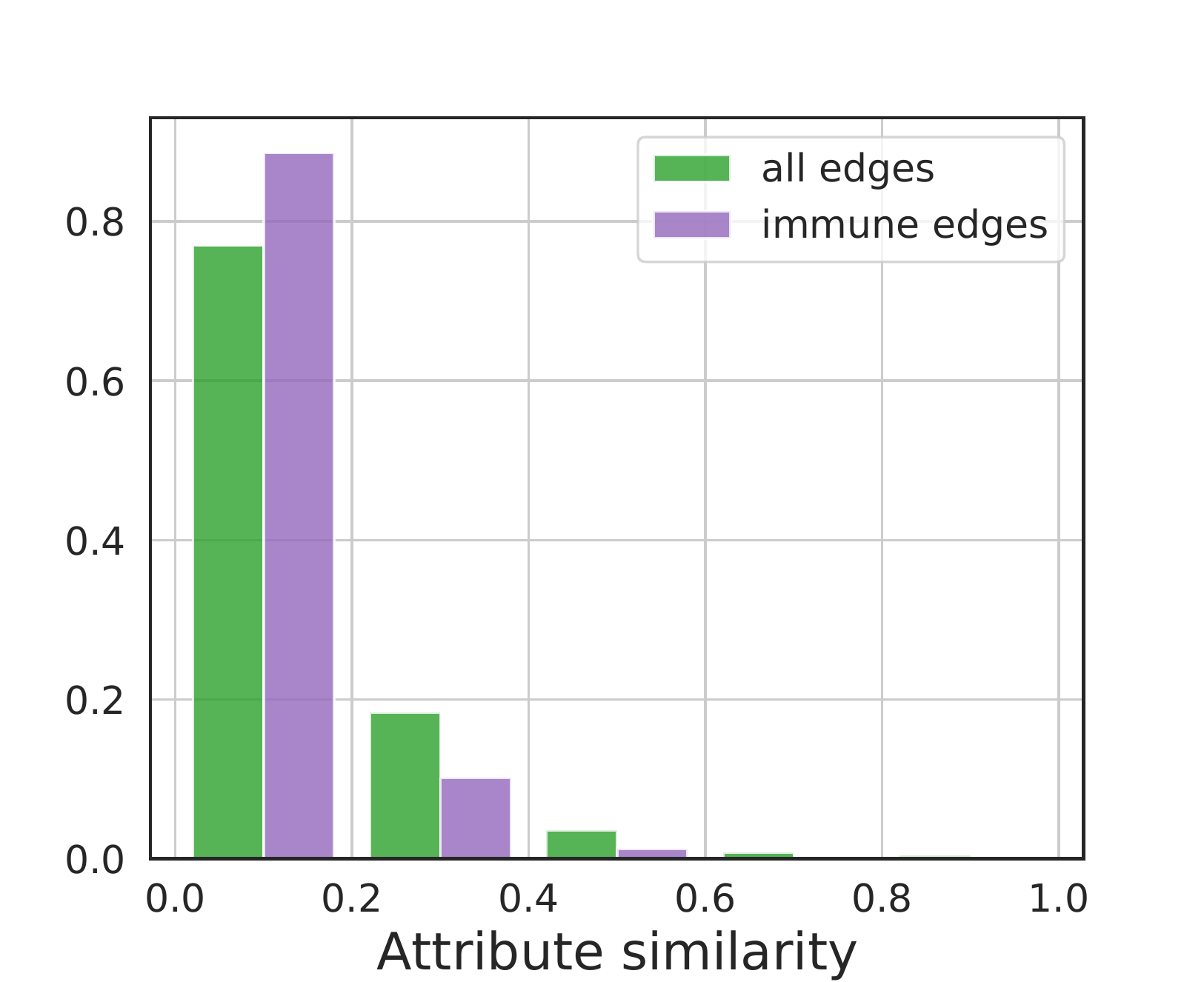}
\includegraphics[width=2.15cm]{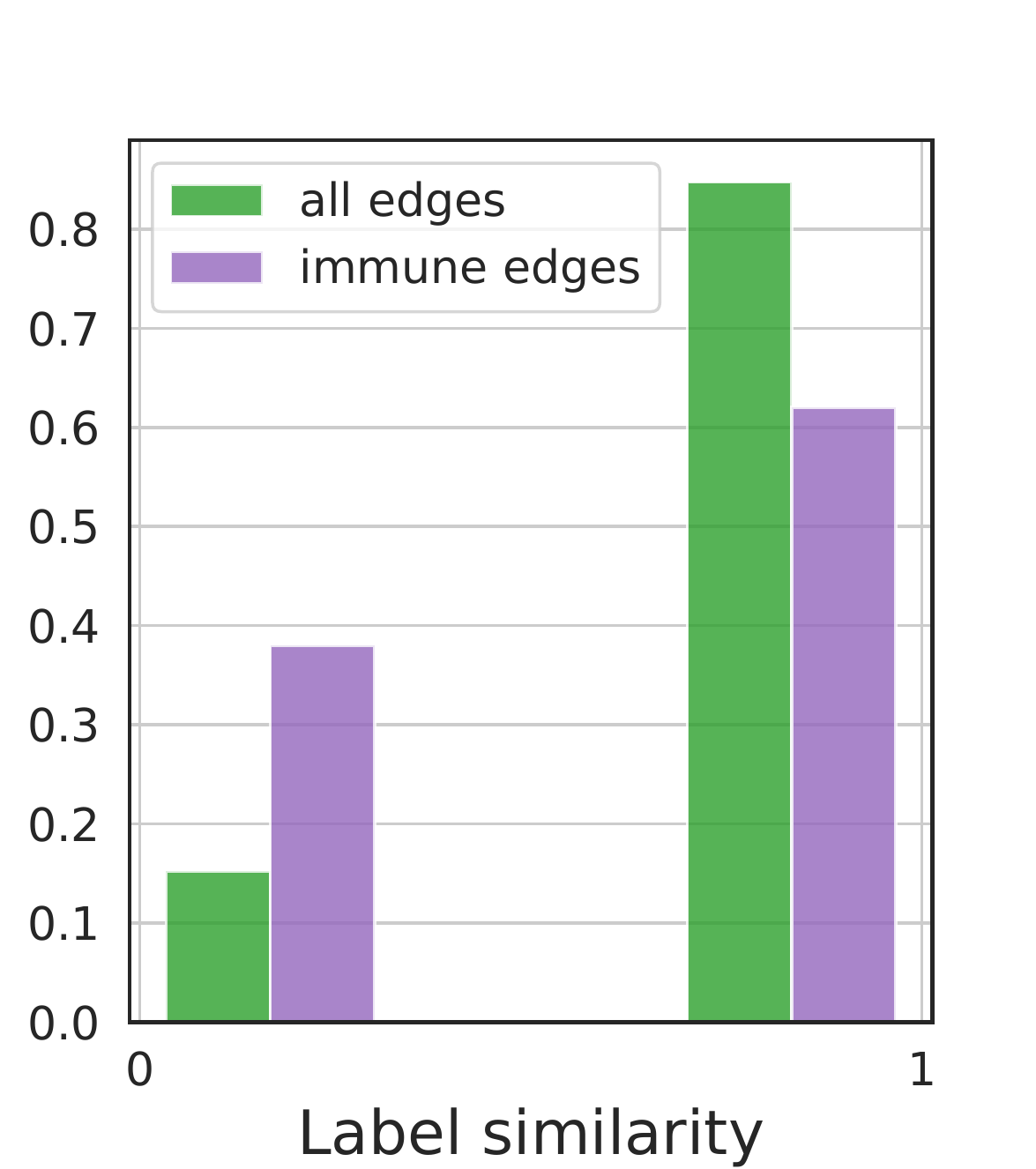}
 \label{subfig:edge_dist}
}
\caption{(a) Visualization of AdvImmune-Edge on Cora-ML.
The colorings of nodes indicate different classes. 
Larger nodes represent nodes that become robust through immunization. 
 The immune edges are red solid lines and immune node pairs are red dashed lines. 
 Insets enlarge the nodes connected to immune edges.
(b) The distribution of immune edges on structure similarity, attribute similarity, and label similarity.
} 
\label{fig:case_study_edge}
\end{figure}

\subsection{Case Study}

To obtain an intuitive understanding of our  AdvImmune-Edge and AdvImmune-Node, we adopt Cora-ML as a case study to examine immune node pairs and immune nodes from three perspectives: structure, attributes, and labels.

\subsubsection{Edge-level immunization}
Figure~\ref{subfig:edge_imm} visualizes the immune edges and the nodes that become robust through AdvImmune-Edge immunization. 
The immune edge budget is $0.001\%$ node pairs, approximately $79$. These immune edges (red edges) aid $26$ nodes become robust (larger nodes).
The distribution of immune edges across the three aspects is presented in Figure~\ref{subfig:edge_dist}.

\textbf{Structure analysis.}
Bridgeness is used as a measure to analyze the structural characteristics of immune edges. 
As shown in the left figure of Figure~\ref{subfig:edge_dist}, the bridgeness of immune edges is smaller than other edges.
We observe that in Figure~\ref{subfig:edge_imm}, the enlarged immune edges are radial around the center nodes so that these external spoke nodes have no common neighbors with the central node.

\textbf{Attribute analysis.}
We examine the node attribute similarity of immune edges. In the middle figure of Fig.~\ref{subfig:edge_dist}, the attribute similarity of nodes at both ends of immune edges primarily ranges from $0$ to $0.2$. The node attribute similarity of immune edges is lower. This might explain why attribute-based baselines do not perform well in Figure~\ref{fig:edge}.

\textbf{Label analysis.}
We evaluate the nodes' label similarity of immune edges. Over $80\%$ edges have the same node label at both ends, whereas the ratio for immune edges is only $60\%$. This could be attributed to the personalized PageRank of $\pi$-PPNP: $\mathbf{\Pi} =(1-\alpha)\left(\mathbf{I}_{N}-\alpha \mathbf{D}^{-1} \mathbf{A} \right)^{-1}$, which involves an inverse operation, resulting in a dense diffusion matrix and making heterogeneous edges useful by altering diffusion weights. It's a noteworthy discovery that immunizing such heterogeneous edges also contributes to the robustness of the graph, warranting further investigation.

\begin{figure}
%\begin{subfigure}
\centering
\subfigure[Visualization of AdvImmune-Node]{
\centering \includegraphics[width=6cm]{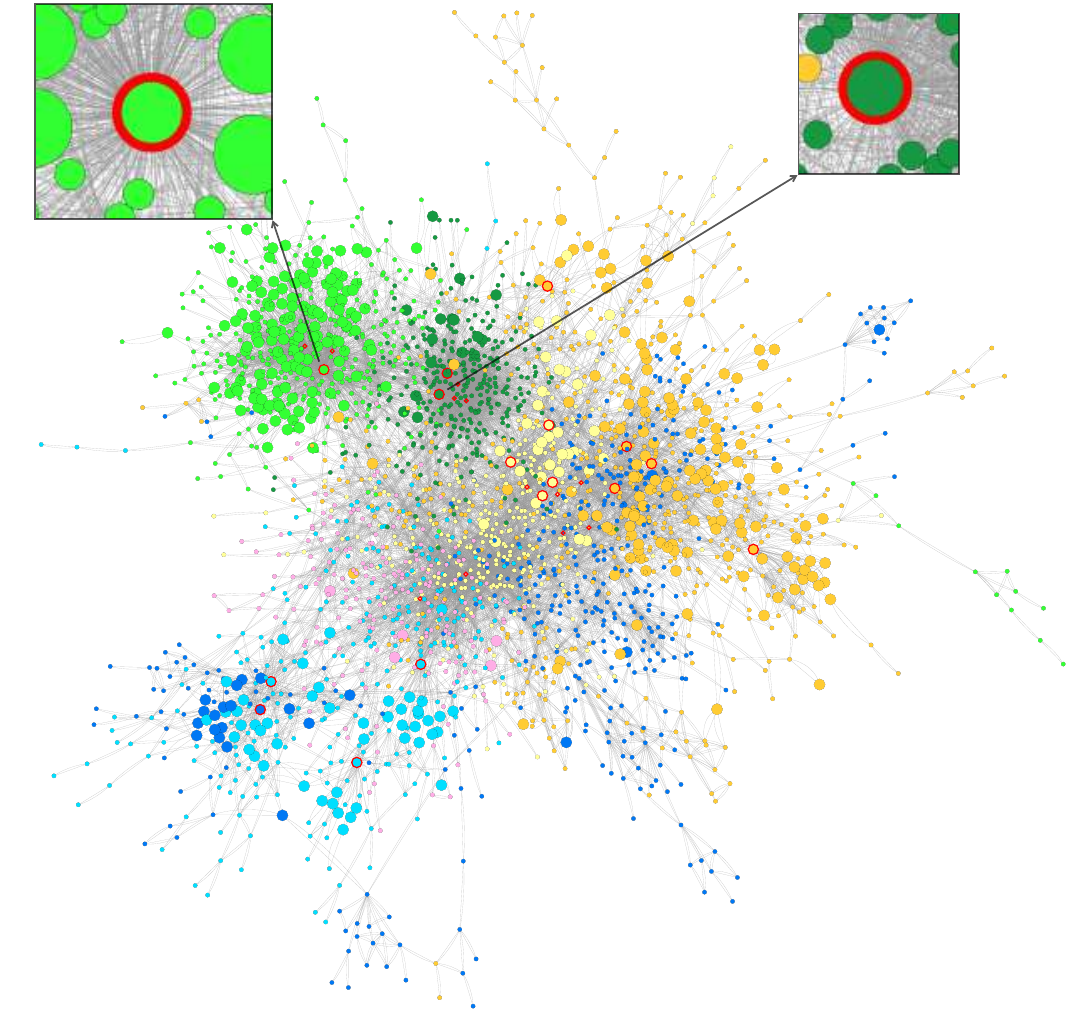}
\label{subfig:node_imm}
%\end{center}
}
%\end{subfigure}
\subfigure[Distribution of immune nodes on different indicators]{
  \includegraphics[width=3.21cm]{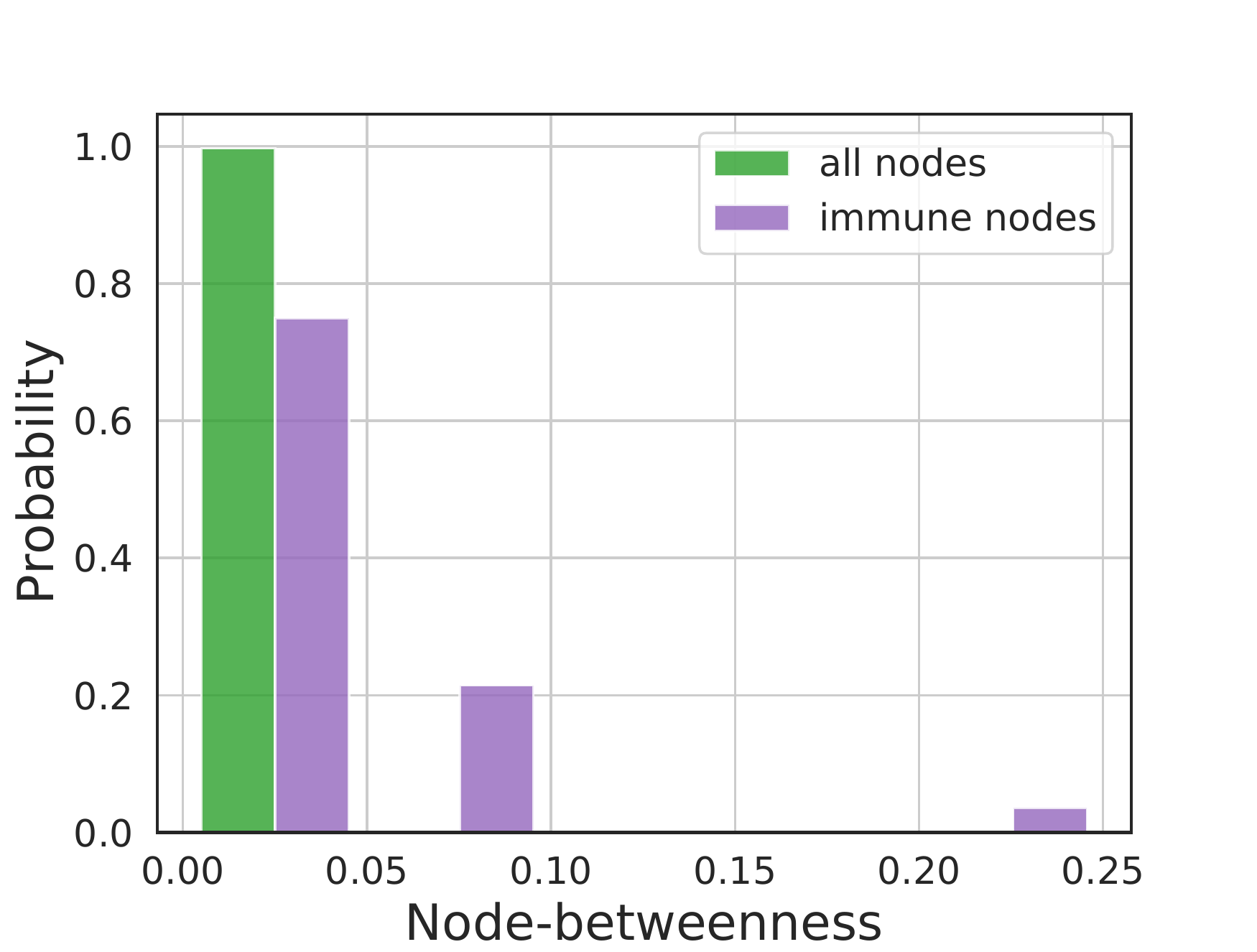}
 \includegraphics[width=2.98cm]{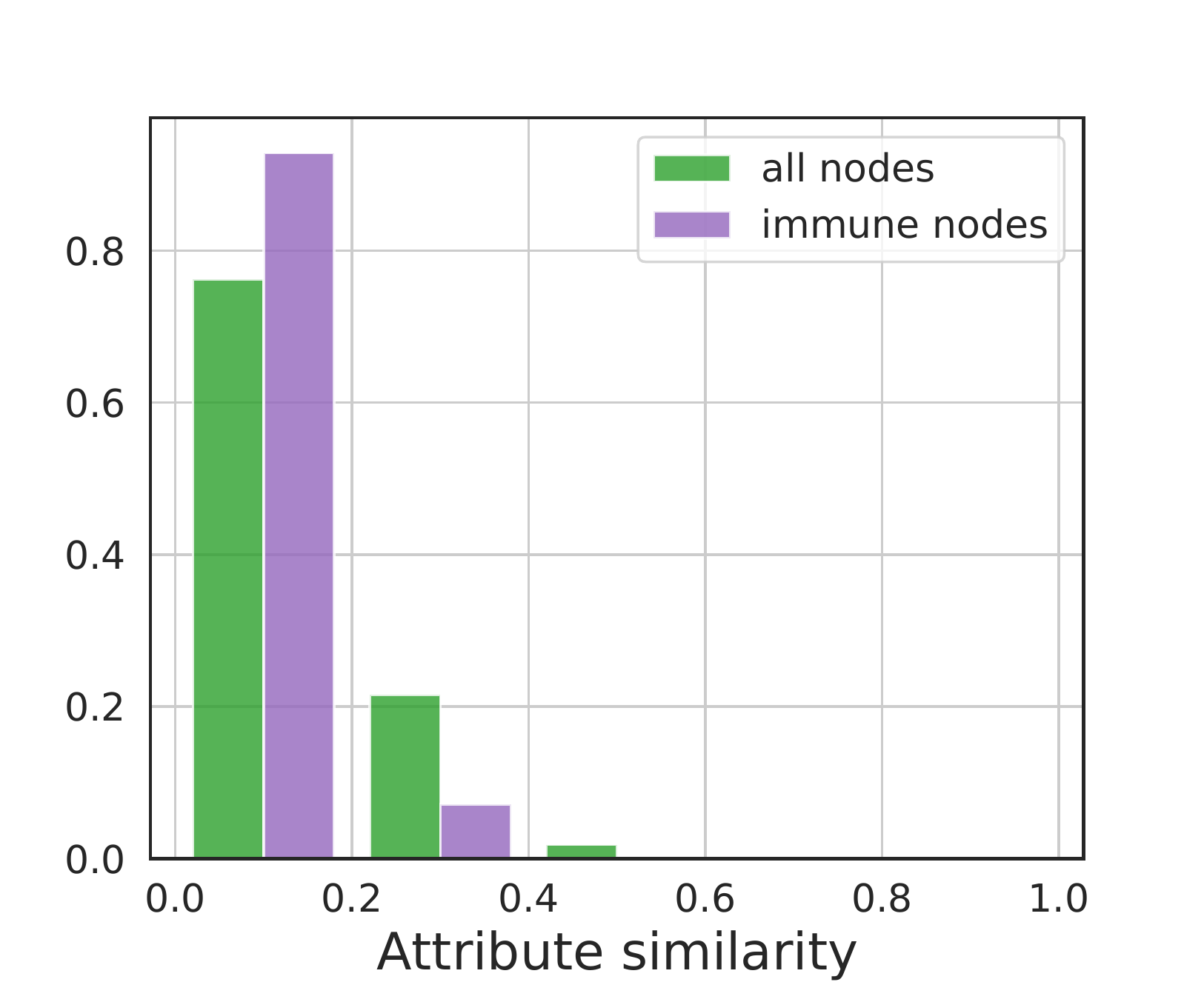}
\includegraphics[width=2.16cm]{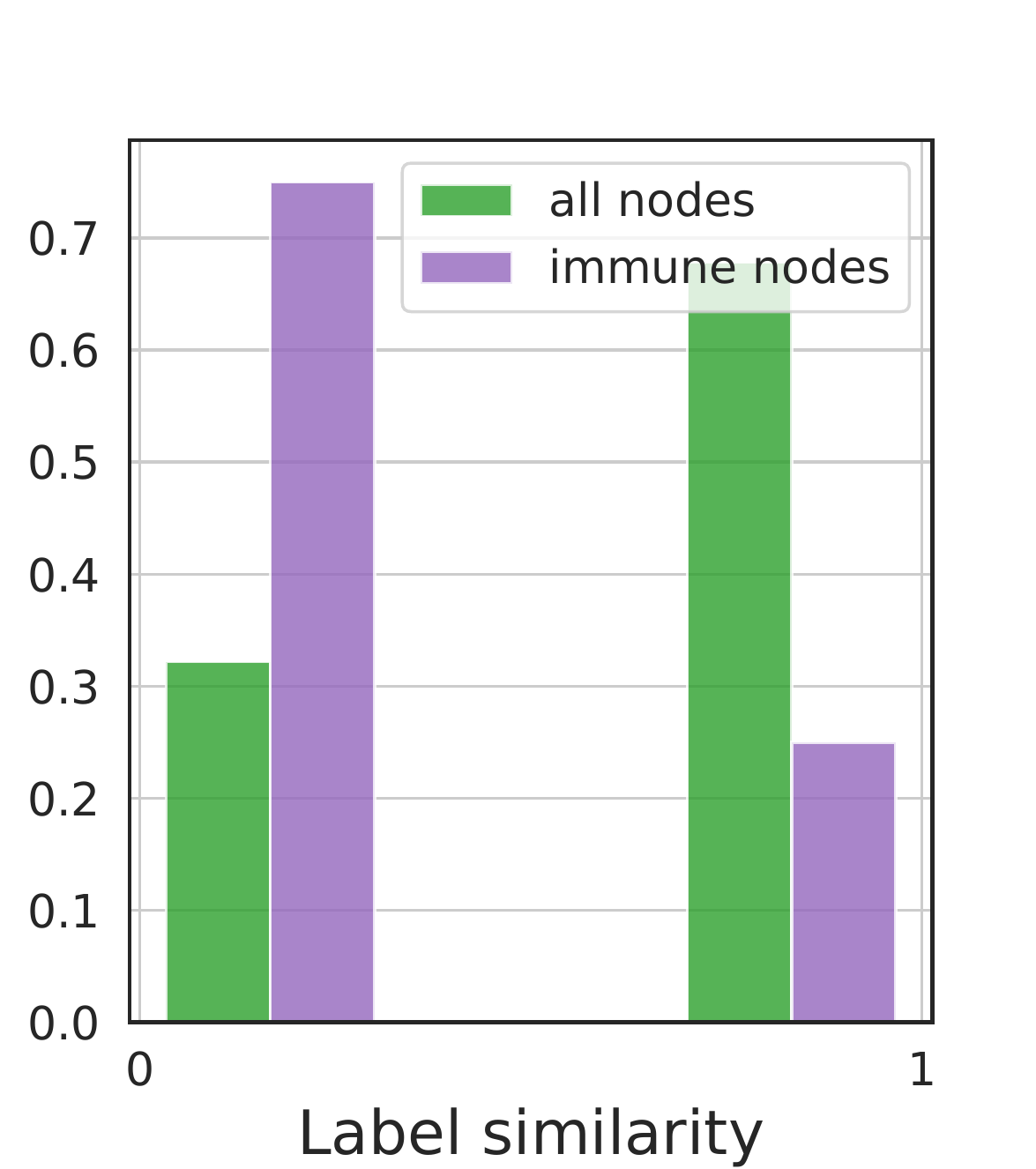}
 \label{subfig:node_dist}
}
\caption{(a) Visualization of AdvImmune-Node.
%The colorings of nodes indicate different classes. 
The red circles indicate the immune nodes. 
Larger nodes mean nodes that become robust through immunization. 
 Insets enlarge some immune nodes.
(b) The distribution of immune nodes on node-betweenness, attribute similarity, and label similarity.
} 
\label{fig:case_study_node}
\end{figure}

\subsubsection{Node-level immunization}
Figure~\ref{subfig:node_imm}  provides a visualization of  the immune nodes and the nodes that become robust through AdvImmune-Node immunization. 
The immune budget is $1\%$ nodes, about $28$. After immunization, $420$ nodes become robust. 
Figure~\ref{subfig:node_dist} shows the distribution of immune nodes in terms of structure, attributes, and labels.

\textbf{Structure analysis.}
The left figure of Figure~\ref{subfig:node_dist} shows that the node-betweenness of immune nodes is statistically larger than that of other nodes, explaining why the Betweenness method is the strongest heuristic immunization baseline. In Figure~\ref{subfig:node_imm}, the light green node with the largest degree that has been immunized is enlarged. The second largest immune node is also enlarged. This suggests that \textit{AdvImmune-Node} tends to protect nodes with significant structural dominance, which is crucial for enhancing the certifiable robustness of the graph.

\textbf{Attribute analysis.}
The middle figure of Figure~\ref{subfig:node_dist} presents the attribute similarity of immune nodes and their neighbors. The attribute similarity of immune nodes and their neighbors is notably lower than that of other nodes, which could explain why attribute-based baselines underperform as shown in Figure~\ref{fig:node}.

\textbf{Label analysis.}
We present the label similarity between immune nodes and their neighbors. Approximately 70\% of nodes share the same label with their neighbors, while only 25\% of immune nodes do. The reason may be similar to the edge-level immunization above.

Experiments conducted at both the node and edge levels on Citeseer and Reddit yielded similar results, which are not included here due to space constraints.

\section{Related Work}
GNNs have demonstrated impressive success on various graph mining tasks such as node classification~\cite{Klicpera2018PredictTP,hamilton2017inductive}, graph classification~\cite{rieck2019persistent}, and fraud detection~\cite{Ma2021Survey,Cheng2022Fraud}. However, their vulnerability to adversarial attacks presents substantial security risks, impeding the implementation of GNNs in the real-world application.~\cite{zugner2018adversarial,Dai2018AdversarialAO}.

\subsection{Adversarial attacks}
Adversarial attacks~\cite{Chen2020ASO, Wang2019AttackingGC, Sun2018AdversarialAA,Zhang2022Proj} can be mainly divided into graph modification attacks and node injection attacks~\cite{ZouTDGIA}.
For graph modification attack,  Nettack~\cite{zugner2018adversarial,zugner2020AdversarialA} employs gradient to attack node attributes and graph structure. RL-S2V~\cite{Dai2018AdversarialAO} leverages reinforcement learning to flip edges. Metattack~\cite{zugner_adversarial_2019} contaminates the graph structure with meta-gradient, representing the latest advancement in graph modification attacks.
Some other researches perturb structure with approximation techniques~\cite{Wang2019AttackingGC}, or by extracting subgraph~\cite{Li2021AdversarialA}, or in balck-box scenario~\cite{Chang20ARestrict, Chang2022Adversarial}. 

Node injection attacks are emerging and practical, injecting malicious nodes without modifying existing graph~\cite{Sun2020AdversarialAO,Wang2020ScalableAO,ZouTDGIA}. 
Specifically, NIPA~\cite{Sun2020AdversarialAO} employs hierarchical reinforcement learning to sequentially generate labels and edges for malicious nodes. 
Unfortunately, NIPA suffers from high computational costs and is not scalable enough for large-scale datasets~\cite{ZouTDGIA}.
TDGIA~\cite{ZouTDGIA} heuristically selects the defective edges for injecting nodes and adopts smooth optimization to generate features for injected nodes.
G-NIA~\cite{TaoGNIA} focuses on the single node injection attack, achieving great attack performance when only injecting one node.

\subsection{Defense methods}
Defense methods against attacks can be mainly categorized into three types: adversarial training, model modification, and certified defense~\cite{Jin2020AdversarialAA, Sun2018AdversarialAA}.
Adversarial training is a widely used defense method~\cite{Jin2020AdversarialAA}, where the model is trained on adversarial examples and clean examples, such as GraphAT~\cite{feng2019graph}, AdvT~\cite{Dai2019AdversarialTM}, FLAG~\cite{kong2020flag}, and so on.
Model modification focuses on redesigning the model structure or purifying adversarial perturbations~\cite{Jin2020AdversarialAA}.
%RobustGCN~\cite{Zhu2019RobustGC} adopts Gaussian distributions as hidden representations, which can defend against nettack~\cite{zugner2018adversarial} and RL-S2V~\cite{Dai2018AdversarialAO}. 
RobustGCN employs Gaussian distributions in graph convolutional layers to mitigate the impact of adversarial attacks~\cite{Zhu2019RobustGC}. 
%GNNGuard~\cite{ZhangGNNGuard2020} mitigates adversarial effects by modifying neural message passing of GNN.
GNNGuard applies the network theory of homophily to learn how to assign higher weights to edges connecting similar nodes while pruning edges between unrelated nodes~\cite{ZhangGNNGuard2020}. 
%GNN-SVD~\cite{Entezari2020AllYN} adopts a low-rank approximation of the graph to defend attacks.
Jaccard~\cite{Wu2019AdversarialEF} prunes edges that connect two dissimilar nodes.
GaSoliNe\cite{Xu2022GraphSanitation} employs an approximation of hyper-gradient to refine graph.
ProGNN~\cite{Jin2020GraphSL} concurrently learns the graph structure and GNN parameters through optimization of feature smoothness, low-rank and sparsity.
Bayesian graph neural networks~\cite{Zhang2018BayesianGC} and transfer learning~\cite{Tang2020TransferringRF} are also used in other defense methods. 

Throughout the development of graph adversarial learning, there has been a constant struggle between attack methods being defended, and defense methods falling short against the next attack. This ongoing cycle may impede the progress of graph adversarial learning. 
Recently, certified defense methods (robustness certification~\cite{Liu2020CertifiableRT} and robust training~\cite{bojchevski_sparsesmoothing_2020,Jia2020CertifiedRO}) have emerged to address this issue,  garnering significant attention~\cite{WangJCG21,SchuchardtBKG21}.
% Z\"ugner \textit{et al.}~\cite{Zgner2019CertifiableRA} verify certifiable robustness w.r.t. perturbations on attributes. 
Bojchevski \textit{et al.}~\cite{Bojchevski2019CertifiableRT} provide certification w.r.t. perturbations on structures. 
%Additionally, There are several recent  based on randomized smoothing~\cite{bojchevski_sparsesmoothing_2020}. 
Several robustness certifications, which are based on randomized smoothing methodologies \cite{bojchevski_sparsesmoothing_2020,SchuchardtBKG21}, calculate the certified radius for the smoothed classifier. They likely mitigate the impact of potential attacks.
We choose certifiable robustness of Bojchevski \textit{et al.}~\cite{Bojchevski2019CertifiableRT}, since the requirements for being certified as a robust node are more strict, that is, only those that can be predicted correctly under the worst-case perturbation are certifiable robust nodes.

Unlike previous studies, this paper is the first to investigate the potential and practicality of adversarial immunization for certifiable robustness of graphs.

\section{Conclusion}
This paper introduces a novel concept, \emph{graph adversarial immunization}, aimed at improve the certifiable robustness of graphs against any admissible attack. Adversarial immunization vaccinates a fraction of node pairs or nodes in advance, corresponding to node-level and edge-level immunization. To bypass the computational challenges of combinatorial optimization, we propose two efficient algorithms, AdvImmune-Edge and AdvImmune-Node. These algorithms address the immunization problems using meta-gradient and robustness gain, respectively. Experiments conducted on three benchmark datasets demonstrate that our AdvImmune methods significantly improve the certifiable robustness of graphs, outperforming all baselines. We also confirm the defense capabilities of our methods against various attacks. Edge-level immunization is suitable for scenarios that demand precise immunization, while node-level immunization is more practical and effective against emerging node injection attacks. In summary, we present the first action guideline to improve certifiable robustness from a graph data perspective without compromising performance on clean graphs, offering a fresh perspective on graph adversarial learning.

\textbf{Limitations and future work.} 
Our initial exploration into adversarial immunization has revealed some areas for improvement. The robustness certification on which it relies is dependent on PPNP-like GNN models, leading to a high degree of complexity. 
Currently, our immunization mainly focused on worst-case-based robustness certification~\cite{Bojchevski2019CertifiableRT}. This immunization mechanism can also be extended to randomized-smoothing-based certifications~\cite{bojchevski_sparsesmoothing_2020,SchuchardtBKG21}. 
%For example, we can improve these certification by protecting critical edges/nodes, thereby increasing the lower bound of the probability that $\boldsymbol{x}$ is still correctly classified after randomized smoothing ($\underline{p_{y}(\boldsymbol{x})}$ in~\cite{bojchevski_sparsesmoothing_2020}).
%The exploration of immunization for robustness certifications based on randomized smoothing is earmarked for future direction. 
%We regard adversarial immunization as a valuable area for future research.
Investigating immunization strategies for randomized-smoothing-based robustness certifications is identified as a noteworthy direction for future research. We regard adversarial immunization as a significant field warranting further scholarly exploration.
%We leave the immunization for randomized-smoothing-based certifications as a future research direction.
%Furthermore, adversarial immunization struggles to generalize the certifiable robustness based on randomized smoothing, since this robustness is solely dependent on model prediction outcomes and is not influenced by the data. 

% if have a single appendix:
%\appendix[Proof of the Zonklar Equations]
% or
%\appendix  % for no appendix heading
% do not use \section anymore after \appendix, only \section*
% is possibly needed

% use appendices with more than one appendix
% then use \section to start each appendix
% you must declare a \section before using any
% \subsection or using \label (\appendices by itself
% starts a section numbered zero.)
%

%\appendices
%\section{Proof of the First Zonklar Equation}
%Appendix one text goes here.
%
%% you can choose not to have a title for an appendix
%% if you want by leaving the argument blank
%\section{}
%Appendix two text goes here.

% use section* for acknowledgment
\ifCLASSOPTIONcompsoc
  % The Computer Society usually uses the plural form
  \section*{Acknowledgments}
\else
  % regular IEEE prefers the singular form
  \section*{Acknowledgment}
\fi
This work is funded by the National Key R\&D Program of China (2022YFB3103700, 2022YFB3103701), and the National Natural Science Foundation of China under Grant Nos. 62102402, 62272125, U21B2046. Huawei Shen is also supported by Beijing Academy of Artificial Intelligence (BAAI).

% Can use something like this to put references on a page
% by themselves when using endfloat and the captionsoff option.
\ifCLASSOPTIONcaptionsoff
  \newpage
\fi

% trigger a \newpage just before the given reference
% number - used to balance the columns on the last page
% adjust value as needed - may need to be readjusted if
% the document is modified later
%\IEEEtriggeratref{8}
% The "triggered" command can be changed if desired:
%\IEEEtriggercmd{\enlargethispage{-5in}}

% references section

% can use a bibliography generated by BibTeX as a .bbl file
% BibTeX documentation can be easily obtained at:
% http://mirror.ctan.org/biblio/bibtex/contrib/doc/
% The IEEEtran BibTeX style support page is at:
% http://www.michaelshell.org/tex/ieeetran/bibtex/
\bibliographystyle{IEEEtran}
% argument is your BibTeX string definitions and bibliography database(s)
\bibliography{adv_immune_node.bib}

% Generated by IEEEtran.bst, version: 1.14 (2015/08/26)
\begin{thebibliography}{10}
\providecommand{\url}[1]{#1}
\csname url@samestyle\endcsname
\providecommand{\newblock}{\relax}
\providecommand{\bibinfo}[2]{#2}
\providecommand{\BIBentrySTDinterwordspacing}{\spaceskip=0pt\relax}
\providecommand{\BIBentryALTinterwordstretchfactor}{4}
\providecommand{\BIBentryALTinterwordspacing}{\spaceskip=\fontdimen2\font plus
\BIBentryALTinterwordstretchfactor\fontdimen3\font minus
  \fontdimen4\font\relax}
\providecommand{\BIBforeignlanguage}[2]{{%
\expandafter\ifx\csname l@#1\endcsname\relax
\typeout{** WARNING: IEEEtran.bst: No hyphenation pattern has been}%
\typeout{** loaded for the language `#1'. Using the pattern for}%
\typeout{** the default language instead.}%
\else
\language=\csname l@#1\endcsname
\fi
#2}}
\providecommand{\BIBdecl}{\relax}
\BIBdecl

\bibitem{kipf2017semi}
T.~N. Kipf and M.~Welling, ``Semi-supervised classification with graph
  convolutional networks,'' in \emph{International Conference on Learning
  Representations}, 2017.

\bibitem{Klicpera2018PredictTP}
J.~Klicpera, A.~Bojchevski, and S.~G{\"u}nnemann, ``Predict then propagate:
  Graph neural networks meet personalized pagerank,'' in \emph{International
  Conference on Learning Representations}, 2019.

\bibitem{xu2018gwnn}
B.~Xu, H.~Shen, Q.~Cao, Y.~Qiu, and X.~Cheng, ``Graph wavelet neural network,''
  in \emph{International Conference on Learning Representations}, 2019.

\bibitem{Cao2020PopularityPO}
Q.~Cao, H.~Shen, J.~Gao, B.~Wei, and X.~Cheng, ``Popularity prediction on
  social platforms with coupled graph neural networks,'' in \emph{Proceedings
  of the 13th International Conference on Web Search and Data Mining}, ser.
  WSDM '20, 2020, pp. 70--78.

\bibitem{fan2019graph}
W.~Fan, Y.~Ma, Q.~Li, Y.~He, E.~Zhao, J.~Tang, and D.~Yin, ``Graph neural
  networks for social recommendation,'' in \emph{The World Wide Web
  Conference}, ser. WWW '19, 2019, pp. 417--426.

\bibitem{Ma2021Survey}
X.~Ma, J.~Wu, S.~Xue, J.~Yang, C.~Zhou, Q.~Z. Sheng, H.~Xiong, and L.~Akoglu,
  ``A comprehensive survey on graph anomaly detection with deep learning,''
  \emph{IEEE Transactions on Knowledge and Data Engineering}, pp. 1--1, 2021.

\bibitem{Cheng2022Fraud}
D.~Cheng, X.~Wang, Y.~Zhang, and L.~Zhang, ``Graph neural network for fraud
  detection via spatial-temporal attention,'' \emph{IEEE Transactions on
  Knowledge and Data Engineering}, vol.~34, no.~8, pp. 3800--3813, 2022.

\bibitem{Dai2018AdversarialAO}
H.~Dai, H.~Li, T.~Tian, X.~Huang, L.~Wang, J.~Zhu, and L.~Song, ``Adversarial
  attack on graph structured data,'' in \emph{Proceedings of the 35th
  International Conference on Machine Learning}, ser. ICML '18, 2018, pp.
  1123--1132.

\bibitem{zugner2018adversarial}
D.~Z\"{u}gner, A.~Akbarnejad, and S.~G\"{u}nnemann, ``Adversarial attacks on
  neural networks for graph data,'' in \emph{Proceedings of the 24th ACM SIGKDD
  International Conference on Knowledge Discovery \& Data Mining}, ser. KDD
  '18, 2018, pp. 2847--2856.

\bibitem{Bojchevski2018AdversarialAO}
A.~Bojchevski and S.~G{\"u}nnemann, ``Adversarial attacks on node embeddings
  via graph poisoning,'' in \emph{Proceedings of the 36th International
  Conference on Machine Learning}, ser. ICML '19, 2019, pp. 695--704.

\bibitem{TaoGNIA}
S.~Tao, Q.~Cao, H.~Shen, J.~Huang, Y.~Wu, and X.~Cheng, ``Single node injection
  attack against graph neural networks,'' in \emph{Proceedings of the 30th ACM
  International Conference on Information and Knowledge Management}, ser. CIKM
  '21, 2021, p. 1794–1803.

\bibitem{ZouTDGIA}
X.~Zou, Q.~Zheng, Y.~Dong, X.~Guan, E.~Kharlamov, J.~Lu, and J.~Tang, ``Tdgia:
  Effective injection attacks on graph neural networks,'' in \emph{Proceedings
  of the 27th ACM SIGKDD International Conference on Knowledge Discovery \&
  Data Mining}, 2021, p. 2461–2471.

\bibitem{feng2019graph}
F.~Feng, X.~He, J.~Tang, and T.-S. Chua, ``Graph adversarial training:
  Dynamically regularizing based on graph structure,'' \emph{IEEE Transactions
  on Knowledge and Data Engineering}, vol.~33, no.~6, pp. 2493--2504, 2021.

\bibitem{Dai2019AdversarialTM}
Q.~Dai, X.~Shen, L.~Zhang, Q.~Li, and D.~Wang, ``Adversarial training methods
  for network embedding,'' in \emph{Proceedings of The Web Conference 2019},
  ser. WWW '19, 2019, pp. 329--339.

\bibitem{kong2020flag}
K.~Kong, G.~Li, M.~Ding, Z.~Wu, C.~Zhu, B.~Ghanem, G.~Taylor, and T.~Goldstein,
  ``Robust optimization as data augmentation for large-scale graphs,'' in
  \emph{Proceedings of the IEEE conference on computer vision and pattern
  recognition}, ser. CVPR'22, 2022.

\bibitem{ZhangGNNGuard2020}
X.~Zhang and M.~Zitnik, ``Gnnguard: Defending graph neural networks against
  adversarial attacks,'' in \emph{Proceedings of Neural Information Processing
  Systems}, ser. NeurIPS '20, 2020, pp. 9263--9275.

\bibitem{Zhu2019RobustGC}
D.~Zhu, Z.~Zhang, P.~Cui, and W.~Zhu, ``Robust graph convolutional networks
  against adversarial attacks,'' in \emph{Proceedings of the 25th ACM SIGKDD
  International Conference on Knowledge Discovery \& Data Mining}, ser. KDD
  '19, 2019, pp. 1399--1407.

\bibitem{zhang2019comparing}
Y.~Zhang, S.~Khan, and M.~Coates, ``Comparing and detecting adversarial attacks
  for graph deep learning,'' in \emph{Proc. Representation Learning on Graphs
  and Manifolds Workshop, Int. Conf. Learning Representations, New Orleans, LA,
  USA}, ser. RLGM @ ICLR '19, 2019.

\bibitem{Entezari2020AllYN}
N.~Entezari, S.~A. Al-Sayouri, A.~Darvishzadeh, and E.~E. Papalexakis, ``All
  you need is low (rank): Defending against adversarial attacks on graphs,'' in
  \emph{Proceedings of the 13th International Conference on Web Search and Data
  Mining}, ser. WSDM '20, 2020, pp. 169--177.

\bibitem{Jin2020GraphSL}
W.~Jin, Y.~Ma, X.~Liu, X.-F. Tang, S.~Wang, and J.~Tang, ``Graph structure
  learning for robust graph neural networks,'' in \emph{Proceedings of the 26th
  ACM SIGKDD International Conference on Knowledge Discovery \& Data Mining},
  ser. KDD '20, 2020.

\bibitem{Wu2019AdversarialEF}
H.~Wu, C.~Wang, Y.~Tyshetskiy, A.~Docherty, K.~Lu, and L.~Zhu, ``Adversarial
  examples for graph data: Deep insights into attack and defense,'' in
  \emph{Proceedings of the 28th International Joint Conference on Artificial
  Intelligence}, ser. IJCAI '19, 2019, pp. 4816--4823.

\bibitem{Xu2019TopologyAA}
K.~Xu, H.~Chen, S.~Liu, P.-Y. Chen, T.-W. Weng, M.~Hong, and X.~Lin, ``Topology
  attack and defense for graph neural networks: An optimization perspective,''
  in \emph{Proceedings of the 28th International Joint Conference on Artificial
  Intelligence}, ser. IJCAI '19, 2019, pp. 3961--3967.

\bibitem{Bojchevski2019CertifiableRT}
A.~Bojchevski and S.~G\"{u}nnemann, ``Certifiable robustness to graph
  perturbations,'' in \emph{Proceedings of Neural Information Processing
  Systems}, ser. NeurIPS '19, 2019, pp. 8319--8330.

\bibitem{zugner2020CertiRob}
D.~Z{\"u}gner and S.~G{\"u}nnemann, ``Certifiable robustness of graph
  convolutional networks under structure perturbations,'' in \emph{Proceedings
  of the 26th ACM SIGKDD International Conference on Knowledge Discovery \&
  Data Mining}, ser. KDD '20, 2020.

\bibitem{Liu2020CertifiableRT}
Y.~Liu, X.~Xia, L.~Chen, X.~He, C.~Yang, and Z.~Zheng, ``Certifiable robustness
  to discrete adversarial perturbations for factorization machines,'' in
  \emph{Proceedings of the 43rd International ACM SIGIR Conference on Research
  and Development in Information Retrieval}, ser. SIGIR '20, 2020, pp.
  419--428.

\bibitem{zugner_adversarial_2019}
D.~Z{\"u}gner and S.~G{\"u}nnemann, ``Adversarial attacks on graph neural
  networks via meta learning,'' in \emph{International Conference on Learning
  Representations}, 2019.

\bibitem{bojchevski_sparsesmoothing_2020}
A.~Bojchevski, J.~Klicpera, and S.~G{\"u}nnemann, ``Efficient robustness
  certificates for discrete data: Sparsity-aware randomized smoothing for
  graphs, images and more,'' in \emph{Proceedings of the 37th International
  Conference on Machine Learning}, ser. ICML '20, 2020, pp. 11\,647--11\,657.

\bibitem{WangJCG21}
B.~Wang, J.~Jia, X.~Cao, and N.~Z. Gong, ``Certified robustness of graph neural
  networks against adversarial structural perturbation,'' in \emph{Proceedings
  of the 27th ACM SIGKDD International Conference on Knowledge Discovery \&
  Data Mining}, 2021, pp. 1645--1653.

\bibitem{SchuchardtBKG21}
J.~Schuchardt, A.~Bojchevski, J.~Klicpera, and S.~G{\"{u}}nnemann, ``Collective
  robustness certificates: Exploiting interdependence in graph neural
  networks,'' in \emph{International Conference on Learning Representations},
  2021.

\bibitem{Jin2020AdversarialAA}
W.~Jin, Y.~Li, H.~Xu, Y.~Wang, and J.~Tang, ``Adversarial attacks and defenses
  on graphs: A review and empirical study,'' \emph{ArXiv}, vol. abs/2003.00653,
  2020.

\bibitem{Tao2021AdvImmune}
S.~Tao, H.~Shen, Q.~Cao, L.~Hou, and X.~Cheng, ``Adversarial immunization for
  certifiable robustness on graphs,'' in \emph{Proceedings of the 14th ACM
  International Conference on Web Search and Data Mining}, ser. WSDM'21, 2021.

\bibitem{Sun2020AdversarialAO}
Y.~Sun, S.~Wang, X.-F. Tang, T.-Y. Hsieh, and V.~G. Honavar, ``Adversarial
  attacks on graph neural networks via node injections: A hierarchical
  reinforcement learning approach,'' in \emph{Proceedings of The Web Conference
  2020}, ser. WWW '20, 2020, pp. 673--683.

\bibitem{Wang2020ScalableAO}
J.~Wang, M.~Luo, F.~Suya, J.~Li, Z.~Yang, and Q.~Zheng, ``Scalable attack on
  graph data by injecting vicious nodes,'' \emph{arXiv preprint
  arXiv:2004.13825}, 2020.

\bibitem{velickovic2018graph}
P.~Veli{\v{c}}kovi{\'{c}}, G.~Cucurull, A.~Casanova, A.~Romero, P.~Li{\`{o}},
  and Y.~Bengio, ``{Graph Attention Networks},'' in \emph{International
  Conference on Learning Representations}, 2018.

\bibitem{Page1999ThePC}
\BIBentryALTinterwordspacing
L.~Page, S.~Brin, R.~Motwani, and T.~Winograd, ``The pagerank citation ranking
  : Bringing order to the web,'' in \emph{The Web Conference}, 1999. [Online].
  Available: \url{https://api.semanticscholar.org/CorpusID:1508503}
\BIBentrySTDinterwordspacing

\bibitem{Finn2017ModelAgnosticMF}
C.~Finn, P.~Abbeel, and S.~Levine, ``Model-agnostic meta-learning for fast
  adaptation of deep networks,'' in \emph{Proceedings of the 34th International
  Conference on Machine Learning}, ser. ICML '17, 2017, pp. 1126--1135.

\bibitem{de2011second}
P.~de~Fouquieres, S.~G. Schirmer, S.~J. Glaser, and I.~Kuprov, ``Second order
  gradient ascent pulse engineering,'' \emph{Journal of Magnetic Resonance},
  vol. 212, no.~2, pp. 412--417, 2011.

\bibitem{Leskovec2007CELF}
J.~Leskovec, A.~Krause, C.~Guestrin, C.~Faloutsos, J.~M. VanBriesen, and N.~S.
  Glance, ``Cost-effective outbreak detection in networks,'' in
  \emph{Proceedings of the 13th ACM SIGKDD International Conference on
  Knowledge Discovery \& Data Mining}, ser. KDD '07, 2007, pp. 420--429.

\bibitem{Zeng2019GraphSAINTGS}
H.~Zeng, H.~Zhou, A.~Srivastava, R.~Kannan, and V.~Prasanna, ``{GraphSAINT}:
  Graph sampling based inductive learning method,'' in \emph{International
  Conference on Learning Representations}, 2020.

\bibitem{Trustworthy2022}
H.~Zhang, B.~Wu, X.~Yuan, S.~Pan, H.~Tong, and J.~Pei, ``Trustworthy graph
  neural networks: Aspects, methods and trends,'' \emph{ArXiv}, vol.
  abs/2205.07424, 2022.

\bibitem{Zgner2019CertifiableRA}
D.~Z{\"u}gner and S.~G{\"u}nnemann, ``Certifiable robustness and robust
  training for graph convolutional networks,'' in \emph{Proceedings of the 25th
  ACM SIGKDD International Conference on Knowledge Discovery \& Data Mining},
  ser. KDD '19, 2019, pp. 246--256.

\bibitem{Girvan2002CommunitySI}
M.~Girvan and M.~E. Newman, ``Community structure in social and biological
  networks,'' \emph{Proceedings of the national academy of sciences}, vol.~99,
  no.~12, pp. 7821--7826, 2002.

\bibitem{Sun2018AdversarialAA}
L.~Sun, J.~Wang, P.~S. Yu, and B.~Li, ``Adversarial attack and defense on graph
  data: A survey,'' \emph{ArXiv}, vol. abs/1812.10528, 2018.

\bibitem{Madry2017TowardsDL}
A.~Madry, A.~Makelov, L.~Schmidt, D.~Tsipras, and A.~Vladu, ``Towards deep
  learning models resistant to adversarial attacks,'' in \emph{International
  Conference on Learning Representations}, 2018.

\bibitem{hamilton2017inductive}
W.~Hamilton, Z.~Ying, and J.~Leskovec, ``Inductive representation learning on
  large graphs,'' in \emph{Proceedings of Neural Information Processing
  Systems}, ser. NIPS '17, 2017, pp. 1024--1034.

\bibitem{rieck2019persistent}
B.~Rieck, C.~Bock, and K.~Borgwardt, ``A persistent weisfeiler-lehman procedure
  for graph classification,'' in \emph{Proceedings of the 36th International
  Conference on Machine Learning}, ser. ICML '19, 2019, pp. 5448--5458.

\bibitem{Chen2020ASO}
L.~Chen, J.~Li, J.~Peng, T.~Xie, Z.~Cao, K.~Xu, X.~He, and Z.~Zheng, ``A survey
  of adversarial learning on graphs,'' \emph{ArXiv}, vol. abs/2003.05730, 2020.

\bibitem{Wang2019AttackingGC}
B.~Wang and N.~Z. Gong, ``Attacking graph-based classification via manipulating
  the graph structure,'' in \emph{Proceedings of the 2019 ACM SIGSAC Conference
  on Computer and Communications Security}, ser. CCS '19, 2019, pp. 2023--2040.

\bibitem{Zhang2022Proj}
H.~Zhang, X.~Yuan, C.~Zhou, and S.~Pan, ``Projective ranking-based gnn evasion
  attacks,'' \emph{IEEE Transactions on Knowledge and Data Engineering}, 2022.

\bibitem{zugner2020AdversarialA}
D.~Z\"{u}gner, O.~Borchert, A.~Akbarnejad, and S.~G\"{u}nnemann, ``Adversarial
  attacks on graph neural networks: Perturbations and their patterns,''
  \emph{ACM Trans. Knowl. Discov. Data}, vol.~14, no.~5, 2020.

\bibitem{Li2021AdversarialA}
J.~Li, T.~Xie, C.~Liang, F.~Xie, X.~He, and Z.~Zheng, ``Adversarial attack on
  large scale graph,'' \emph{IEEE Transactions on Knowledge and Data
  Engineering}, pp. 1--1, 2021.

\bibitem{Chang20ARestrict}
H.~Chang, Y.~Rong, T.~Xu, W.~Huang, H.~Zhang, P.~Cui, W.~Zhu, and J.~Huang, ``A
  restricted black-box adversarial framework towards attacking graph embedding
  models,'' in \emph{Proceedings of the 34th {AAAI} Conference on Artificial
  Intelligence}.

\bibitem{Chang2022Adversarial}
H.~Chang, Y.~Rong, T.~Xu, W.~Huang, H.~Zhang, P.~Cui, X.~Wang, W.~Zhu, and
  J.~Huang, ``Adversarial attack framework on graph embedding models with
  limited knowledge,'' \emph{IEEE Transactions on Knowledge and Data
  Engineering}, pp. 1--1, 2022.

\bibitem{Xu2022GraphSanitation}
Z.~Xu, B.~Du, and H.~Tong, ``Graph sanitation with application to node
  classification,'' in \emph{Proceedings of The Web Conference 2022}, ser. WWW
  '22, 2022, pp. 1136--1147.

\bibitem{Zhang2018BayesianGC}
Y.~Zhang, S.~Pal, M.~Coates, and D.~{\"U}stebay, ``Bayesian graph convolutional
  neural networks for semi-supervised classification,'' in \emph{Proceedings of
  the 33th {AAAI} Conference on Artificial Intelligence}.

\bibitem{Tang2020TransferringRF}
X.-F. Tang, Y.~Li, Y.~Sun, H.~Yao, P.~Mitra, and S.~Wang, ``Transferring
  robustness for graph neural network against poisoning attacks,'' in
  \emph{Proceedings of the 13th International Conference on Web Search and Data
  Mining}, ser. WSDM '20, 2020, pp. 600--608.

\bibitem{Jia2020CertifiedRO}
J.~Jia, B.~Wang, X.~Cao, and N.~Z. Gong, ``Certified robustness of community
  detection against adversarial structural perturbation via randomized
  smoothing,'' in \emph{Proceedings of The Web Conference 2020}, ser. WWW '20,
  2020, pp. 2718--2724.

\end{thebibliography}
%
% <OR> manually copy in the resultant .bbl file
% set second argument of \begin to the number of references
% (used to reserve space for the reference number labels box)
%\begin{thebibliography}{1}
%
%\bibitem{IEEEhowto:kopka}
%H.~Kopka and P.~W. Daly, \emph{A Guide to \LaTeX}, 3rd~ed.\hskip 1em plus
%  0.5em minus 0.4em\relax Harlow, England: Addison-Wesley, 1999.
%
%\end{thebibliography}

% biography section
% 
% If you have an EPS/PDF photo (graphicx package needed) extra braces are
% needed around the contents of the optional argument to biography to prevent
% the LaTeX parser from getting confused when it sees the complicated
% \includegraphics command within an optional argument. (You could create
% your own custom macro containing the \includegraphics command to make things
% simpler here.)
%\begin{IEEEbiography}[{\includegraphics[width=1in,height=1.25in,clip,keepaspectratio]{mshell}}]{Michael Shell}
% or if you just want to reserve a space for a photo:
%\newpage

\begin{IEEEbiography}[{\includegraphics[width=1in,height=1.25in,clip,keepaspectratio]{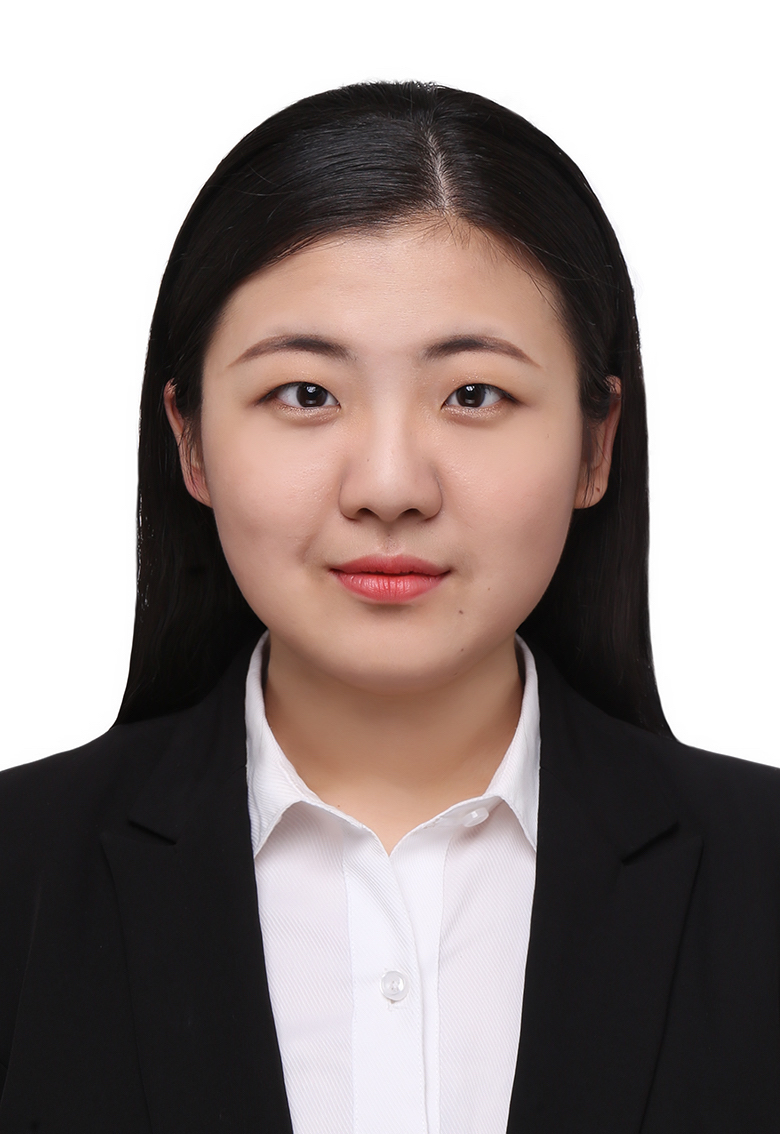}}]{Shuchang Tao}
is currently a Ph.D. student at the Data Intelligence System Research Center, Institute of Computing Technology, Chinese Academy of Sciences (CAS) and University of Chinese Academy of Sciences (UCAS). She received her Bachelor Degree from Tianjin University of China in 2018. Her research interests include graph neural networks, robustness, adversarial attack and defense.

\end{IEEEbiography}

% if you will not have a photo at all:
\begin{IEEEbiography}[{\includegraphics[width=1in,height=1.25in,clip,keepaspectratio]{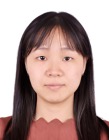}}]{Qi Cao}
is an associate professor at Institute of Computing Technology, Chinese Academy of Sciences (CAS). She received her Ph.D. degree from Institute of Computing Technology, CAS in 2020. Her research interests include social media computing, trustworthy graph machine learning.
\end{IEEEbiography}

% insert where needed to balance the two columns on the last page with
% biographies
%\newpage

\begin{IEEEbiography}[{\includegraphics[width=1in,height=1.25in,clip,keepaspectratio]{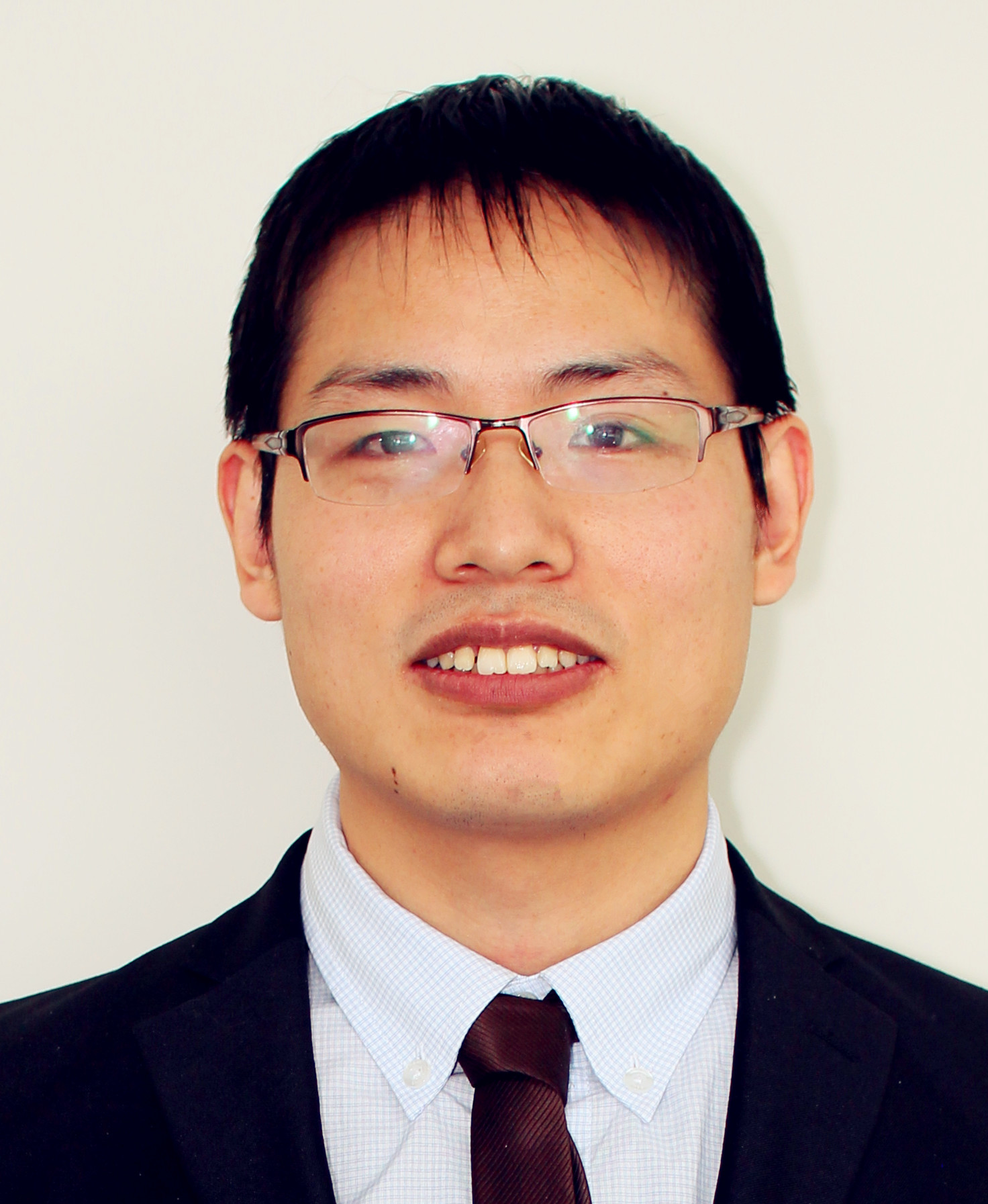}}]{Huawei Shen}
is a professor in the Institute of Computing Technology, Chinese Academy of Sciences (CAS) and the University of Chinese Academy of Sciences. He received his PhD degree from the Institute of Computing Technology in 2010. His major research interests include network science, social media analytics and artificial intelligence. He has published more than 100 papers in prestigious journals and top international conferences. He was the Track Chair, Area Chair, and Program Committee Member of more than 30 international conferences.
\end{IEEEbiography}

\begin{IEEEbiography}[{\includegraphics[width=1in,height=1.25in,clip,keepaspectratio]{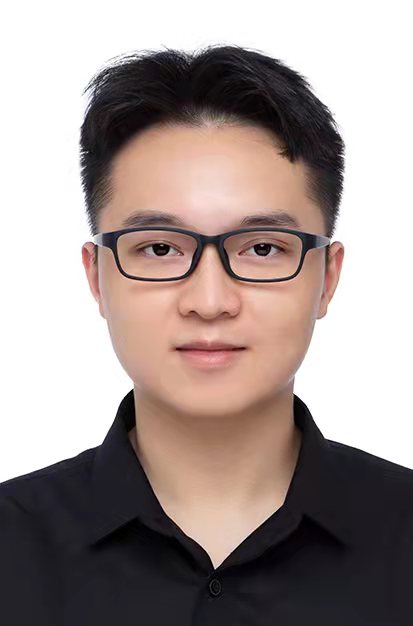}}]{Yunfan Wu}
is currently a Ph.D. candidate at Institute of Computing Technology, Chinese Academy of Sciences and University of Chinese Academy of Sciences, advised by Huawei Shen. He received the B.S. degree from University of Chinese Academy of Sciences in 2019. His research interests include recommender systems and adversary attacks.
\end{IEEEbiography}

\begin{IEEEbiography}[{\includegraphics[width=1in,height=1.25in,clip,keepaspectratio]{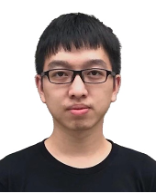}}]{Liang Hou}
is currently a senior algorithm engineer at Kuaishou Technology. Before that, he obtained the Ph.D. degree in computer science at Institute of Computing Technology, Chinese Academy of Sciences and University of Chinese Academy of Sciences in 2023, advised by Xueqi Cheng and Huawei Shen. He received the B.S. degree in computer science at Sichuan University in 2017. His research interests include generative adversarial networks, generative models, and efﬁcient deep learning.
\end{IEEEbiography}

\begin{IEEEbiography}
[{\includegraphics[width=1in,height=1.25in,clip,keepaspectratio]{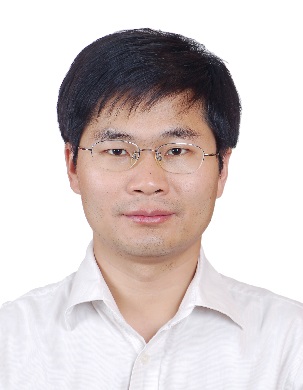}}]{Xueqi Cheng}
is a professor in the Institute of Computing Technology, Chinese Academy of Sciences (ICT-CAS) and the University of Chinese Academy of Sciences, and the director of the CAS Key Laboratory of Network Data Science and Technology. His main research interests include network science, web search and data mining, big data processing and distributed computing architecture. He has published more than 200 publications in prestigious journals and conferences, including IEEE Transactions on Information Theory, IEEE Transactions on Knowledge and Data Engineering, Journal of  Statistical Mechanics, Physical Review E., ACM SIGIR, WWW, ACM CIKM, WSDM, AAAI, IJCAI, ICDM, and so on. He has won the Best Paper Award in CIKM (2011), the Best Student Paper Award in SIGIR (2012), and the Best Paper Award Runner up of CIKM (2017). He is currently serving on the editorial board of the Journal of Computer Science and Technology, the Journal of Computer, and so on. He received the China Youth Science and Technology Award (2011), the Young Scientist Award of Chinese Academy of Sciences (2010), the Second prize for the National Science and Technology Progress (2012). He is a member of the IEEE.
\end{IEEEbiography}

% You can push biographies down or up by placing
% a \vfill before or after them. The appropriate
% use of \vfill depends on what kind of text is
% on the last page and whether or not the columns
% are being equalized.

%\vfill

% Can be used to pull up biographies so that the bottom of the last one
% is flush with the other column.
%\enlargethispage{-5in}

% that's all folks
\end{document}